\documentclass[10pt,twocolumn,letterpaper]{article}
\usepackage[pagenumbers]{cvpr} % To force page numbers, e.g. for an arXiv version

\usepackage[utf8]{inputenc} % allow utf-8 input
\usepackage[T1]{fontenc}    % use 8-bit T1 fonts
\usepackage{url}            % simple URL typesetting
\usepackage{booktabs}       % professional-quality tables
\usepackage{amsfonts}       % blackboard math symbols
\usepackage{nicefrac}       % compact symbols for 1/2, etc.
\usepackage{microtype}      % microtypography
\usepackage{comment}
\usepackage{multirow}
\usepackage[table,xcdraw]{xcolor}
\usepackage{xspace}
\usepackage{makecell} 
\usepackage{amssymb}% http://ctan.org/pkg/amssymb
\usepackage{pifont}% http://ctan.org/pkg/pifont
\usepackage{listings}
\usepackage{xcolor}
\usepackage{wrapfig}
\usepackage{tabularx}
\usepackage{siunitx}      % S column for aligned numbers
\usepackage{graphicx}      % already present in most templates
\usepackage{subcaption}    % for subfigure environments

\newcommand{\mypar}[1]{\vspace{1mm}\noindent\textbf{#1}}
\newcommand{\myparit}[1]{\vspace{1mm}\noindent\emph{#1}}

\def\eg{\emph{e.g.}\xspace}

\def\etc{\emph{etc.}\xspace}

\def\llm3{Kyvo\xspace}
\def\json3{\texttt{3Djson}\xspace}

\definecolor{darkgray}{gray}{0.3}
\definecolor{darkgreen}{RGB}{50, 150, 50}
\definecolor{darkred}{RGB}{180, 50, 50}
\definecolor{darkblue}{RGB}{80, 106, 156}
\definecolor{col1}{HTML}{506A88}
\definecolor{col2}{HTML}{2E3D47}
\definecolor{col3}{HTML}{98A9B5}
\definecolor{col4}{HTML}{D19C83}
\definecolor{fom}{RGB}{0,153,139}
\definecolor{greenkeyword}{RGB}{34,139,34}
\definecolor{blueattribute}{RGB}{0,0,255}
\definecolor{lightgray}{RGB}{250,250,245} % Light gray background

\usepackage[breakable, theorems, skins]{tcolorbox}

\usepackage{caption}
\usepackage{algorithm}
\usepackage{algorithmic}
\usepackage{longtable}
\usepackage{booktabs}
\usepackage{pifont}
\usepackage{cuted}

\definecolor{codegray}{gray}{0.95}   % light-gray background
\definecolor{codepink}{rgb}{1,0.97,0.97}  % light-pink background
\lstdefinestyle{pystyle}{
  language=Python,
  backgroundcolor=\color{codepink},
  basicstyle=\ttfamily\scriptsize,
  showstringspaces=false,
  numbers=none,
  frame=single,
  rulecolor=\color{codepink},        % border matches background
  breaklines=true,
  xleftmargin=0pt,
  mathescape=true
}

\definecolor{cvprblue}{rgb}{0.21,0.49,0.74}
\usepackage[pagebackref,breaklinks,colorlinks,allcolors=cvprblue]{hyperref}
\usepackage{placeins}

\def\paperID{****} % *** Enter the Paper ID here
\def\confName{CVPR}
\def\confYear{2026}

\title{Aligning Text, Images and 3D Structure Token-by-Token}

\author{Aadarsh Sahoo$^*$\quad Vansh Tibrewal$^*$\quad Georgia Gkioxari\\
California Institute of Technology\\
}

\begin{document}
% \maketitle

\twocolumn[{%
\renewcommand\twocolumn[1][]{#1}%
\maketitle
\begin{center}
    \centering
    \vspace{-8mm}
    \includegraphics[width=0.95\linewidth]{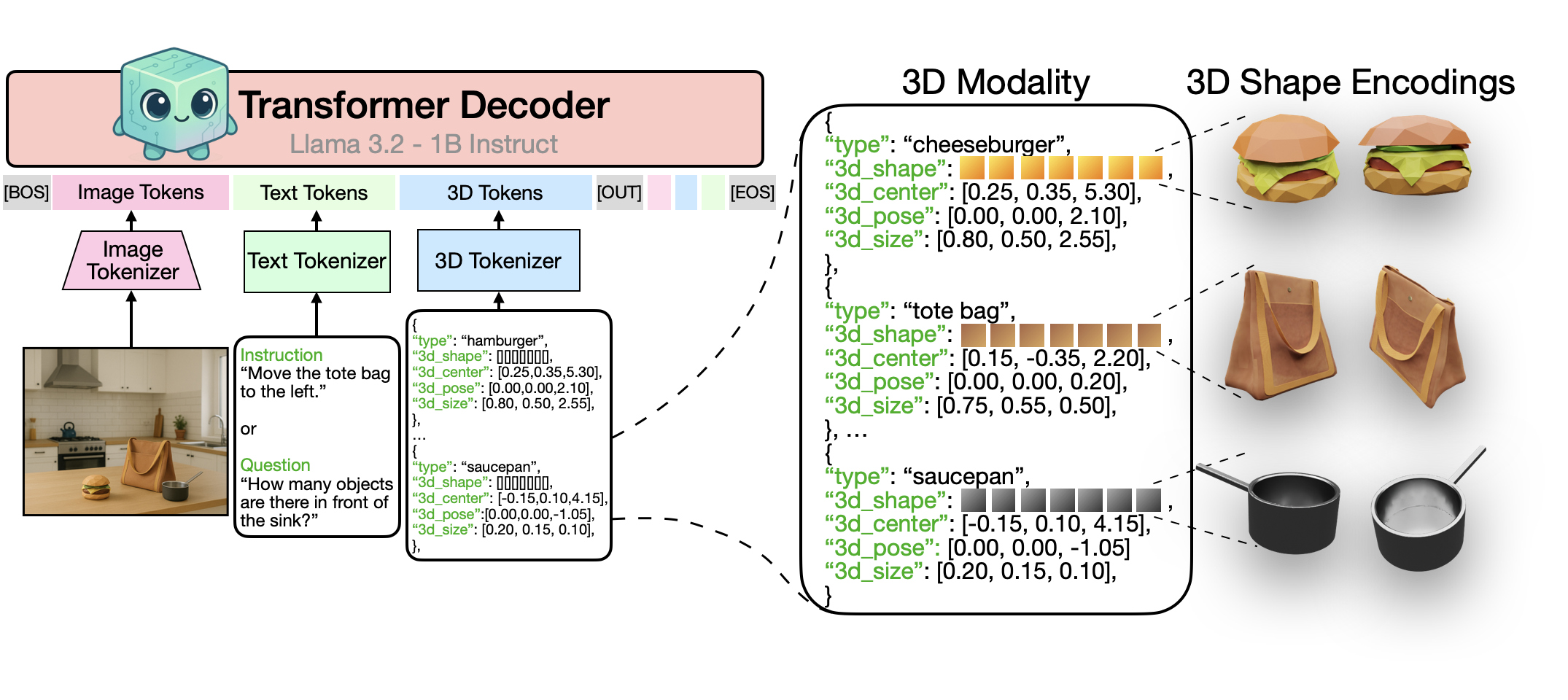}
    \vspace{-12mm}
    \captionof{figure}{Kyvo: a decoder-only transformer aligns a structured 3D modality with language and vision.  This 3D modality represents scenes as lists of objects, each defined by its 3D shape, type, 3D position, pose and size parameters. Kyvo unifies the token space of images, text, and 3D to enable a variety of complex visual 3D tasks.}
    \label{fig:teaser}
\end{center}
}]
\renewcommand{\thefootnote}{}
\footnotetext{$^*$Equal contribution}

\begin{abstract}
Creating machines capable of understanding the world in 3D is essential in assisting designers that build and edit 3D environments and robots navigating and interacting within a three-dimensional space.
Inspired by advances in language and image modeling, we investigate the potential of autoregressive models for a new modality: structured 3D scenes. 
To this end, we propose a unified LLM framework that aligns language, images, and 3D scenes and provide a detailed ``cookbook'' outlining critical design choices for achieving optimal training and performance addressing key questions related to data representation, modality-specific objectives, and more. We show how to tokenize complex 3D objects to incorporate into our structured 3D scene modality.
We evaluate performance across four core 3D tasks -- rendering, recognition, instruction-following, and question-answering -- and four 3D datasets, synthetic and real-world.
We show our model’s effectiveness on reconstructing complete 3D scenes consisting of complex objects from a single image and on real-world 3D object recognition tasks. Project webpage: \href{https://glab-caltech.github.io/kyvo/}{https://glab-caltech.github.io/kyvo}.
\end{abstract}    
\section{Introduction}
\label{sec:introduction}

Large language models (LLMs) that fuse text and images have unlocked unprecedented visual capabilities, such as visual captioning and text-guided image generation. Inspired by this success, we explore a third modality – structured 3D scenes – and show how aligning it with text and images allows LLMs to tackle a new suite of visual tasks in 3D, such as 3D reconstruction from a single image and 3D-conditioned image generation. We start from a language-pretrained transformer and extend it with a structured 3D modality that is designed to try to leverage its linguistic reasoning and generalization capabilities.

\begin{figure*}[!t]

\begin{center}

\includegraphics[width=\linewidth]{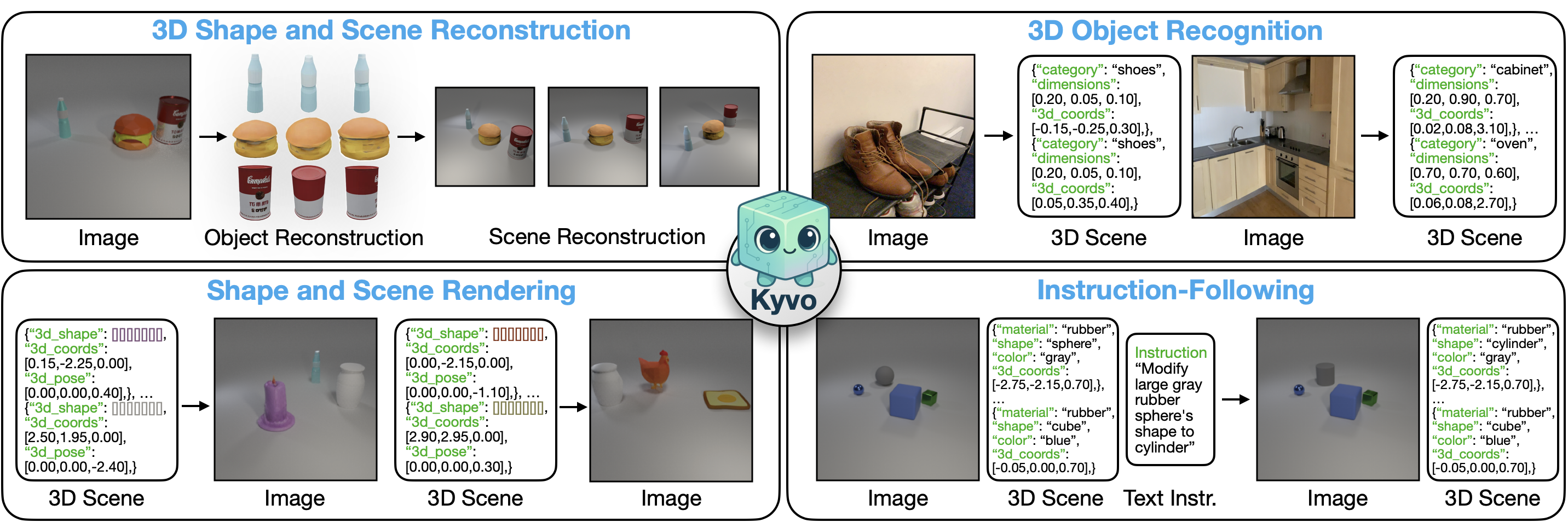} 

\end{center}

\vspace{-6mm}
\caption{\small\textbf{3D task examples} with Kyvo's unified autoregressive framework using a structured 3D modality. (1) \textit{3D shape and scene reconstruction:} From a single input image, Kyvo reconstructs individual objects with accurate geometry and spatial relationships. (2) \textit{3D object recognition:} Given an input image, Kyvo identifies objects and predicts their 3D positions in real-world scenes. (3) \textit{Shape and scene rendering:} Kyvo generates semantically consistent images from structured 3D scene inputs. (4) \textit{Instruction-Following:} Given an image, 3D scene and text instruction, Kyvo produces coherent modifications to both image and the 3D representation.
}
\label{fig:task-teaser}
\vspace{-4mm}
\end{figure*}

Our structured 3D modality encodes a scene as a list of its objects, where every object is specified by its 3D shape (\eg, 3D mesh, 3DGS), type, 3D position, 3D size and 3D pose, shown in Fig.~\ref{fig:teaser}. 
This modality captures aspects of the physical world that are not directly conveyed through language or images alone.
Additionally, it slots naturally into the unified token space shared by language and vision through an object-by-object tokenization scheme that integrates seamlessly with image and text tokens, so any modality can serve as input or output.
As a result, it supports a broad range of tasks in 3D, such as image generation conditioned on the 3D scene structure (3D → image), predicting 3D objects, their shapes and locations from a single image (image → 3D), and language-guided object-centric 3D editing (3D + image + text → 3D + image) -- all within the same model design -- shown in Fig.~\ref{fig:task-teaser}.
These capabilities can reshape workflows. Robots can parse an image into a 3D scene composed of 3D objects, while designers can create complex scenes of objects through language in one forward pass of an LLM instead of wrestling with hard-to-use software like Blender.

But, how does one design and train an LLM that aligns this structured 3D modality with image and text? 
While there is extensive work on language-only~\cite{achiam2023gpt,grattafiori2024llama3herdmodels,team2023gemini} or vision-and-language models~\cite{tong2024cambrian,laurenccon2024matters,beyer2024paligemma}, there is limited work on training with an additional structured 3D modality.
We explore the vast design space and evaluate the impact of design choices in architecture, training strategies, data representation, modality-specific objectives, and more.
Our testbed consists of four core 3D tasks -- image generation, recognition, instruction-following, and question-answering -- and four challenging datasets, both synthetic and real-world. 
Through an extensive ``cookbook'' of rigorously validated empirical findings we hope to provide guidelines to design 3D-aligned multi-modal LLMs, including guidelines on tokenizing complex 3D shapes.
We demonstrate the effectiveness of our unified LLM framework, \llm3 (Greek for ``3D cube''), on single-view 3D reconstruction of complex object geometries.
Finally, we apply our model to real-world domains of indoor and outdoor scenes for the task of image-based 3D object recognition, where we show that our model can tackle tasks that previously required task-specific vision specialists.

\section{Related Work}
\label{sec:relatedwork}

\mypar{LLMs.} The success of LLMs is attributed to the scalability of transformers~\cite{vaswani2017attention} taking the form of encoder-decoder~\cite{tay2022ul2, chung2024scaling} and most popular decoder-only~\cite{achiam2023gpt, grattafiori2024llama3herdmodels, team2023gemini} models. LLMs showcase the effectiveness of autoregressive formulations in achieving task generalization, a finding we extend to 3D structured representations.

\mypar{Vision-Language Models (VLMs).} 
Early VLMs~\cite{radford2021learning, jia2021scaling, alayrac2022flamingo} learned from image–text pairs for tasks like zero-shot classification, retrieval, and QA.
Modern VLMs~\cite{liu2024visual, chen2024internvl, lu2024deepseek, ma2024janusflow, wang2024qwen2, wu2024deepseek} extend these abilities using internet-scale data and stronger alignment.
Recent studies~\cite{laurenccon2024matters, beyer2024paligemma, tong2024cambrian} analyze vision-centric design choices but treat images only as input, while others~\cite{team2024chameleon, zhou2024transfusion, tong2024metamorph} explore early fusion and diffusion for image generation.
Nonetheless, current VLMs remain weak in 3D reasoning and geometry prediction, tasks that we tackle in this work.

\mypar{LLMs \& 3D.} 
Prior work integrates LLMs with diverse 3D formats for tasks like VQA and grounding~\cite{ma2024llms}. Some~\cite{ma2024spatialpin,cheng2024spatialrgpt,liao2024reasoning} extend VLMs to 3D via depth-aware queries, while others~\cite{hong20233d,tang2024minigpt,xu2025pointllm,ma2024find,marsili2025visual,kerr2023lerf,shen2023distilled} embed 3D data (\eg, point clouds, NeRFs) with CLIP-style features for open-ended QA.
3D-LLM~\cite{hong20233d} uses holistic 3D point clouds for captioning and QA, whereas our model’s structured 3D modality decomposes scenes into objects, enabling direct alignment across language and vision. This supports tasks like 3D shape and pose prediction from a single image and image generation from object-centric 3D inputs.
Similarly, while SceneScript~\cite{avetisyan2025scenescript} autoregressively predicts 3D boxes from videos and full scene clouds, our approach aligns images with structured 3D object and scene representations.

\mypar{3D Tokenization.} 
Prior work on 3D tokenization targets limited tasks, mainly single-asset generation. SAR3D~\cite{chen2024sar3d} uses triplanes, while AToken~\cite{lu2025atokenunifiedtokenizervision} builds on Trellis~\cite{xiang2024structured} to train a joint tokenizer for images, 3D, and video.
In contrast, we design a 3D tokenizer for structured scene encoding, optimized for compactness to represent multiple objects per scene. Whereas AToken and SAR3D use over $20\mathrm{k}$ and $2040$ tokens per asset respectively, our tokenizer achieves comparable or better reconstruction with far fewer tokens (see~\cref{subsection:cookbook}).

\section{Building our model token-by-token}
\label{section:building-our-model}

We present our approach, \llm3, that aligns the 3D modality with text and images. The result is a unified multimodal LLM framework that can perform a range of visual 3D tasks -- rendering, reconstruction, recognition and instruction-following -- as shown in~\cref{fig:task-teaser}.

~\cref{subsection:setup-tasks-and-datasets} outlines our experimental setup, including tasks and datasets. \cref{subsection:cookbook} presents a bottom-up analysis of key design choices -- covering data representation, 3D tokenization, sequence design, and scaling -- and addresses challenges in modality-specific tokenization and optimization. We summarize empirical insights and adopt optimal configurations for each subsequent experiment. Finally, we explore generalization to complex scene layouts (\cref{subsection:generalization-to-complex-shapes}), 3D shape representations (\cref{subsection:generalization-to-open-geometry}), and real-world recognition (\cref{subsection:generalization-to-real-world-recognition}).

\subsection{Setup: tasks and datasets}
\label{subsection:setup-tasks-and-datasets}

We design and train an autoregressive model aligning 3D with images and language, capable of performing 3D tasks.

\mypar{Tasks.} We focus on four core 3D tasks: image generation from 3D scene specifications (rendering), 3D object recognition \& reconstruction from a single image, instruction-following 3D editing and question-answering. 

\myparit{Rendering}: 3D → Image.
Given a 3D scene described by our structured modality (object types, shapes, locations, poses), the model generates a corresponding image. This task tests whether rendering -- typically requiring tools (\eg, Blender) and complex processes (\eg, ray tracing, rasterization) -- can be reframed as feed-forward next-token prediction.

\myparit{Reconstruction}: Image → 3D.
This is a dual of rendering. Given an image, the model predicts the underlying 3D scene structure, including object shapes, types, positions and poses.

\myparit{Instruction-Following}: $(\text{Image}, \text{3D}, \text{Text}_\text{I})$ → $(\text{Image}, \text{3D})$.
We define a set of tasks to manipulate and modify 3D scenes given a text instruction. 
These include: (1) \textit{modifying the appearance of objects}, (2) \textit{adding new objects}, (3) \textit{removing objects}, and (4) \textit{moving an object to a desired location}. For each subtask, we craft templated natural language instructions to evaluate the model's ability to follow instructions in 3D, \eg ``Remove the red mug behind the yellow bottle''. 

\myparit{Question-Answering}: $(\text{Image}, \text{3D}, \text{Text}_\text{Q})$ → $\text{Text}_\text{A}$.
We generate QA pairs using templated questions. Given the 3D scene, an image, and a query (\eg “Is there a green object the same size as the glass vase?”), the model predicts a natural language answer. Details are in the Appendix.

All four tasks require spatial reasoning and grounding. Two of them explicitly involve text as input or output and all four implicitly require linguistic reasoning in the use of the structured 3D modality as input or output.

\mypar{Datasets.} 
We consider four datasets for our experiments. 
Two synthetic ones: CLEVR~\cite{johnson2017clevr} features scenes of simple shapes in varying layouts; ObjaWorld features complex objects of any geometry sourced from Objaverse~\cite{deitke2023objaverse}.
Two real-world ones: Objectron~\cite{ahmadyan2021objectron} and ARKitScenes~\cite{dehghan2021arkitscenes}
comprising real-world indoor and outdoor scenes of various object types.

\mypar{Evaluation.} 
Per our task definitions, Kyvo's output can be of either modality: text, image, or 3D. 

\myparit{Recognition.} 
We evaluate predicted 3D scenes using the Jaccard Index,
$J = \frac{tp}{tp + fp + fn}$,
which measures object-level agreement between prediction and ground truth based on matching attributes (type, size, color, material) and spatial proximity within threshold $\tau$.
True positives are matched objects within $\tau$, false positives are extra predictions, and false negatives are misses.
We report the mean Jaccard Index over $\tau \in \{0.05, 0.10, 0.15, 0.20, 0.25\}$, reflecting both recognition and spatial accuracy.

\myparit{Question-Answering.} 
We report answer accuracy based on exact match between predicted and ground-truth text (typically 1–2 tokens). A random baseline yields $0.359$ accuracy, while a frequency-based baseline reaches $0.430$. See Appendix for more details.

\myparit{Rendering.} 
In the rendering task, models generate images from 3D scene inputs. Standard image metrics (L2, SSIM~\cite{wang2004image}, FID~\cite{heusel2017gans}, PSNR) fail to capture object placement or attribute accuracy.
We therefore use human evaluations: annotators rank anonymized outputs against ground truth, and we report the mean rank. While L2 and SSIM roughly follow these trends, human judgment better reflects scene correctness (details in Appendix).

\myparit{Instruction-Following.} 
Here, the model predicts both images and 3D scenes. We report the Jaccard Index for 3D outputs, as this is the primary focus of our work.

%%%%%%%%%%%%%

\subsection{Cookbook}
\label{subsection:cookbook}
 
Our model, \llm3, starts from language as the principal modality. Our backbone is the decoder-only Llama-3.2-1B-Instruct~\cite{grattafiori2024llama3herdmodels} initialized from language-only pretrained weights. We extend it with modality-specific tokenizers for images and the structured 3D modality, along with modified input embeddings and output projections (\cref{fig:teaser}). We do this in order to leverage the generalization and reasoning capabilities of LLMs, a hypothesis we later confirm in \cref{table:training-recipes}.
We evaluate key architectural choices -- each with significant performance impact -- which we hope offer insights to guide future multimodal LLM development in the 3D domain.

We begin by exploring how to represent and tokenize our structured 3D modality. We explore the tokenization of individual 3D objects of diverse geometry and appearance from Objaverse as well as the tokenization of scenes containing multiple objects of known geometry in CLEVR.
This setup helps establish best practices for integrating 3D into a unified LLM framework. We then extend these findings to complex scenes in ObjaWorld (\cref{subsection:generalization-to-complex-shapes} \& \cref{subsection:generalization-to-open-geometry}) and validate generalization to real-world scenes in Objectron and ARKitScenes (\cref{subsection:generalization-to-real-world-recognition}).

\subsubsection{What is the optimal 3D representation?}

Our model handles three modalities -- images, text, and structured 3D scenes -- by converting each into token sequences for autoregressive modeling. 

\mypar{Text.} We employ an off-the-shelf text tokenizer from Llama-3.2~\cite{grattafiori2024llama3herdmodels} with a $128,\!000$ size vocabulary. Text instructions, questions, and answers are tokenized and enclosed within special, learnable tokens \texttt{[TEXT-START]} and \texttt{[TEXT-END]}.

\mypar{Images.} Our framework addresses tasks involving images as both inputs and outputs. To enable image generation within an autoregressive paradigm, we adopt discrete image representations using VQGAN~\cite{esser2021taming}. This approach maps continuous image features to discrete tokens through a learned codebook. Specifically, we fine-tune a pre-trained VQGAN model to optimize the codebook for our visual domain.
Image tokens are enclosed within special, learnable tokens, \texttt{[IMAGE-START]} and \texttt{[IMAGE-END]}. 

\mypar{3D scene tokenization.} 
We represent 3D scenes as structured token sequences, where each list element encodes one object.
Object attributes -- shape, size, location, color, material -- are expressed through special learnable tokens (e.g., \texttt{[SHAPE]}), whose values may be text (``car'', ``yellow''), numbers (for size or position), or learned 3D embeddings.
Each object is enclosed by \texttt{[OBJECT-START]} and \texttt{[OBJECT-END]}, and the full scene by \texttt{[SCENE-START]} and \texttt{[SCENE-END]}.
An example scene with two objects is shown below:

\lstset{
    basicstyle=\ttfamily\small,
    escapeinside={(*@}{@*)}, % Define escape delimiters
    breaklines=true,
    showstringspaces=false,
    breakindent=0pt,
    backgroundcolor=\color{lightgray} % Set light gray background
}
% Color definitions
\definecolor{assetcolor}{RGB}{204,0,255} % Brighter purple for [ASSET]
\definecolor{vcolor}{RGB}{0,150,255}      % Bright blue for vector elementselements

\begin{lstlisting}
(*@\textcolor{greenkeyword}{[SCENE-START]}@*)(*@\textcolor{greenkeyword}{[OBJECT-START]}@*)(*@\textcolor{assetcolor}{[SHAPE]}@*)<(*@\textcolor{vcolor}{$v_{1}^{1}$}@*),(*@\textcolor{vcolor}{$v_{2}^{1}$}@*),...,(*@\textcolor{vcolor}{$v_{512}^{1}$}@*)>(*@\textcolor{blueattribute}{[LOCATION]}@*)-0.15 1.05 0.00(*@\textcolor{blueattribute}{[POSE]}@*)0.00 0.00 3.00(*@\textcolor{greenkeyword}{[OBJECT-END]}@*)(*@\textcolor{greenkeyword}{[OBJECT-START]}@*)(*@\textcolor{assetcolor}{[SHAPE]}@*)<(*@\textcolor{vcolor}{$v_{1}^{2}$}@*),(*@\textcolor{vcolor}{$v_{2}^{2}$}@*),...,(*@\textcolor{vcolor}{$v_{512}^{2}$}@*)>(*@\textcolor{blueattribute}{[LOCATION]}@*)0.25 2.10 0.00(*@\textcolor{blueattribute}{[POSE]}@*)0.00 0.00 -2.05(*@\textcolor{greenkeyword}{[OBJECT-END]}@*)(*@\textcolor{greenkeyword}{[SCENE-END]}@*)
\end{lstlisting}

We discuss two key design decisions:

\begin{figure}[t]
% \vspace*{-3mm}
% \begin{center}
\centering
\includegraphics[width=\columnwidth, trim=2pt 2pt 2pt 2pt,clip]{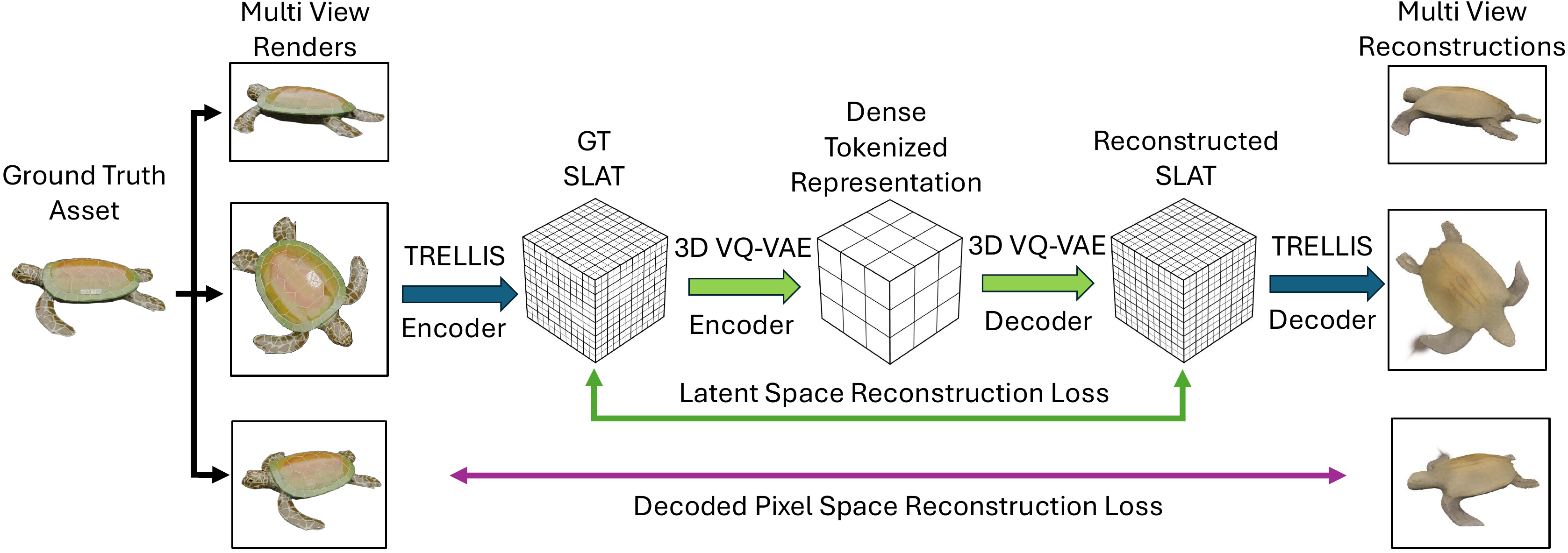}
% \end{center}
\vspace{-6mm}
\caption{\small 3D VQ-VAE training involves the standard VQ-VAE losses (including reconstruction loss) applied in latent space as well as an auxiliary reconstruction loss applied in decoded pixel space.
}
\label{fig:3d-vq-vae-training}
\vspace{-5mm}
\end{figure}

\begin{figure*}[t]
\centering
\setlength{\tabcolsep}{4pt} % horizontal padding between columns

\begin{tabular}{@{}c|c|c@{}}
% --- (a) auxiliary reconstruction loss ---
\begin{minipage}[t]{0.3\textwidth}
    \centering
    \includegraphics[width=\linewidth, trim=2pt 2pt 2pt 2pt,clip]{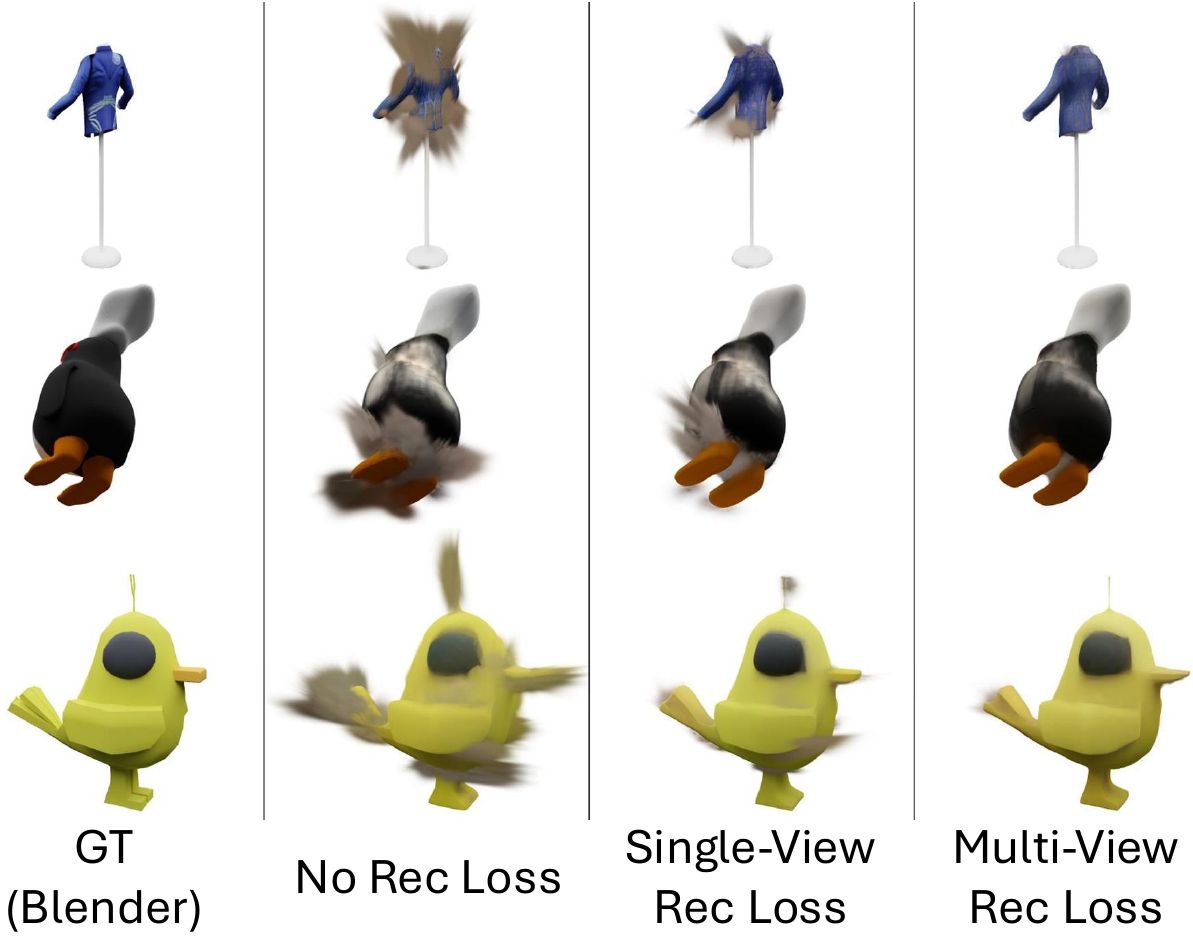}
    \phantomsubcaption\label{fig:decoded-rec-loss-comparison}
    \vspace{1mm}
    {\small (a)}
    \vspace{-2mm}
\end{minipage}
&
% --- (b) 3D tokenizer comparison ---
\begin{minipage}[t]{0.27\textwidth}
    \centering
    \includegraphics[width=\linewidth, trim=2pt 2pt 2pt 2pt,clip]{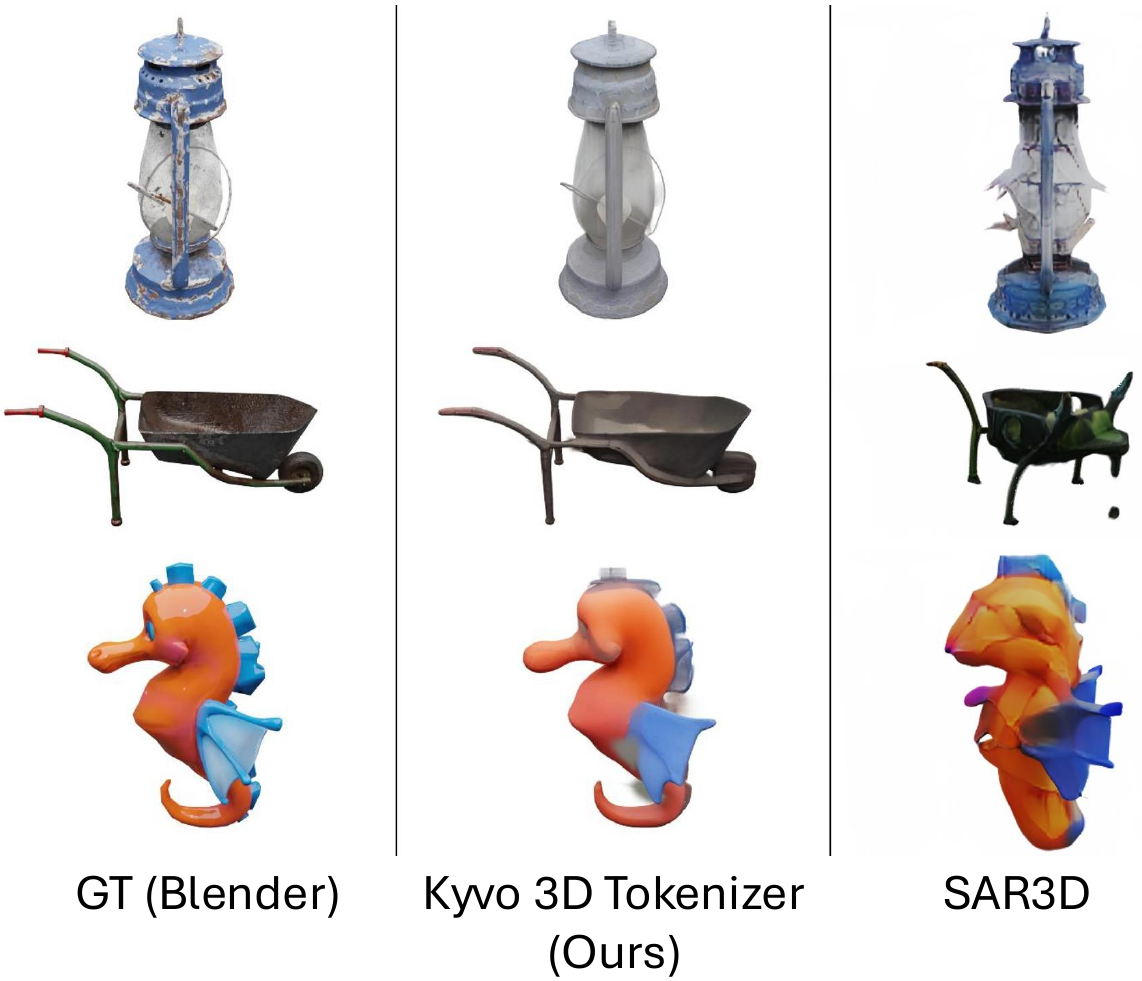}
    \phantomsubcaption\label{fig:3d-reconstruction-sar3d}
    \vspace{1mm}
    {\small (b)}
    \vspace{-2mm}
\end{minipage}
&
% --- (c) image-to-3d / 3d-to-image ---
\begin{minipage}[t]{0.38\textwidth}
    \centering
    \includegraphics[width=0.63\linewidth, trim=2pt 0pt 2pt 0pt,clip]{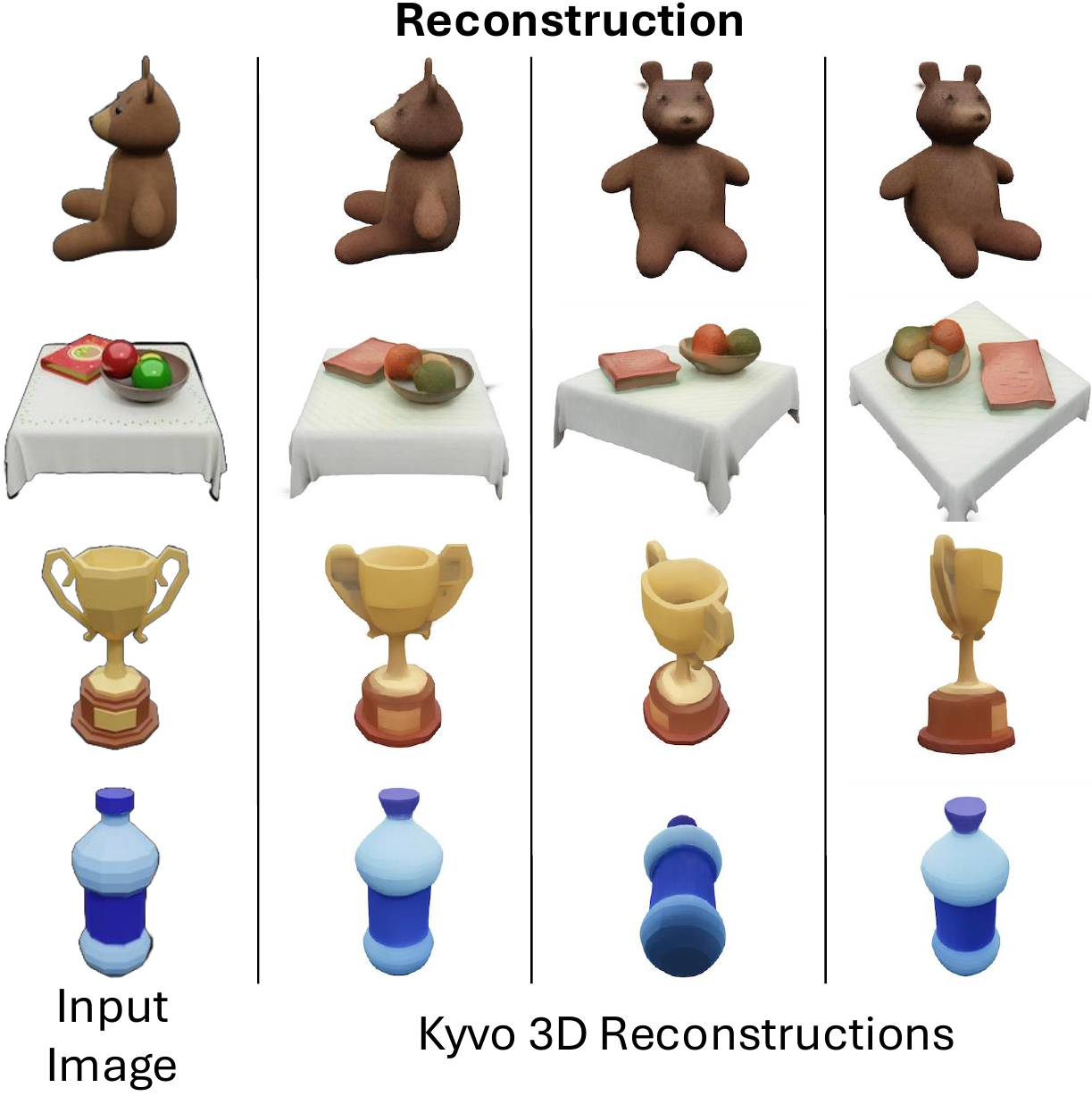}%
    \hfill
    \vrule width 0.5pt
    \hfill
    \includegraphics[width=0.35\linewidth, trim=2pt 0pt 2pt 0pt,clip]{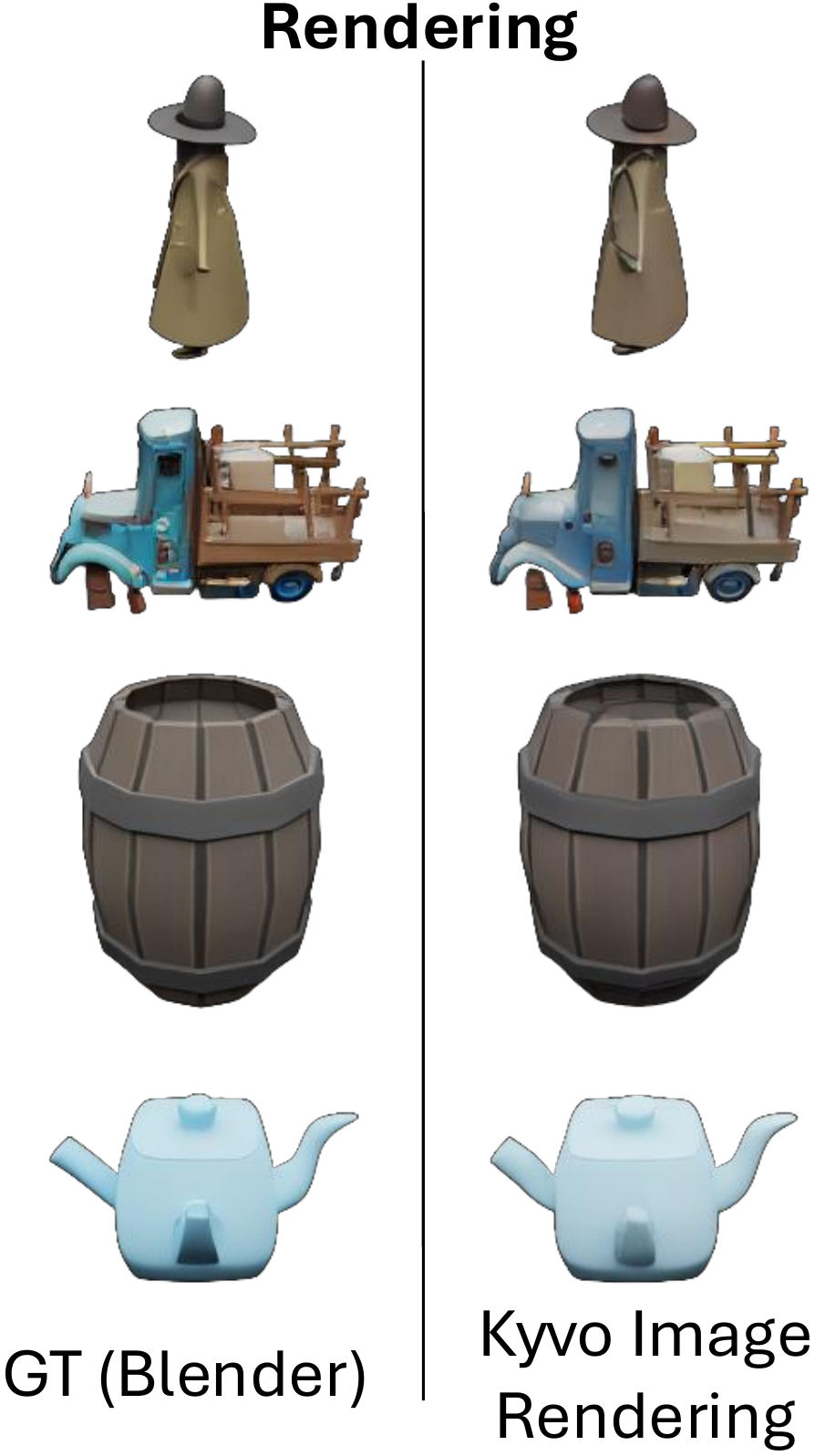}
    \phantomsubcaption\label{fig:asset-level}
    \vspace{1mm}
    {\small (c)}
    \vspace{-2mm}
\end{minipage}
\end{tabular}

\vspace{-2mm}
\caption{\small
\textbf{3D Tokenization findings and comparisons.}
(a) \textbf{Effect of auxiliary reconstruction loss.} An auxiliary pixel-space reconstruction loss on decoded renders from multiple views of the 3D object leads to much better reconstructions.
(b) \textbf{3D tokenizer comparison.} Reconstructions from our Trellis-based VQ-VAE exceed the quality of SAR3D reconstructions with fewer tokens; improved textures stem from the Trellis slat representation rather than triplanes.
(c) \textbf{Learned 3D shape encodings are effective during decoding.} The 3D tokens used in Kyvo are sufficient for both reconstruction and rendering using Llama 3.2 as decoder.
}
\label{fig:combined-recon}
\vspace{-4mm}
\end{figure*}
% ---------------- side-by-side tables in one column ----------------
\setlength{\tabcolsep}{3pt}
\begin{table}[t]
  \vspace{-4mm}
  \centering
  \scriptsize
  \renewcommand{\arraystretch}{0.95}
  \newcommand{\cmark}{\ding{51}} % ✓
  \newcommand{\xmark}{\ding{55}} % ✗

  \begin{minipage}[t]{0.48\columnwidth}
    \centering
    \begin{tabular}{@{}c S[table-format=1.3]@{}}
      \toprule
      \textbf{Aux Rec Loss} &
      {Mean Rank$({\downarrow})$} \\ \midrule
      \xmark & 2.828 \\
      Fixed Single View & 1.672 \\
      Multiple Views & $\mathbf{1.500}$ \\
      \bottomrule
    \end{tabular}
    \vspace{-2mm}
    \caption{\small Effect of auxiliary reconstruction loss.}
    \label{table:decoded-rec-loss}
  \end{minipage}
  \hfill
  \begin{minipage}[t]{0.48\columnwidth}
    \centering
    \begin{tabular}{@{}c S[table-format=1.3]@{}}
      \toprule
      \textbf{3D Tokenizer} &
      {Mean Rank$({\downarrow})$} \\ \midrule
      SAR3D & 1.605 \\
      Kyvo 3D VQ-VAE & $\mathbf{1.395}$ \\
      \bottomrule
    \end{tabular}
    \vspace{-2mm}
    \caption{\small 3D tokenization comparison using human evaluation.}
    \label{table:3d-reconstruction-sar3d}
  \end{minipage}

  \vspace{-6mm}
\end{table}
% -------------------------------------------------------------------

\mypar{(1) How to tokenize 3D shapes (geometry \& texture)?} 
We aim for our structured 3D modality to encode complex objects via compact, autoregressively decodable 3D shape representations. To this end, we adopt Trellis~\cite{xiang2024structured}, which encodes geometry and texture as sparse voxel features (slats) $z = \{(z_i, p_i)\}_{i=1}^L$, where each $z_i \in \mathbb{R}^8$ represents local features and $p_i$ indexes active voxels in an $N^3$ grid. Although slats are sparse ($L \ll N^3$, typically $L \approx 20\text{k}$ for $N=64$), their length makes autoregressive modeling intractable.

We therefore train a 3D VQ-VAE~\cite{van2017neural} to compress slats from $64^3 \times 8$ to a dense $8^3 \times 128$ latent, vector-quantized with an 8192-token codebook. Each object is thus represented by 512 discrete tokens—a $\sim\!40\times$ reduction—enabling efficient autoregressive decoding.

How do we train such a 3D VQ-VAE in latent slat space while preserving essential geometric and appearance information? We find that training the VQ-VAE in the latent space of slats is insufficient to learn an effective representation for reconstruction. To overcome this, we apply an auxiliary reconstruction loss to the decoded reconstructed slats in pixel-space, as shown in~\cref{fig:3d-vq-vae-training}. We use the same pixel-space reconstruction loss ($\mathcal{L}_1$, D-SSIM and LPIPS) as in~\cite{xiang2024structured}. This yields significantly better reconstructions, despite achieving a similar latent space reconstruction loss, as shown in \cref{fig:decoded-rec-loss-comparison}. Additionally, we find that imposing a multi-view reconstruction loss of the asset (randomly sampled from 150 views) leads to better reconstructions than a single fixed view. We quantify the improvement using human evaluations in \cref{table:decoded-rec-loss}.

Our Trellis-based 3D VQ-VAE matches or surpasses SAR3D~\cite{chen2024sar3d} while using 4× fewer tokens (512 vs.\ 2040). Qualitative and quantitive comparisons are shown in \cref{fig:3d-reconstruction-sar3d} and \cref{table:3d-reconstruction-sar3d}. Importantly, our compact 3D tokens integrate seamlessly into our unified vocabulary with image and language tokens, enabling efficient multi-object scene encoding and autoregressive decoding, while SAR3D tackles single-asset generation only.

While our representation enables effective 3D asset tokenization and reconstruction, we further test its suitability for autoregressive decoding. Using the same Llama-3.2-Instruct backbone, we evaluate reconstruction and rendering on unseen assets, in \cref{fig:asset-level}. The learned 3D encodings generalize well, 
confirming their effectiveness for both reconstruction and decoding.
These 3D tokens are then incorporated into our unified vocabulary alongside image and language tokens for training and inference.

\mypar{(2) How to tokenize 3D location and orientation?} 
In addition to 3D shape, the 3D location and orientation of an object
are key attributes within our 3D modality and critical for understanding 3D spatial relationships and performing grounded 3D actions based on instructions. Thus, accurately encoding coordinates is essential. However, LLMs are struggle with numbers~\cite{mirzadeh2024gsm, singh2024tokenization}. \setlength{\tabcolsep}{2pt}
% ------------- wrapped, compact table (booktabs + siunitx) -------------
\begin{table}[t]     % right, 40 % of text width
  \vspace{2mm}
  \centering
  \scriptsize
  \renewcommand{\arraystretch}{0.95}      % tighten rows a bit

  % scale the tabular so it can never exceed the wraptable box
  \resizebox{\columnwidth}{!}{%
  \begin{tabular}{@{}S[table-format=1.3]
    S[table-format=1.3]  % Rendering
    S[table-format=1.4]  % Recognition
    S[table-format=1.4]  % Instruction
    S[table-format=1.4]@{}} % QA
    \toprule
    \textbf{Granularity} &
      {Rendering$({\downarrow})$} &
      {Recognition$({\uparrow})$} &
      {Instruction$({\uparrow})$} &
      {QA$({\uparrow})$} \\ \midrule
    0.005 & 1.380 & 0.5707 & 0.8643 & $\mathbf{0.5185}$ \\
    0.05  & $\mathbf{1.200}$ & $\mathbf{0.9212}$ & $\mathbf{0.8666}$ & 0.4980 \\
    0.5   & 2.020 & 0.2352 & 0.2427 & 0.4730 \\
    \bottomrule
  \end{tabular}}%
  \vspace{-1mm}
  \caption{\small\textbf{Effect of granularity.}  
           A value of $0.05$ yields the best overall performance on CLEVR.}
  \label{table:number-granularity}
  \vspace{-6mm}
\end{table}
% -----------------------------------------------------------------------

\begin{figure}[t]
\vspace*{-3mm}
\centering
\includegraphics[width=\columnwidth]{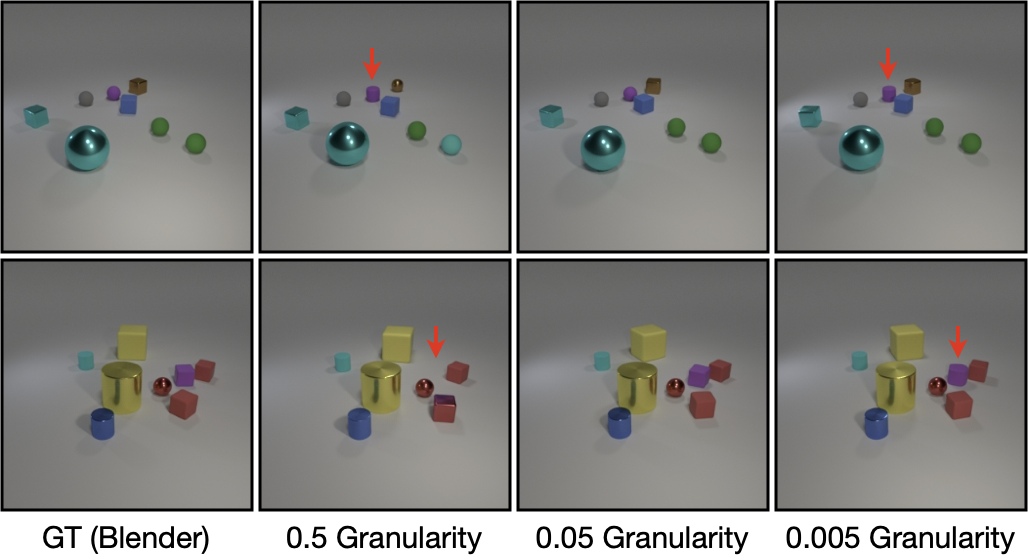}
\vspace{-6mm}
\caption{\small \textbf{Effect of Granularity.} 
A $0.05$ granularity more accurately captures object locations and shapes.}
\label{fig:granularity-rendering}
\vspace{-4mm}
\end{figure}

To mitigate this, we encode each object's $x,y,z$ coordinates separately as individual tokens, allowing the model to learn distinct embeddings for each coordinate. We discretize the coordinate values using equally spaced bins based on a chosen granularity. We find this granularity to be decisive for performance -- if it is too coarse, spatial inaccuracies arise; if it is too fine, it leads to an 
exponential increase in tokens, with fewer training samples per bin and thus difficulty in learning. \cref{table:number-granularity} shows the effect of granularity across the four tasks on CLEVR. We choose CLEVR, which features simple, known shapes, to isolate and study the effect of number encodings in next-token prediction frameworks. 
We find that a granularity of $0.05$ outperforms the coarser $0.5$ and the finer $0.005$. ~\cref{fig:granularity-rendering} compares image generation for the rendering task on the test set across varying levels of granularity.

Furthermore, our tokenization approach substantially improves efficiency by reducing sequence length compared to standard text tokenizers, which typically fragment floating-point values (e.g., "$0.000$" becomes "$0$", "$.$", "$000$"). We achieve a mean sequence length reduction from $271.4$ tokens using the standard Llama-3.2 tokenizer to $93.2$ tokens -- a $2.91\times$ compression ratio. These efficiency gains directly translate to decreased memory requirements and faster training and inference times.
The final vocabulary comprises $137,\!607$ tokens covering all modalities.

%%%%%%%%%%%%%
\subsubsection{What matters in input sequence design?}
\label{subsection:input-design}

We combine three modalities to construct the input sequence to the model. How should we order the modalities? And how should we encode numbers? We discuss our findings for input sequence design. 

\begin{figure}[t]
% \vspace*{-3mm}
% \begin{center}
\centering
\includegraphics[width=0.8\columnwidth, trim=2pt 2pt 2pt 2pt,clip]{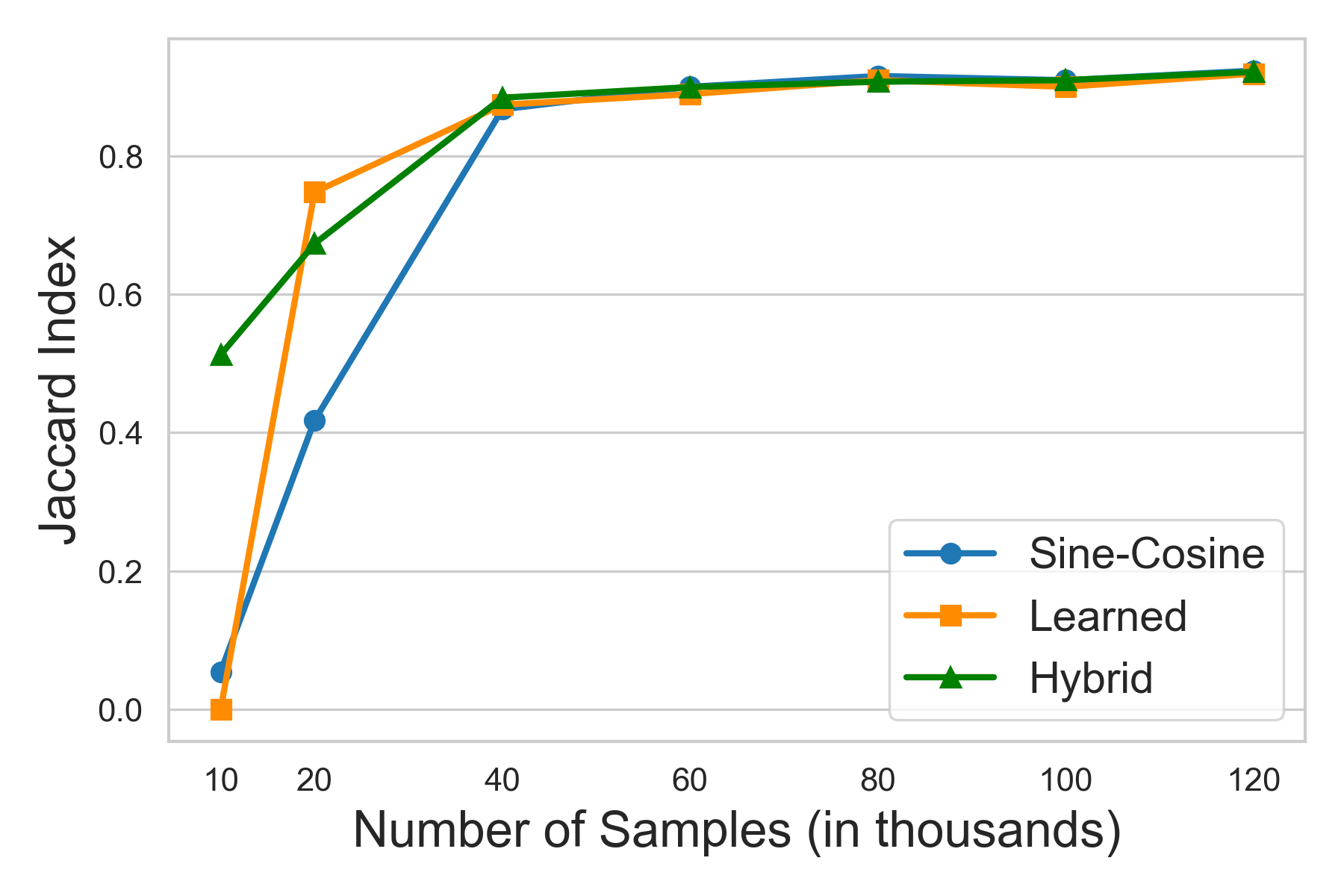}
% \end{center}
\vspace{-4mm}
\caption{\small Hybrid number encodings are robust across data scales. 
}
\label{fig:scaling-laws-number-encoding}
\vspace{-6mm}
\end{figure}

\mypar{Sine-cosine encoding of numbers.} 
While we independently tokenized coordinate values, naively learning embeddings for these tokens may fail to capture the inherent ordering of the numbers (\eg, $2$ is between $1$ and $3$).
We investigate whether augmenting the learned embeddings with sine-cosine encodings can enforce numeric ordering relationships.
Specifically, we evaluate three encoding strategies: (1) fixed sine-cosine encodings, (2) learned embeddings initialized from scratch, and (3) a hybrid approach where embeddings are learned but augmented with sine-cosine encodings.
\cref{fig:scaling-laws-number-encoding} plots the effect of these encoding strategies on the recognition task with varying training data sizes. 
While all methods perform on par in the high data regime, standalone fixed sine-cosine and learned embeddings collapse with low data. 
Consequently, we adopt the hybrid approach as it demonstrates robustness across data regimes. We show the performance of these encoding strategies across all four tasks in the Appendix.

\mypar{Should the image or 3D come first?} 
We investigate the impact of the ordering between the image and 3D modality for instruction-following and QA tasks. Our experiments reveal that placing the image before the 3D sequence leads to better performance compared to the reverse order. Specifically, we achieve an accuracy of $0.8666$ with the sequence $(I, \text{3D}, T_I) \rightarrow (I, \text{3D})$ compared to $0.8350$ with $(\text{3D}, I, T_I) \rightarrow (\text{3D}, I)$. Moreover, we obtain an accuracy of $0.4980$ with $(I, \text{3D}, T_Q) \rightarrow T_A$ compared to $0.4720$ with $(\text{3D}, I, T_Q) \rightarrow T_A$. This improvement could be attributed to the 3D tokens attending to the entirety of the preceding image tokens, enabling better conditioning and performance.

%%%%%%%%%%%%%
\subsubsection{What matters in output sequence design?}

We outline the important design decisions concerning the output sequence and objectives. 

\mypar{Initial token prediction matters.} 
The next-token prediction scheme caused challenges in generating reliable image outputs during inference. 
Despite achieving low training loss, the model’s decoded (next-token) predictions during inference deviated from expected outputs. At a high level, the issue stems from predicting rich outputs (images) from less informative conditions (3D specifications). 

\begin{figure*}[!t]  % or [!htbp] depending on template
  \centering

  \captionsetup[subfigure]{
    font=small,        % caption text
    labelfont=small,   % the “(a) (b) …” label
    skip=0pt          % vertical space image ↔︎ caption
}  
  
  % ------------ Left panel -------------
  \begin{subfigure}[b]{0.42\textwidth}
    \centering
    % top + bottom stacked inside this subfigure
    \includegraphics[width=0.7\linewidth]{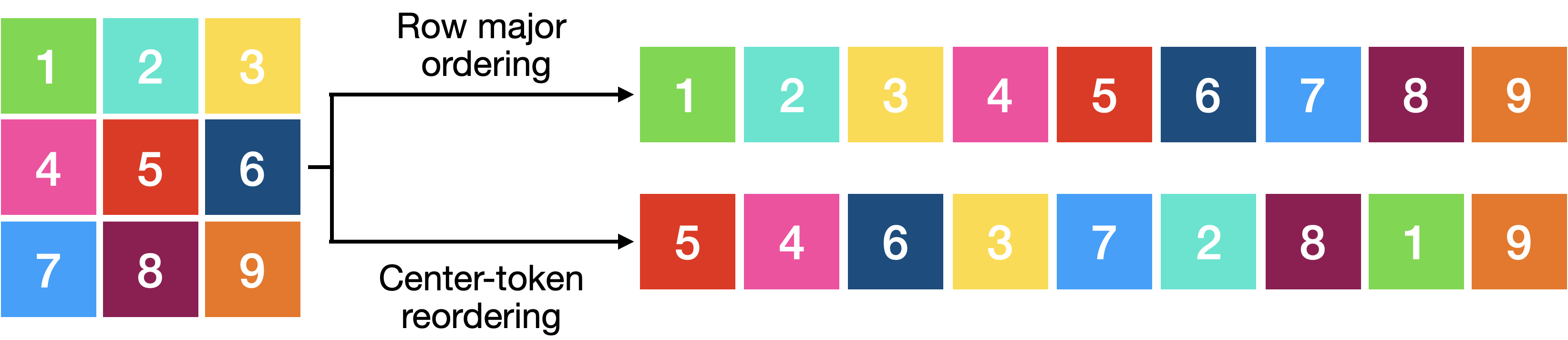}\\[-1mm]
    \includegraphics[width=0.8\linewidth]{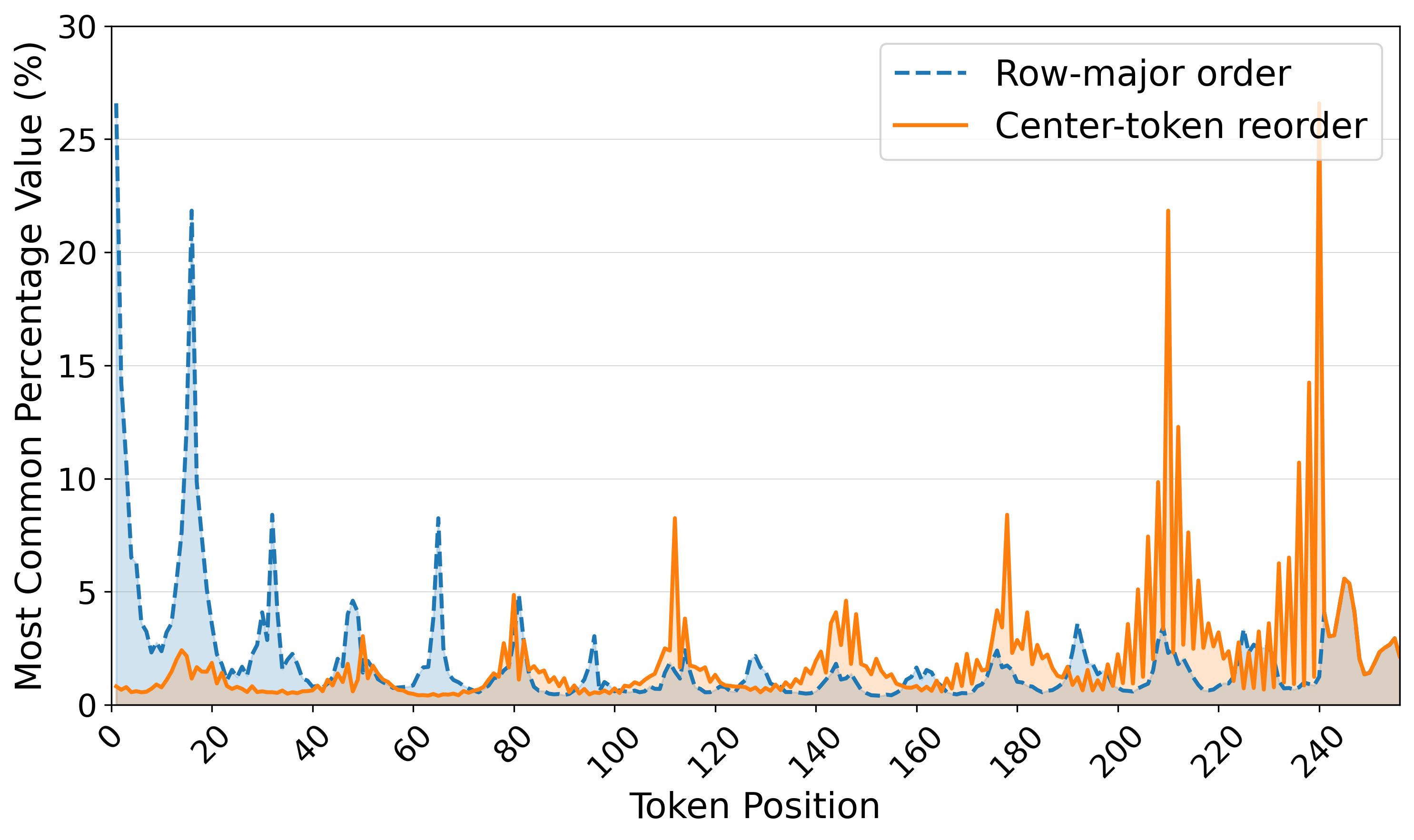}
    \caption{\small }
    \label{subfig:center-token-reordering}
  \end{subfigure}
  \hfill
  % ------------ Right panel -------------
  \begin{subfigure}[b]{0.57\textwidth}
    \centering
    \includegraphics[width=\linewidth]{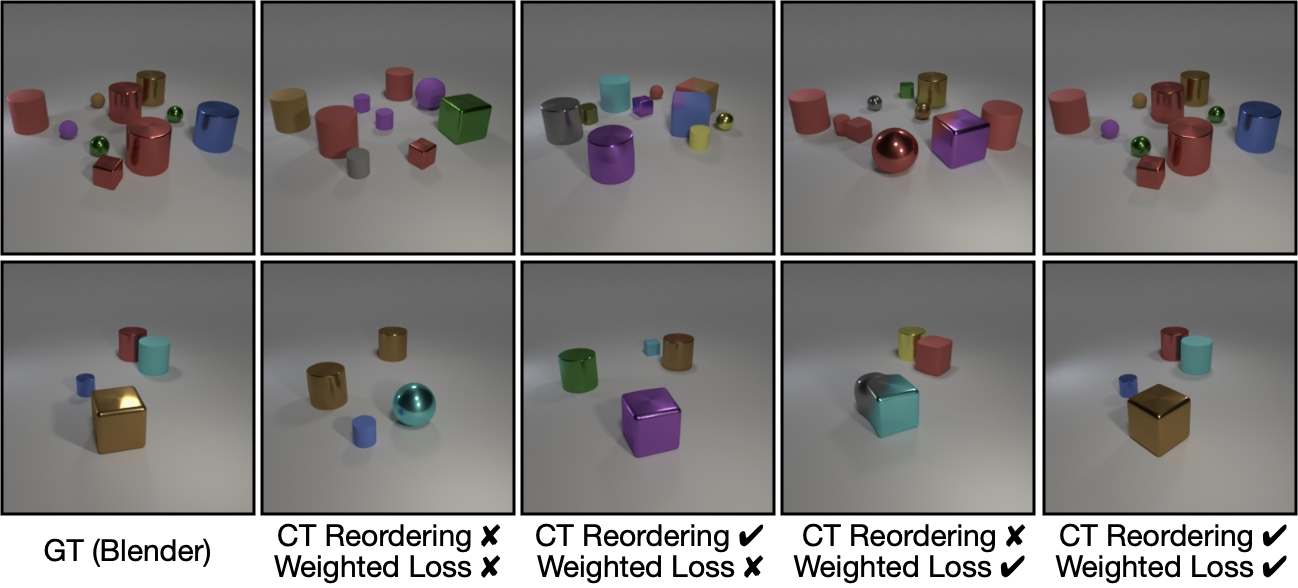}
    \caption{\small }
    \label{subfig:center-token-reordering-rendering}
  \end{subfigure}
  
  % ------------ Global caption ----------
  \vspace{-2mm}
  \caption{\small
    (a) \textbf{Top:} Schematic representation of center token reordering. \textbf{Bottom:} We convert CLEVR images into $256$-token sequences and analyze token frequency. Over $25\%$ of images share the same first token, causing biased predictions. Center token reordering significantly reduces this issue. (b) Inaccurate first-token predictions can cause catastrophic object- and scene-level diversions in autoregressive generations.
  }
  \label{fig:center-token-reordering-full}
  \vspace{-7mm}
\end{figure*}

Further investigation revealed the issue: the first token.
During inference, the model decodes sequentially, with the first token guiding the output.
For CLEVR images, this token, representing the top-left corner, was biased toward a few codes -- see Fig.~\ref{subfig:center-token-reordering} (blue dash plot) -- due to CLEVR's uniform gray background. This caused overfitting and during inference a wrongly predicted first token caused the decoding to diverge. This finding is not just applicable to CLEVR but to any image set with a more uniform background, often the case in graphic design. Moreover, this trend is even evident with real-world images which we show in the Appendix.

% ---------------- wrapped, compact table ----------------
\setlength{\tabcolsep}{3pt}
\begin{table}[t]
  \vspace{-4mm}
  \centering
  \scriptsize
  \renewcommand{\arraystretch}{0.95}

  % local settings—thousands comma if you ever need it, honour \mathbf
    % ticks / crosses (needs \usepackage{pifont} in the preamble)
    \newcommand{\cmark}{\ding{51}} % ✓
    \newcommand{\xmark}{\ding{55}} % ✗

    \begin{tabular}{@{}c c
      S[table-format=1.2]@{}}     % numeric column, aligned on decimal
      \toprule
      \textbf{CT reorder} & \textbf{Weighted loss} &
      {Rendering$({\downarrow})$} \\ \midrule
      \xmark & \xmark & 2.66 \\
      \cmark & \xmark & 3.56 \\
      \xmark & \cmark & 2.78 \\
      \cmark & \cmark & $\mathbf{1.00}$ \\
      \bottomrule
    \end{tabular}%
  \vspace{-2mm}
  \caption{\small Effect of center-token reordering and weighted loss.}
  \label{table:center-token-reordering-weighted-loss}
  \vspace{-6mm}
\end{table}
% --------------------------------------------------------

To mitigate this issue, we incorporate a \textit{center-token reordering} scheme to balance the token distribution at the sequence's starting position, by starting from the center token of the image and alternating hops after that as shown in Fig.~\ref{subfig:center-token-reordering}. This reordering means the first token now captures a representative part of the scene, instead of an uninformative background patch.
The token distribution and method and results are shown in Fig.~\ref{subfig:center-token-reordering} and Fig.~\ref{subfig:center-token-reordering-rendering} respectively.

\mypar{Token-specific loss weighting.}
We apply a \textit{weighted loss} during training by assigning a higher weight ($10.0$) to the loss for the first five tokens of the output image sequence. This enforces a stronger constraint to correctly predict the initial tokens, which proved critical for autoregressive decoding. 

~\cref{table:center-token-reordering-weighted-loss} shows the impact of center-token reordering and weighted loss for the rendering task. Best performance is achieved when both are combined, also shown in~\cref{subfig:center-token-reordering-rendering}.

\subsection{Generalization to complex scene layouts}
\label{subsection:generalization-to-complex-shapes}

Above, we validated our structured 3D modality and developed a cookbook for training unified autoregressive models with tokenized 3D shapes and attributes.
We verified our findings for 3D shape encodings on diverse Objaverse objects and for 3D layout attributes (e.g., locations) on CLEVR scenes with known shapes but diverse layouts.
We now show these findings generalize to ObjaWorld scenes with complex shapes (from Objaverse assets~\cite{deitke2023objaverse}) and diverse layouts.

We focus on two categories: park scenes (\textit{person, bench, lamppost, bird}) and living room scenes (\textit{person, sofa, coffee table}), featuring realistic textured objects with varied geometry at diverse locations and heights.
We generate $50,\!000$ scenes per category (total $100,\!000$) to train rendering and recognition models, and $2,\!000$ additional scenes with unseen object layouts for testing. The rendering and reconstruction tasks are defined the same as with CLEVR.

For recognition, \llm3 achieves a Jaccard Index of $0.6415$, well below its CLEVR score ($0.9212$, \cref{table:number-granularity}), reflecting ObjaWorld’s increased complexity. Llama3.2-V, prompted with in-context examples for the same attributes, performs near zero, failing to accurately predict 3D locations.

\begin{figure}[!t]  % 1-column figure
  \centering

  \captionsetup[subfigure]{
    font=small,        % caption text
    labelfont=small,   % the “(a) (b) …” label
    skip=-2pt          % vertical space image ↔︎ caption
  }

  % ------------ First row: (a) and (b) -------------
  \begin{subfigure}[b]{0.48\linewidth}
    \centering
    \includegraphics[width=\linewidth]{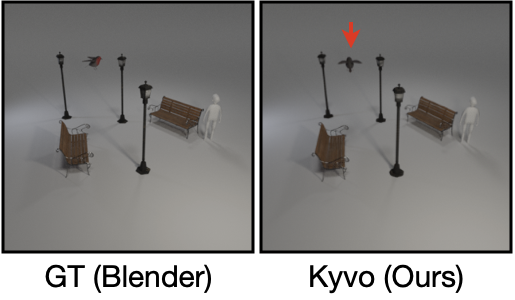}
    \caption{\small}
    \label{subfig:objaworld-rendering}
  \end{subfigure}
  \hfill
  \begin{subfigure}[b]{0.48\linewidth}
    \centering
    \includegraphics[width=\linewidth]{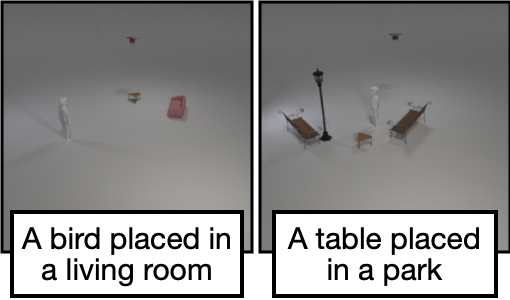}
    \vspace{-3mm}
    \caption{\small}
    \label{subfig:ood-generalization}
  \end{subfigure}

  \vspace{2mm}

  % ------------ Second row: (c) full width -------------
  \begin{subfigure}[b]{0.8\linewidth}
    \centering
    \includegraphics[width=\linewidth]{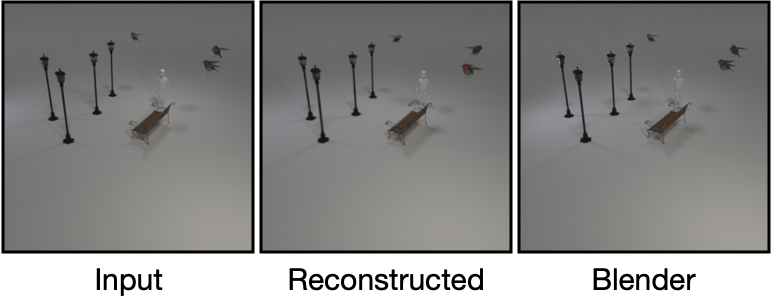}
    \vspace{-3mm}
    \caption{\small}
    \label{subfig:task-decomposition-objaworld}
  \end{subfigure}
  
  % ------------ Global caption ----------
  \vspace{-2mm}
  \caption{\small
    (a) \textbf{Rendering examples.}  
The model predicts images autoregressively from 3D inputs. Errors are largely pose mispredictions, \eg, the bird.  (b) \textbf{Novel scene configurations.} (c) \textbf{Chaining tasks.} Our recognition model predicts the 3D scene representation for an input image, which is then visualized through both our rendering model and Blender.
  }
  \label{fig:center-token-reordering-full}
  \vspace{-6mm}
\end{figure}

For rendering, the model renders capture accurate object types, counts, and positions, though some errors occur in fine pose alignment (Fig.~\ref{subfig:objaworld-rendering}).

\mypar{Generalizing to novel scene configurations.}  We test on out-of-distribution inputs (\eg, park-only objects placed in living room and vice-versa). \cref{subfig:ood-generalization} shows our model can generalize to such novel scenarios for the rendering task.

\mypar{Chaining tasks.} Chaining our recognition and rendering models enables image reconstruction to test model robustness. \cref{subfig:task-decomposition-objaworld} shows an example of input image, chained-model reconstruction, and Blender rendering of the predicted scene. Blender shows accurate object types, poses, and positions, and the reconstructed images also closely match the input, with only minor pose errors (\eg, the birds).

\subsection{Unified 3D shape and scene understanding}
\label{subsection:generalization-to-open-geometry}

\cref{subsection:generalization-to-complex-shapes} focused on the effectiveness of our structured 3D modality for diverse scene layouts with \textit{known} complex shapes. Therefore, we used type-specific object descriptions (\eg, ``bird'' or ``person''). 
Here, we bring together all our findings from~\cref{subsection:cookbook}, \textit{including} the learned 3D shape tokenization. We show that \llm3 can reconstruct and render complex 3D objects and scenes, using a variant of ObjaWorld with complex objects (\eg, barrel, chicken) from Objaverse~\cite{deitke2023objaverse}. We train recognition and rendering models on 100k image-scene pairs.

\begin{figure*}[!t]  % or [!htbp] depending on template
  \centering

  \captionsetup[subfigure]{
    font=small,        % caption text
    labelfont=small,   % the “(a) (b) …” label
    skip=0pt          % vertical space image ↔︎ caption
}   
  
  % ------------ Left panel -------------
  \begin{subfigure}[b]{0.77222\textwidth}
    \centering
    % top + bottom stacked inside this subfigure
    \includegraphics[width=\linewidth]{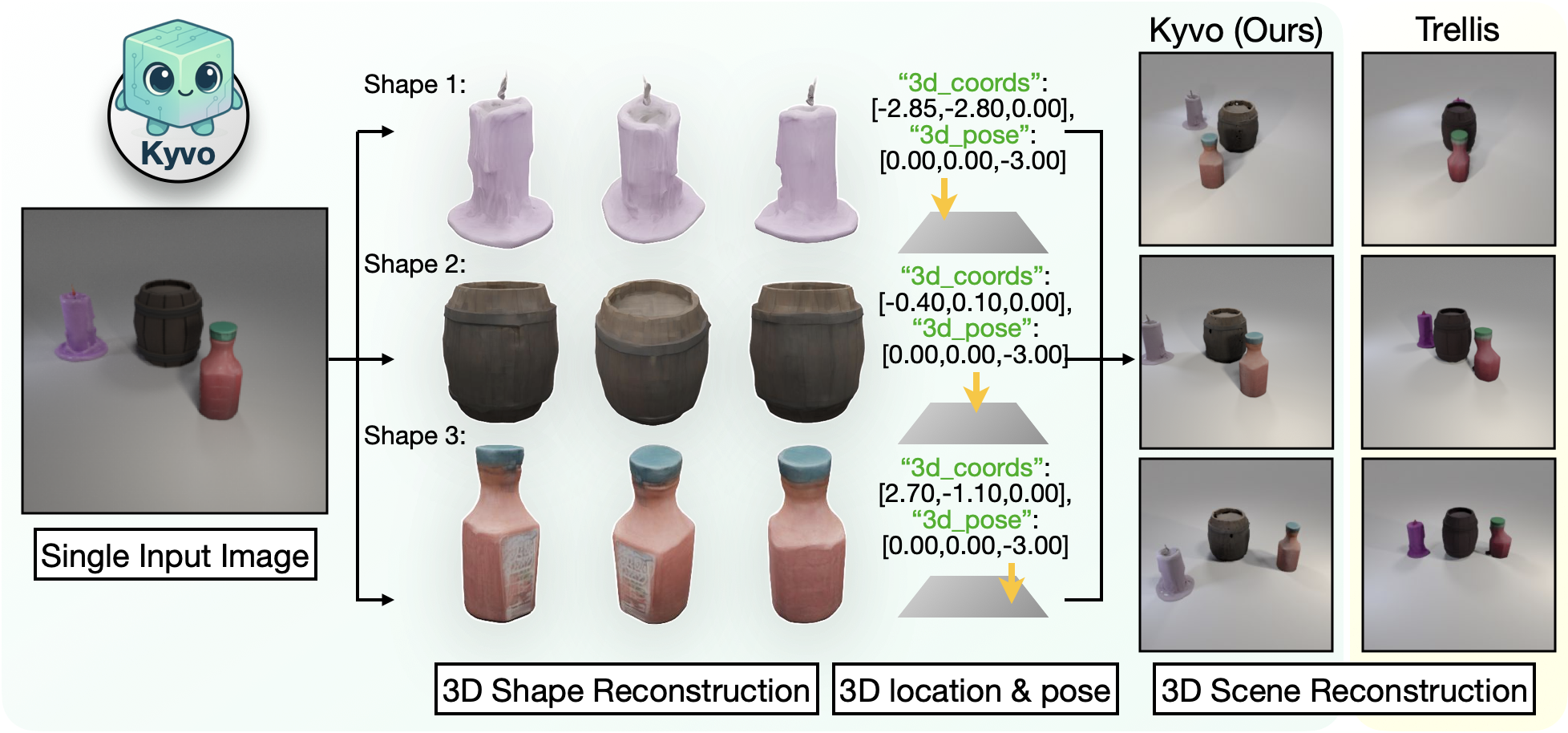}
    \caption{\small }
    \label{subfig:shape-and-scene-reconstruction}
  \end{subfigure}
  \hfill
  % ------------ Right panel -------------
  \begin{subfigure}[b]{0.21799\textwidth}
    \centering
    \includegraphics[width=\linewidth]{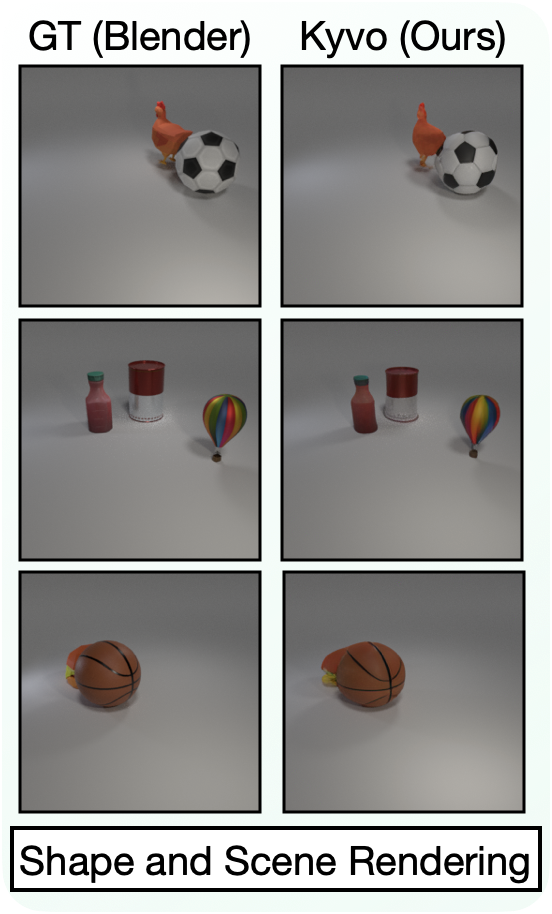}
    \caption{\small }
    \label{subfig:shape-and-scene-rendering}
  \end{subfigure}
  
  % ------------ Global caption ----------
  \vspace{-2mm}
  \caption{\small (a) \textbf{Unified shape and scene reconstruction example.} Given a single input image, Kyvo predicts shape sequences and reconstructs individual objects (candle, barrel, bottle) along with their 3D locations and poses via our structured 3D modality, effectively reconstructing the 3D scene with consistent spatial relations between the objects, visualized using Blender. (b) \textbf{Shape and scene rendering examples.} Given the structured 3D modality as input, Kyvo renders images with consistent object appearance and spatial relationships.
  }
  \label{fig:open-geometry-reconstruction}
  \vspace{-2mm}
\end{figure*}

The recognition model must (1) reconstruct full 3D geometry for each object, and (2) infer each object's 3D position and pose from a \emph{single image}.
Fig.~\ref{subfig:shape-and-scene-reconstruction} shows results on unseen scenes: our model recovers object geometries and spatial layouts via our structured, object-centric 3D modality, while Trellis~\cite{xiang2024structured}, a diffusion-based image-to-3D model trained on Objaverse that treats the scene holistically, often yields distorted shapes (\eg, a deformed bottle) and misaligned layouts (\eg, linearly arranged objects).

For rendering, the model takes the structured 3D modality with encoded shape tokens and outputs the corresponding image, mapping shape to appearance and placing objects at the correct positions and poses.
Fig.~\ref{subfig:shape-and-scene-rendering} compares our outputs with Blender renderings, showing that our model reliably captures shapes and spatial relationships, with only minor distortions in challenging cases (\eg, the occluded cheeseburger behind the basketball in the third image).

\begin{figure}[!thbp]
\centering
\includegraphics[width=\linewidth]{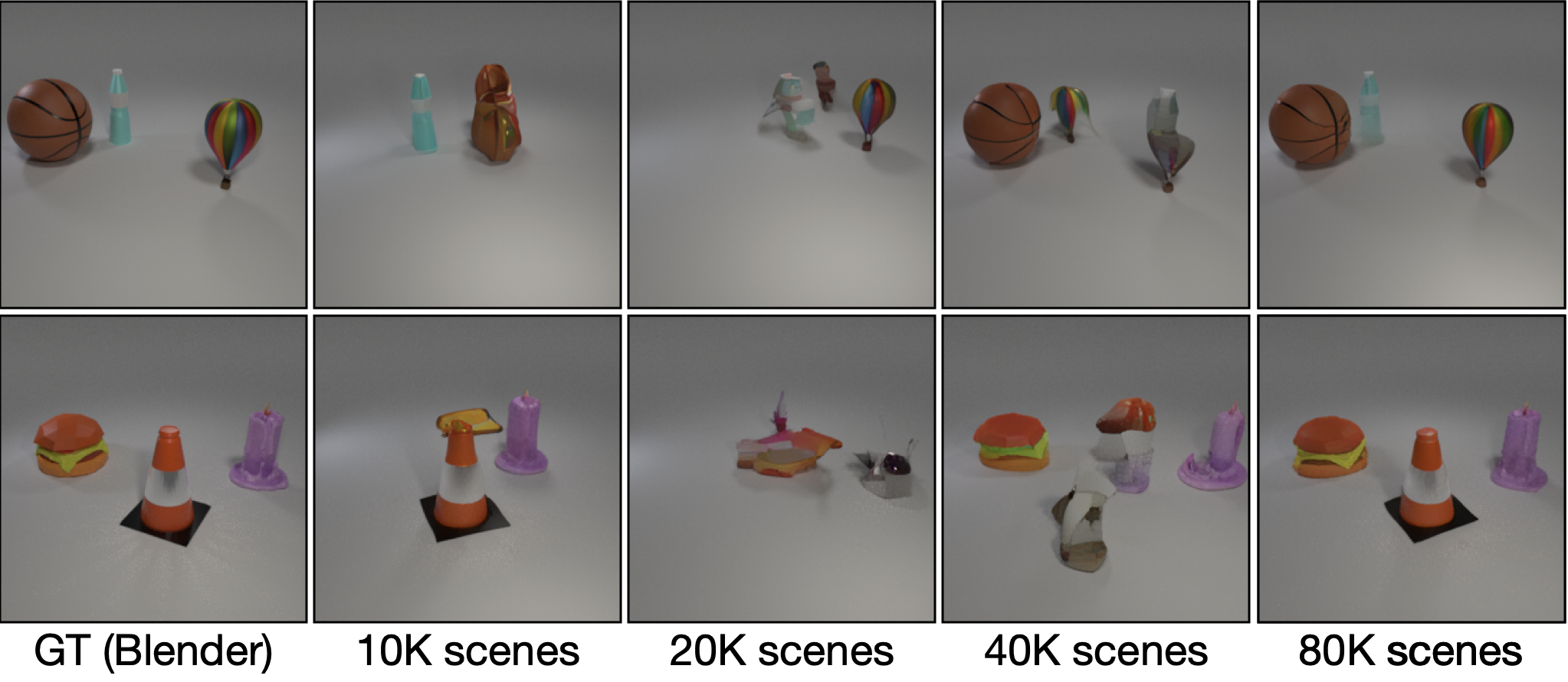}
\vspace{-6mm}
\caption{\small Qualitative scaling behavior of rendering model.}
\label{fig:scaling-laws-rendering}
\vspace{-6mm}
\end{figure}

\mypar{Qualitative scaling behavior.} 
Fig.~\ref{fig:scaling-laws-rendering} shows that with limited data the rendering model produces amorphous color blobs that lack semantic coherence and captures only coarse layouts, while increasing training data progressively improves object geometry, appearance, and spatial relationships, yielding more accurate and consistent renderings.

\lstset{
    basicstyle=\ttfamily\small,
    escapeinside={(*@}{@*)}, % escape delimiters for LaTeX in listings
    breaklines=true,
    breakindent=0pt,
    showstringspaces=false,
    backgroundcolor=\color{lightgray},
    extendedchars=false        % avoid non‑ASCII issues in listings
}

\subsection{Generalization to real-world recognition}
\label{subsection:generalization-to-real-world-recognition}

We now evaluate our model's effectiveness for 3D object recognition in real-world scenes.
We conduct experiments on two challenging real-world datasets: Objectron~\cite{ahmadyan2021objectron}, which features indoor and outdoor scenes of diverse object categories (\eg, bicycle, camera, car, cereal box); and ARKitScenes~\cite{dehghan2021arkitscenes}, featuring complex indoor environments of many object categories (\eg, bathtub, bed, cabinet, chair). ARKitScenes presents additional complexity due to its scene density and ground truth annotation noise. 
Following the train-test splits by~\cite{brazil2023omni3d}, we train a recognition model to detect objects by type and predict their 3D center coordinates and size dimensions in metric space. During training, we augment the data with random horizontal flipping. 
\begin{table*}[t]
  \centering
  \scriptsize
  \setlength{\tabcolsep}{2pt}    % tighter columns
  \renewcommand{\arraystretch}{0.95}

  % ----------------------------- (a) Objectron / ARKitScenes -----------------------------
  \begin{subtable}[t]{0.26\textwidth}
    \centering
    \resizebox{\textwidth}{!}{%
      \begin{tabular}{@{}l
                      S[table-format=1.4]
                      S[table-format=1.4]@{}}
        \toprule
        \textbf{Model} &
        {Objectron} &
        {ARKitScenes} \\ \midrule
        Cube R-CNN (ResNet-34) & 0.3276 & 0.2043 \\
        Cube R-CNN (DLA-34)    & 0.4012 & 0.2208 \\
        \llm3 (Ours)           & 0.4784 & 0.2118 \\
        \bottomrule
      \end{tabular}%
    }
    \caption{\small Jaccard Acc. vs Cube R-CNN.}
    \label{table:objectron-and-arkitscenes}
  \end{subtable}
  \hfill
  % ----------------------------- (b) Training recipes -----------------------------------
  \begin{subtable}[t]{0.33\textwidth}
    \centering
    \resizebox{\textwidth}{!}{%
      \begin{tabular}{@{}l
        S[table-format=1.2]   % Rendering
        S[table-format=1.4]   % Recognition
        S[table-format=1.4]   % Instruction
        S[table-format=1.4]@{}}% QA
        \toprule
        \textbf{Recipe} &
          {Rendering$({\downarrow})$} &
          {Recognition$({\uparrow})$} &
          {Instruction$({\uparrow})$} &
          {QA$({\uparrow})$} \\ \midrule
        Scratch & 1.36 & 0.6265 & 0.7744 & 0.4645 \\
        LoRA    & 1.82 & 0.8684 & $\mathbf{0.8680}$ & 0.3950 \\
        FFT     & $\mathbf{1.26}$ & $\mathbf{0.9212}$ & 0.8666 & $\mathbf{0.4980}$ \\
        \bottomrule
      \end{tabular}%
    }
    \caption{\small\textbf{Training recipe:} Scratch vs LoRA vs FFT.}
    \label{table:training-recipes}
  \end{subtable}
  \hfill
  % ----------------------------- (c) Backbone ablations ---------------------------------
  \begin{subtable}[t]{0.39\textwidth}
    \centering
    \resizebox{\textwidth}{!}{%
      \begin{tabular}{@{}l
        S[table-format=1.2]   % Rendering
        S[table-format=1.4]   % Recognition
        S[table-format=1.4]   % Instruction
        S[table-format=1.4]@{}}% QA
        \toprule
        \textbf{Backbone} &
          {Rendering$({\downarrow})$} &
          {Recognition$({\uparrow})$} &
          {Instruction$({\uparrow})$} &
          {QA$({\uparrow})$} \\ \midrule
        Llama-3.2-1B          & 1.38 & 0.8948 & 0.8674 & 0.4490 \\
        Llama-3.2-1B-Instruct & 1.28 & $\mathbf{0.9212}$ & 0.8666 & $\mathbf{0.4980}$ \\
        Llama-3.2-3B-Instruct & $\mathbf{1.18}$ & 0.8626 & $\mathbf{0.8763}$ & 0.2345 \\
        \bottomrule
      \end{tabular}%
    }
    \caption{\small\textbf{Effect of backbone} size and instruction-tuning.}
    \label{table:backbone-architectures}
  \end{subtable}

  \vspace{-2mm}
  \caption{\small Comparing real-world recognition performance; Quantifying the effect of training recipes and different backbones for \llm3.}
  \label{tab:three-tables-side-by-side}
  \vspace{-4mm}
\end{table*}

Table~\ref{table:objectron-and-arkitscenes} reports the Jaccard Index of our \llm3 and a state-of-the-art 3D object detector Cube R-CNN~\cite{brazil2023omni3d} -- we apply a $0.05$ confidence threshold to their predictions, as recommended by the authors. \llm3 significantly outperforms Cube R-CNN with two backbone variants on Objectron and performs on par on the more challenging ARKitScenes dataset. 
This result demonstrates the potential of a general autoregressive framework that aligns images with the 3D modality.

\subsection{Analysis and observations}

We explore the effects of training strategies and model backbones for Kyvo.

\mypar{Training recipes.} 
Table~\ref{table:training-recipes} compares three approaches for model adaptation: training from scratch, LoRA~\cite{hu2021lora}, and full fine-tuning (FFT). FFT from pre-trained language-only weights yields superior performance even when adapting to image and 3D modalities unseen during pre-training -- suggesting effective cross-modal transfer with limited domain-specific data. Notably, LoRA performs worse despite its established efficacy for text-only adaptation, indicating limitations when incorporating entirely new modalities.

\mypar{Instruction-tuned backbones and model sizes.}  
Instruction-tuned backbones match or outperform non-instruction-tuned ones across all tasks, as shown in Table~\ref{table:backbone-architectures}. Increasing model size from 1B to 3B provides no significant gains (even performing notably worse for question-answering), indicating that the 1B model sufficiently captures our dataset's complexity while avoiding overfitting.
\section{Conclusion \& Limitations}
\label{sec:conclusion}

We introduce \llm3, a autoregressive model that aligns structured 3D with language and vision to support a broad range of 3D tasks. Our empirical “cookbook”, based on training \textbf{307 models}, outlines effective design choices, including for tokenizing 3D scene attributes and complex 3D shapes. We will release code and data.

We cover limitations through scaling behavior, qualitative and quantitative results. 
A key challenge is the limited availability of 3D data. We show that strong performance and within-domain generalization can be achieved with relatively modest training data, but achieving cross-domain generalization demands larger datasets not readily available. 
A promising direction is extending \llm3 to handle mixed training data, enabling generalization to new domains even when 3D data is not always available as a paired modality.
\section*{Acknowledgments}
We thank Damiano Marsili, Raphi Kang, Ilona Demler, and Ziqi Ma for their valuable feedback.
Aadarsh is supported by the Kortschak Scholarship. Georgia is supported by the Powell Foundation, Meta through the LLM evaluation research grant, Google, and Amazon.

{
    \small
    \bibliographystyle{ieeenat_fullname}
    \bibliography{main}

@misc{lu2025atokenunifiedtokenizervision,
      title={AToken: A Unified Tokenizer for Vision}, 
      author={Jiasen Lu and Liangchen Song and Mingze Xu and Byeongjoo Ahn and Yanjun Wang and Chen Chen and Afshin Dehghan and Yinfei Yang},
      year={2025},
      eprint={2509.14476},
      archivePrefix={arXiv},
      primaryClass={cs.CV},
      url={https://arxiv.org/abs/2509.14476}, 
}

@inproceedings{chen2024sar3d,
    title={SAR3D: Autoregressive 3D Object Generation and Understanding via Multi-scale 3D VQVAE},
    author={Chen, Yongwei and Lan, Yushi and Zhou, Shangchen and Wang, Tengfei and Pan, Xingang},
    booktitle={CVPR},
    year={2025}
}

@inproceedings{brazil2023omni3d,
  title={Omni3d: A large benchmark and model for 3d object detection in the wild},
  author={Brazil, Garrick and Kumar, Abhinav and Straub, Julian and Ravi, Nikhila and Johnson, Justin and Gkioxari, Georgia},
  booktitle={Proceedings of the IEEE/CVF conference on computer vision and pattern recognition},
  pages={13154--13164},
  year={2023}
}

@inproceedings{ahmadyan2021objectron,
  title={Objectron: A large scale dataset of object-centric videos in the wild with pose annotations},
  author={Ahmadyan, Adel and Zhang, Liangkai and Ablavatski, Artsiom and Wei, Jianing and Grundmann, Matthias},
  booktitle={Proceedings of the IEEE/CVF conference on computer vision and pattern recognition},
  pages={7822--7831},
  year={2021}
}

@inproceedings{johnson2017clevr,
  title={Clevr: A diagnostic dataset for compositional language and elementary visual reasoning},
  author={Johnson, Justin and Hariharan, Bharath and Van Der Maaten, Laurens and Fei-Fei, Li and Lawrence Zitnick, C and Girshick, Ross},
  booktitle={Proceedings of the IEEE conference on computer vision and pattern recognition},
  pages={2901--2910},
  year={2017}
}

@inproceedings{deitke2023objaverse,
  title={Objaverse: A universe of annotated 3d objects},
  author={Deitke, Matt and Schwenk, Dustin and Salvador, Jordi and Weihs, Luca and Michel, Oscar and VanderBilt, Eli and Schmidt, Ludwig and Ehsani, Kiana and Kembhavi, Aniruddha and Farhadi, Ali},
  booktitle={Proceedings of the IEEE/CVF Conference on Computer Vision and Pattern Recognition},
  pages={13142--13153},
  year={2023}
}

@article{wang2004image,
  title={Image quality assessment: from error visibility to structural similarity},
  author={Wang, Zhou and Bovik, Alan C and Sheikh, Hamid R and Simoncelli, Eero P},
  journal={IEEE transactions on image processing},
  volume={13},
  number={4},
  pages={600--612},
  year={2004},
  publisher={IEEE}
}

@article{heusel2017gans,
  title={Gans trained by a two time-scale update rule converge to a local nash equilibrium},
  author={Heusel, Martin and Ramsauer, Hubert and Unterthiner, Thomas and Nessler, Bernhard and Hochreiter, Sepp},
  journal={Advances in neural information processing systems},
  volume={30},
  year={2017}
}

@article{van2017neural,
  title={Neural discrete representation learning},
  author={Van Den Oord, Aaron and Vinyals, Oriol and others},
  journal={Advances in neural information processing systems},
  volume={30},
  year={2017}
}

@inproceedings{esser2021taming,
  title={Taming transformers for high-resolution image synthesis},
  author={Esser, Patrick and Rombach, Robin and Ommer, Bjorn},
  booktitle={Proceedings of the IEEE/CVF conference on computer vision and pattern recognition},
  pages={12873--12883},
  year={2021}
}

@article{tong2024metamorph,
  title={MetaMorph: Multimodal Understanding and Generation via Instruction Tuning},
  author={Tong, Shengbang and Fan, David and Zhu, Jiachen and Xiong, Yunyang and Chen, Xinlei and Sinha, Koustuv and Rabbat, Michael and LeCun, Yann and Xie, Saining and Liu, Zhuang},
  journal={arXiv preprint arXiv:2412.14164},
  year={2024}
}

@article{singh2024tokenization,
  title={Tokenization counts: the impact of tokenization on arithmetic in frontier llms},
  author={Singh, Aaditya K and Strouse, DJ},
  journal={arXiv preprint arXiv:2402.14903},
  year={2024}
}

@article{mirzadeh2024gsm,
  title={Gsm-symbolic: Understanding the limitations of mathematical reasoning in large language models},
  author={Mirzadeh, Iman and Alizadeh, Keivan and Shahrokhi, Hooman and Tuzel, Oncel and Bengio, Samy and Farajtabar, Mehrdad},
  journal={arXiv preprint arXiv:2410.05229},
  year={2024}
}

@article{laurenccon2024matters,
  title={What matters when building vision-language models?},
  author={Lauren{\c{c}}on, Hugo and Tronchon, L{\'e}o and Cord, Matthieu and Sanh, Victor},
  journal={arXiv preprint arXiv:2405.02246},
  year={2024}
}

@article{zhou2024transfusion,
  title={Transfusion: Predict the next token and diffuse images with one multi-modal model},
  author={Zhou, Chunting and Yu, Lili and Babu, Arun and Tirumala, Kushal and Yasunaga, Michihiro and Shamis, Leonid and Kahn, Jacob and Ma, Xuezhe and Zettlemoyer, Luke and Levy, Omer},
  journal={arXiv preprint arXiv:2408.11039},
  year={2024}
}

@article{hu2021lora,
  title={Lora: Low-rank adaptation of large language models},
  author={Hu, Edward J and Shen, Yelong and Wallis, Phillip and Allen-Zhu, Zeyuan and Li, Yuanzhi and Wang, Shean and Wang, Lu and Chen, Weizhu},
  journal={arXiv preprint arXiv:2106.09685},
  year={2021}
}

@article{beyer2024paligemma,
  title={Paligemma: A versatile 3b vlm for transfer},
  author={Beyer, Lucas and Steiner, Andreas and Pinto, Andr{\'e} Susano and Kolesnikov, Alexander and Wang, Xiao and Salz, Daniel and Neumann, Maxim and Alabdulmohsin, Ibrahim and Tschannen, Michael and Bugliarello, Emanuele and others},
  journal={arXiv preprint arXiv:2407.07726},
  year={2024}
}

@misc{grattafiori2024llama3herdmodels,
      title={The Llama 3 Herd of Models}, 
      author={Aaron Grattafiori and Abhimanyu Dubey and Abhinav Jauhri and Abhinav Pandey and Abhishek Kadian and Ahmad Al-Dahle and Aiesha Letman and Akhil Mathur and Alan Schelten and Alex Vaughan and Amy Yang and Angela Fan and Anirudh Goyal and Anthony Hartshorn and (additional authors not shown)},
      year={2024},
      eprint={2407.21783},
      archivePrefix={arXiv},
      primaryClass={cs.AI},
      url={https://arxiv.org/abs/2407.21783}, 
}

@article{ma2024llms,
  title={When LLMs step into the 3D World: A Survey and Meta-Analysis of 3D Tasks via Multi-modal Large Language Models},
  author={Ma, Xianzheng and Bhalgat, Yash and Smart, Brandon and Chen, Shuai and Li, Xinghui and Ding, Jian and Gu, Jindong and Chen, Dave Zhenyu and Peng, Songyou and Bian, Jia-Wang and others},
  journal={arXiv preprint arXiv:2405.10255},
  year={2024}
}

@article{hong20233d,
  title={3d-llm: Injecting the 3d world into large language models},
  author={Hong, Yining and Zhen, Haoyu and Chen, Peihao and Zheng, Shuhong and Du, Yilun and Chen, Zhenfang and Gan, Chuang},
  journal={Advances in Neural Information Processing Systems},
  year={2023}
}

@article{cheng2024spatialrgpt,
  title={SpatialRGPT: Grounded Spatial Reasoning in Vision Language Model},
  author={Cheng, An-Chieh and Yin, Hongxu and Fu, Yang and Guo, Qiushan and Yang, Ruihan and Kautz, Jan and Wang, Xiaolong and Liu, Sifei},
  journal={arXiv preprint arXiv:2406.01584},
  year={2024}
}

@article{liao2024reasoning,
  title={Reasoning paths with reference objects elicit quantitative spatial reasoning in large vision-language models},
  author={Liao, Yuan-Hong and Mahmood, Rafid and Fidler, Sanja and Acuna, David},
  journal={arXiv preprint arXiv:2409.09788},
  year={2024}
}

@inproceedings{ma2024spatialpin,
  title={Spatialpin: Enhancing spatial reasoning capabilities of vision-language models through prompting and interacting 3d priors},
  author={Ma, Chenyang and Lu, Kai and Cheng, Ta-Ying and Trigoni, Niki and Markham, Andrew},
  booktitle={The Thirty-eighth Annual Conference on Neural Information Processing Systems},
  year={2024}
}

@inproceedings{xu2025pointllm,
  title={Pointllm: Empowering large language models to understand point clouds},
  author={Xu, Runsen and Wang, Xiaolong and Wang, Tai and Chen, Yilun and Pang, Jiangmiao and Lin, Dahua},
  booktitle={ECCV},
  year={2024},
}

@inproceedings{tang2024minigpt,
  title={Minigpt-3d: Efficiently aligning 3d point clouds with large language models using 2d priors},
  author={Tang, Yuan and Han, Xu and Li, Xianzhi and Yu, Qiao and Hao, Yixue and Hu, Long and Chen, Min},
  booktitle={Proceedings of the 32nd ACM International Conference on Multimedia},
  pages={6617--6626},
  year={2024}
}

@inproceedings{kerr2023lerf,
  title={Lerf: Language embedded radiance fields},
  author={Kerr, Justin and Kim, Chung Min and Goldberg, Ken and Kanazawa, Angjoo and Tancik, Matthew},
  booktitle={ICCV},
  year={2023}
}

@article{shen2023distilled,
  title={Distilled feature fields enable few-shot language-guided manipulation},
  author={Shen, William and Yang, Ge and Yu, Alan and Wong, Jansen and Kaelbling, Leslie Pack and Isola, Phillip},
  journal={arXiv preprint arXiv:2308.07931},
  year={2023}
}

@article{ma2024find,
  title={Find Any Part in 3D},
  author={Ma, Ziqi and Yue, Yisong and Gkioxari, Georgia},
  journal={arXiv preprint arXiv:2411.13550},
  year={2024}
}

@inproceedings{avetisyan2025scenescript,
  title={Scenescript: Reconstructing scenes with an autoregressive structured language model},
  author={Avetisyan, Armen and Xie, Christopher and Howard-Jenkins, Henry and Yang, Tsun-Yi and Aroudj, Samir and Patra, Suvam and Zhang, Fuyang and Frost, Duncan and Holland, Luke and Orme, Campbell and others},
  booktitle={ECCV},
  year={2024},
}

@inproceedings{radford2021learning,
  title={Learning transferable visual models from natural language supervision},
  author={Radford, Alec and Kim, Jong Wook and Hallacy, Chris and Ramesh, Aditya and Goh, Gabriel and Agarwal, Sandhini and Sastry, Girish and Askell, Amanda and Mishkin, Pamela and Clark, Jack and others},
  booktitle={International conference on machine learning},
  pages={8748--8763},
  year={2021},
  organization={PMLR}
}

@inproceedings{jia2021scaling,
  title={Scaling up visual and vision-language representation learning with noisy text supervision},
  author={Jia, Chao and Yang, Yinfei and Xia, Ye and Chen, Yi-Ting and Parekh, Zarana and Pham, Hieu and Le, Quoc and Sung, Yun-Hsuan and Li, Zhen and Duerig, Tom},
  booktitle={International conference on machine learning},
  pages={4904--4916},
  year={2021},
  organization={PMLR}
}

@article{team2024chameleon,
  title={Chameleon: Mixed-modal early-fusion foundation models},
  author={Team, Chameleon},
  journal={arXiv preprint arXiv:2405.09818},
  year={2024}
}

@article{alayrac2022flamingo,
  title={Flamingo: a visual language model for few-shot learning},
  author={Alayrac, Jean-Baptiste and Donahue, Jeff and Luc, Pauline and Miech, Antoine and Barr, Iain and Hasson, Yana and Lenc, Karel and Mensch, Arthur and Millican, Katherine and Reynolds, Malcolm and others},
  journal={Advances in neural information processing systems},
  volume={35},
  pages={23716--23736},
  year={2022}
}

@article{liu2024visual,
  title={Visual instruction tuning},
  author={Liu, Haotian and Li, Chunyuan and Wu, Qingyang and Lee, Yong Jae},
  journal={Advances in neural information processing systems},
  volume={36},
  year={2024}
}

@inproceedings{chen2024internvl,
  title={Internvl: Scaling up vision foundation models and aligning for generic visual-linguistic tasks},
  author={Chen, Zhe and Wu, Jiannan and Wang, Wenhai and Su, Weijie and Chen, Guo and Xing, Sen and Zhong, Muyan and Zhang, Qinglong and Zhu, Xizhou and Lu, Lewei and others},
  booktitle={Proceedings of the IEEE/CVF Conference on Computer Vision and Pattern Recognition},
  pages={24185--24198},
  year={2024}
}

@article{lu2024deepseek,
  title={Deepseek-vl: towards real-world vision-language understanding},
  author={Lu, Haoyu and Liu, Wen and Zhang, Bo and Wang, Bingxuan and Dong, Kai and Liu, Bo and Sun, Jingxiang and Ren, Tongzheng and Li, Zhuoshu and Yang, Hao and others},
  journal={arXiv preprint arXiv:2403.05525},
  year={2024}
}

@article{ma2024janusflow,
  title={Janusflow: Harmonizing autoregression and rectified flow for unified multimodal understanding and generation},
  author={Ma, Yiyang and Liu, Xingchao and Chen, Xiaokang and Liu, Wen and Wu, Chengyue and Wu, Zhiyu and Pan, Zizheng and Xie, Zhenda and Zhang, Haowei and Zhao, Liang and others},
  journal={arXiv preprint arXiv:2411.07975},
  year={2024}
}

@article{tong2024cambrian,
  title={Cambrian-1: A fully open, vision-centric exploration of multimodal llms},
  author={Tong, Shengbang and Brown, Ellis and Wu, Penghao and Woo, Sanghyun and Middepogu, Manoj and Akula, Sai Charitha and Yang, Jihan and Yang, Shusheng and Iyer, Adithya and Pan, Xichen and others},
  journal={arXiv preprint arXiv:2406.16860},
  year={2024}
}

@article{wang2024qwen2,
  title={Qwen2-vl: Enhancing vision-language model's perception of the world at any resolution},
  author={Wang, Peng and Bai, Shuai and Tan, Sinan and Wang, Shijie and Fan, Zhihao and Bai, Jinze and Chen, Keqin and Liu, Xuejing and Wang, Jialin and Ge, Wenbin and others},
  journal={arXiv preprint arXiv:2409.12191},
  year={2024}
}

@article{wu2024deepseek,
  title={DeepSeek-VL2: Mixture-of-Experts Vision-Language Models for Advanced Multimodal Understanding},
  author={Wu, Zhiyu and Chen, Xiaokang and Pan, Zizheng and Liu, Xingchao and Liu, Wen and Dai, Damai and Gao, Huazuo and Ma, Yiyang and Wu, Chengyue and Wang, Bingxuan and others},
  journal={arXiv preprint arXiv:2412.10302},
  year={2024}
}

@article{vaswani2017attention,
  title={Attention is all you need},
  author={Vaswani, A},
  journal={Advances in Neural Information Processing Systems},
  year={2017}
}

@article{achiam2023gpt,
  title={Gpt-4 technical report},
  author={Achiam, Josh and Adler, Steven and Agarwal, Sandhini and Ahmad, Lama and Akkaya, Ilge and Aleman, Florencia Leoni and Almeida, Diogo and Altenschmidt, Janko and Altman, Sam and Anadkat, Shyamal and others},
  journal={arXiv preprint arXiv:2303.08774},
  year={2023}
}

@article{tay2022ul2,
  title={Ul2: Unifying language learning paradigms},
  author={Tay, Yi and Dehghani, Mostafa and Tran, Vinh Q and Garcia, Xavier and Wei, Jason and Wang, Xuezhi and Chung, Hyung Won and Shakeri, Siamak and Bahri, Dara and Schuster, Tal and others},
  journal={arXiv preprint arXiv:2205.05131},
  year={2022}
}

@article{chung2024scaling,
  title={Scaling instruction-finetuned language models},
  author={Chung, Hyung Won and Hou, Le and Longpre, Shayne and Zoph, Barret and Tay, Yi and Fedus, William and Li, Yunxuan and Wang, Xuezhi and Dehghani, Mostafa and Brahma, Siddhartha and others},
  journal={Journal of Machine Learning Research},
  volume={25},
  number={70},
  pages={1--53},
  year={2024}
}

@article{team2023gemini,
  title={Gemini: a family of highly capable multimodal models},
  author={Team, Gemini and Anil, Rohan and Borgeaud, Sebastian and Alayrac, Jean-Baptiste and Yu, Jiahui and Soricut, Radu and Schalkwyk, Johan and Dai, Andrew M and Hauth, Anja and Millican, Katie and others},
  journal={arXiv preprint arXiv:2312.11805},
  year={2023}
}

@inproceedings{dehghan2021arkitscenes,
  title={{ARK}itScenes - A Diverse Real-World Dataset for 3D Indoor Scene Understanding Using Mobile {RGB}-D Data},
  author={Gilad Baruch and Zhuoyuan Chen and Afshin Dehghan and Tal Dimry and Yuri Feigin and Peter Fu and Thomas Gebauer and Brandon Joffe and Daniel Kurz and Arik Schwartz and Elad Shulman},
  booktitle={NeurIPS Datasets and Benchmarks Track (Round 1)},
  year={2021},
}

@inproceedings{marsili2025visual,
  title={Visual Agentic AI for Spatial Reasoning with a Dynamic API},
  author={Marsili, Damiano and Agrawal, Rohun and Yue, Yisong and Gkioxari, Georgia},
  booktitle={CVPR},
  year={2025}
}

@inproceedings{xiang2024structured,
    title   = {Structured 3D Latents for Scalable and Versatile 3D Generation},
    author  = {Xiang, Jianfeng and Lv, Zelong and Xu, Sicheng and Deng, Yu and Wang, Ruicheng and Zhang, Bowen and Chen, Dong and Tong, Xin and Yang, Jiaolong},
    booktitle = {CVPR},
    year    = {2025}
}
}

\clearpage
\appendix

% ---------- Full-width block (spans both columns) ----------
\begin{strip}
  \newcommand{\tocline}[3]{%  #1 = indent, #2 = text, #3 = page ref
  \noindent#1#2\dotfill\ #3\par\vspace{0.15em}%
}
\newcommand{\tocsection}[2]{\tocline{}{\textbf{#1}}{\pageref{#2}}}
\newcommand{\tocsubsec}[2]{\tocline{\hspace{1.8em}}{#1}{\pageref{#2}}}
% ---------------------------------------------------------------

{\Large\bfseries Appendix Contents}\par
\bigskip\hrule\bigskip

\tocsection{A\quad Dataset details}{sec:app:dataset-details}
  \tocsubsec{A.1\quad Data generation}{subsec:app:data-generation}
  \tocsubsec{A.2\quad 3D scene serialization}{subsec:app:3d-scene-serialization}
  \tocsubsec{A.3\quad Tokenization}{subsec:app:tokenization}
  \tocsubsec{A.4\quad Task sequence formation}{subsec:app:task-sequence-formation}

\tocsection{B\quad Additional qualitative examples}{sec:app:additional-qualitative-examples}

\tocsection{C\quad Evaluation}{sec:app:evaluation}
  \tocsubsec{C.1\quad 3D scenes}{subsec:app:3d-scenes}
  \tocsubsec{C.2\quad Images}{subsec:app:images}
  \tocsubsec{C.3\quad 3D assets}{subsec:app:assets}
  \tocsubsec{C.4\quad Text}{subsec:app:text}

\tocsection{D\quad Additional implementation details}{sec:app:additional-implementation-details}
  \tocsubsec{D.1\quad Image VQGAN architecture}{subsec:app:vqgan-architecture}
  \tocsubsec{D.2\quad 3D VQ-VAE training}{subsec:app:3d-vq-vae-training}
  \tocsubsec{D.3\quad Encoding of numbers}{subsec:app:encoding-of-numbers}
  \tocsubsec{D.4\quad Compute resources and time}{subsec:app:compute-resources-and-time}

\tocsection{E\quad Additional experiments and observations}{sec:app:additional-experiments-and-observations}

\tocsection{F\quad Failure cases}{sec:app:failure-cases}

\bigskip\hrule\bigskip

\end{strip}
% ------------------------------------------------------------

\section{Dataset details}
\label{sec:app:dataset-details}

In this section, we present a detailed overview of the four datasets used in our experiments. We discuss dataset creation, statistics, serialization, tokenization, and task formulation.

\subsection{Data generation}
\label{subsec:app:data-generation}

\mypar{CLEVR.} We generate CLEVR scenes using the dataset creation code from~\cite{johnson2017clevr} and render the corresponding images with Blender. Each scene outputs a JSON file describing the scene along with its rendered image.

\myparit{Rendering.} We generate $120,\!000$ unique CLEVR scenes as training data for rendering. The JSON files serve as input for the 3D scene representation, while the corresponding images are used to generate output sequences. Further details are provided in the following sections. For evaluation, we use a test set of $2,\!000$ image-JSON pairs.

\myparit{Recognition.} We use the same training data as in rendering but with reversed input-output roles. Here, images serve as input sequences, while JSON files generate the 3D scene output. For evaluation, we use a test set of $2,\!000$ JSON-image pairs.

\myparit{Instruction-following.} We consider four different types of instructions for 3D scene modification: (1) \textit{modifying the appearance of objects}, (2) \textit{adding new objects}, (3) \textit{removing objects}, and (4) \textit{moving an object to a desired location}. Using $16$ to $28$ text instruction templates, we generate $20,\!000$ input-output pairs per instruction type that we build on top of the initial $20,\!000$ CLEVR scenes. Specifically, we sample a CLEVR scene, apply an instruction template, and generate the corresponding modified scene.
Additionally, for instructions involving object appearance modification, we generate an extra $20,\!000$ pairs that do not reference other objects, ensuring modifications apply solely to uniquely identifiable objects within the scene. This results in a total of $100,\!000$ input-output pairs forming the training set.
For evaluation, we sample $500$ input-output pairs per instruction type, creating a test set of $2,\!500$ pairs. This approach enhances dataset diversity, improving the model’s ability to generalize across different instruction types. Example text instructions for each type are listed in Table~\ref{table:app-instruction-templates}.

\myparit{Question-answering.} For question-answer pair generation, we use the question generation engine by~\cite{johnson2017clevr} that uses functional programs and generate $20,\!000$ question-answer pairs for the training data. For evaluation, we use a test set of $2,\!000$ question-answer pairs.

\begin{table*}[htbp]
  \centering
  \scriptsize
  {%
    % local settings – commas and alignment only in this table
    \sisetup{group-separator={,}, group-minimum-digits=4}

    % scale to fit the text width (adjust the 0.9 if you like)
    \resizebox{\textwidth}{!}{%
    \begin{tabular}{@{}l
      S[table-format=5.0]      % “20,000”
      l@{}}
      \toprule
      \textbf{Instruction type} & \textbf{\# pairs} & \textbf{Example text instruction} \\
      \midrule
      \makecell[l]{Modifying the appearance of objects\\(no reference to other objects)}
        & 20000
        & \makecell[l]{\textit{``Change the gray object to have purple color''}\\
                       \textit{``Transform the small yellow rubber sphere to have metal material''}} \\[2pt]

      Modifying the appearance of objects
        & 20000
        & \makecell[l]{\textit{``Change the size of the small purple metal cylinder to the behind of large green rubber sphere to large''}\\
                       \textit{``Set the material of the gray metal cube object to the left of small purple rubber cylinder to rubber''}} \\[2pt]

      Adding new objects
        & 20000
        & \makecell[l]{\textit{``Put a small gray rubber cylinder to the left of small yellow rubber sphere''}\\
                       \textit{``Insert a red rubber cylinder object to the front of large cyan rubber cube''}} \\[2pt]

      Removing objects
        & 20000
        & \makecell[l]{\textit{``Remove the small red rubber cylinder to the right of large yellow rubber cube''}\\
                       \textit{``Take out the small rubber sphere object to the right of small gray rubber cylinder''}} \\[2pt]

      Moving an object to a desired location
        & 20000
        & \makecell[l]{\textit{``Move the small cyan rubber cylinder object to left and behind''}\\
                       \textit{``Change the position of the large rubber sphere object towards right and behind''}} \\
      \bottomrule
    \end{tabular}}%
  }
  \vspace{2mm}
  \caption{\small\textbf{Instruction templates.}  Each template type, its pair count, and two representative text instructions.}
  \label{table:app-instruction-templates}
  % \vspace{-3mm}
\end{table*}

\mypar{ObjaWorld.} We extend the CLEVR framework to support 3D Blender assets from Objaverse~\cite{deitke2023objaverse}, for inclusion of objects beyond basic geometric shapes such as cubes, cylinders, and spheres. Specifically, we adapt the CLEVR code to include objects like person, bird, bench, \etc. 

First, for experiments showing generalization to complex scene layouts (Section 3.3), we consider two scene setups: \textit{park} and \textit{living room}. \textit{Park} scenes are composed of the assets \textit{person, bird, bench}, and \textit{lamppost}, while \textit{living room} scenes use \textit{person, sofa}, and \textit{table} to construct the scenes. 
For training, we generate $50,\!000$ scenes for each setup, resulting in a total of $100,\!000$ scenes. 
Our test set comprises $4,\!000$ scenes, evenly split between $2,\!000$ park and $2,\!000$ living room scenes. 

For experiments showing unified 3D shape and scene understanding (Section 3.4), we select 20 complex Objaverse objects like barrel, chicken, cheeseburger, \etc, and generate 100,000 training scenes containing 2-3 randomly sampled objects. The test set contains 1,000 unseen scenes.

\mypar{Objectron.} We adhere to the official dataset splits by~\cite{brazil2023omni3d} to construct our training and test sets. For each object in a scene, we extract the category name, 3D center camera coordinates, and object dimensions from the annotation files, generating image–3D scene pairs.

\mypar{ARKitScenes.} Similar to Objectron, we follow~\cite{brazil2023omni3d} and use the provided dataset splits and extract object category labels, 3D center coordinates, and dimensions from annotations to generate image-3D scene pairs.

\mypar{Objaverse.} For training the 3D VQ-VAE, we use $\sim168$k Objaverse assets from the Sketchfab subset. However, the computational cost of extracting slats from assets is high using the original TRELLIS pipeline, as it requires extracting 150 renders of the assets which are then passed through the DINOv2 encoder. Instead, we extract a single render of each asset only, and pass it as image-conditioning to the pre-trained TRELLIS slat generator. We use the resultant slats as the inputs to the 3D VQ-VAE encoder during training. We found that the slats generated via this synthetic pipeline are relatively faithful to the original asset. Further, the synthetic slats are effective substitutes in the training data as during evaluation, we use slats extracted from the original TRELLIS pipeline and find that performance transfers well, indicating the distributions of slats are similar with the two methods.

\subsection{3D scene serialization}
\label{subsec:app:3d-scene-serialization}

As described above, each scene consists of an image paired with a structured 3D scene representation in JSON format. Before tokenization, we preprocess these JSON files by parsing and converting them into a single string representation. Specifically, we extract relevant attributes from the JSON and structure them using special markers like \texttt{[SHAPE]}, \texttt{[LOCATION]}, \etc, which vary depending on the dataset. These special markers are registered as special tokens in the tokenizer, which we discuss in the next section. For instance, in ObjaWorld, when we encode the explicit geometry, 512 tokens fetched from the 3D VQ-VAE codebook follow the \texttt{[SHAPE]} marker.
We provide examples of serialized outputs for each dataset below. 

\textit{CLEVR}:
\lstset{
    basicstyle=\ttfamily\small,
    escapeinside={(*@}{@*)}, % Define escape delimiters
    breaklines=true,
    showstringspaces=false,
    breakindent=0pt,
    backgroundcolor=\color{lightgray} % Set light gray background
}
\begin{lstlisting}
(*@\textcolor{greenkeyword}{[SCENE-START]}@*)(*@\textcolor{greenkeyword}{[OBJECT-START]}@*)(*@\textcolor{blueattribute}{[SIZE]}@*)large(*@\textcolor{blueattribute}{[COLOR]}@*)cyan(*@\textcolor{blueattribute}{[MATERIAL]}@*)metal(*@\textcolor{blueattribute}{[SHAPE]}@*)cube(*@\textcolor{blueattribute}{[LOCATION]}@*)-0.55 0.05 0.70(*@\textcolor{greenkeyword}{[OBJECT-END]}@*)(*@\textcolor{greenkeyword}{[OBJECT-START]}@*)(*@\textcolor{blueattribute}{[SIZE]}@*)small(*@\textcolor{blueattribute}{[COLOR]}@*)yellow(*@\textcolor{blueattribute}{[MATERIAL]}@*)metal(*@\textcolor{blueattribute}{[SHAPE]}@*)cylinder(*@\textcolor{blueattribute}{[LOCATION]}@*)1.25 2.50 0.35(*@\textcolor{greenkeyword}{[OBJECT-END]}@*)(*@\textcolor{greenkeyword}{[SCENE-END]}@*)
\end{lstlisting}

\textit{ObjaWorld}:
\lstset{
    basicstyle=\ttfamily\small,
    escapeinside={(*@}{@*)}, % Define escape delimiters
    breaklines=true,
    showstringspaces=false,
    breakindent=0pt,
    backgroundcolor=\color{lightgray} % Set light gray background
}
\begin{lstlisting}
(*@\textcolor{greenkeyword}{[SCENE-START]}@*)(*@\textcolor{greenkeyword}{[OBJECT-START]}@*)(*@\textcolor{blueattribute}{[SHAPE]}@*)table(*@\textcolor{blueattribute}{[LOCATION]}@*)-2.70 -2.20 0.20(*@\textcolor{blueattribute}{[POSE]}@*)0.00 0.00 -0.10(*@\textcolor{greenkeyword}{[OBJECT-END]}@*)(*@\textcolor{greenkeyword}{[OBJECT-START]}@*)(*@\textcolor{blueattribute}{[SHAPE]}@*)person(*@\textcolor{blueattribute}{[LOCATION]}@*)-0.20 -0.70 0.85(*@\textcolor{blueattribute}{[POSE]}@*)0.00 0.00 0.55(*@\textcolor{greenkeyword}{[OBJECT-END]}@*)(*@\textcolor{greenkeyword}{[OBJECT-START]}@*)(*@\textcolor{blueattribute}{[SHAPE]}@*)person(*@\textcolor{blueattribute}{[LOCATION]}@*)-0.75 -2.80 0.85(*@\textcolor{blueattribute}{[POSE]}@*)0.00 0.00 -2.55(*@\textcolor{greenkeyword}{[OBJECT-END]}@*)(*@\textcolor{greenkeyword}{[OBJECT-START]}@*)(*@\textcolor{blueattribute}{[SHAPE]}@*)table(*@\textcolor{blueattribute}{[LOCATION]}@*)2.75 1.90 0.20(*@\textcolor{blueattribute}{[POSE]}@*)0.00 0.00 1.95(*@\textcolor{greenkeyword}{[OBJECT-END]}@*)(*@\textcolor{greenkeyword}{[OBJECT-START]}@*)(*@\textcolor{blueattribute}{[SHAPE]}@*)sofa(*@\textcolor{blueattribute}{[LOCATION]}@*)0.40 2.75 0.30(*@\textcolor{blueattribute}{[POSE]}@*)0.00 0.00 -0.95(*@\textcolor{greenkeyword}{[OBJECT-END]}@*)(*@\textcolor{greenkeyword}{[SCENE-END]}@*)
\end{lstlisting}

\textit{ObjaWorld (with explicit shape representations)}:
% Color definitions
\definecolor{assetcolor}{RGB}{204,0,255} % Brighter purple for [ASSET]
\definecolor{vcolor}{RGB}{0,150,255}      % Bright blue for vector elementselements
% \definecolor{greenkeyword}{RGB}{0,158,115}
% \definecolor{blueattribute}{RGB}{0,114,178}

\begin{lstlisting}
(*@\textcolor{greenkeyword}{[SCENE-START]}@*)(*@\textcolor{greenkeyword}{[OBJECT-START]}@*)(*@\textcolor{assetcolor}{[SHAPE]}@*)<(*@\textcolor{vcolor}{$v_{1}^{1}$}@*),(*@\textcolor{vcolor}{$v_{2}^{1}$}@*),...,(*@\textcolor{vcolor}{$v_{512}^{1}$}@*)>(*@\textcolor{blueattribute}{[LOCATION]}@*)-2.45 0.60 1.20(*@\textcolor{blueattribute}{[POSE]}@*)0.00 0.00 2.30(*@\textcolor{greenkeyword}{[OBJECT-END]}@*)(*@\textcolor{greenkeyword}{[OBJECT-START]}@*)(*@\textcolor{assetcolor}{[SHAPE]}@*)<(*@\textcolor{vcolor}{$v_{1}^{2}$}@*),(*@\textcolor{vcolor}{$v_{2}^{2}$}@*),...,(*@\textcolor{vcolor}{$v_{512}^{2}$}@*)>(*@\textcolor{blueattribute}{[LOCATION]}@*)2.50 -1.35 0.00(*@\textcolor{blueattribute}{[POSE]}@*)0.00 0.00 1.55(*@\textcolor{greenkeyword}{[OBJECT-END]}@*)(*@\textcolor{greenkeyword}{[OBJECT-START]}@*)(*@\textcolor{assetcolor}{[SHAPE]}@*)<(*@\textcolor{vcolor}{$v_{1}^{3}$}@*),(*@\textcolor{vcolor}{$v_{2}^{3}$}@*),...,(*@\textcolor{vcolor}{$v_{512}^{3}$}@*)>(*@\textcolor{blueattribute}{[LOCATION]}@*)-1.80 -0.90 0.00(*@\textcolor{blueattribute}{[POSE]}@*)0.00 0.00 1.45(*@\textcolor{greenkeyword}{[OBJECT-END]}@*)(*@\textcolor{greenkeyword}{[SCENE-END]}@*)
\end{lstlisting}

\textit{Objectron}:
\lstset{
    basicstyle=\ttfamily\small,
    escapeinside={(*@}{@*)}, % Define escape delimiters
    breaklines=true,
    showstringspaces=false,
    breakindent=0pt,
    backgroundcolor=\color{lightgray} % Set light gray background
}
\begin{lstlisting}
(*@\textcolor{greenkeyword}{[SCENE-START]}@*)(*@\textcolor{greenkeyword}{[OBJECT-START]}@*)(*@\textcolor{blueattribute}{[CATEGORY]}@*)bicycle(*@\textcolor{blueattribute}{[CENTER\_CAM]}@*)0.00 -0.10 2.45(*@\textcolor{blueattribute}{[DIMENSIONS]}@*)0.60 1.10 1.00(*@\textcolor{greenkeyword}{[OBJECT-END]}@*)(*@\textcolor{greenkeyword}{[SCENE-END]}@*)
\end{lstlisting}

\textit{ARKitScenes}:
\lstset{
    basicstyle=\ttfamily\small,
    escapeinside={(*@}{@*)}, % Define escape delimiters
    breaklines=true,
    showstringspaces=false,
    breakindent=0pt,
    backgroundcolor=\color{lightgray} % Set light gray background
}
\begin{lstlisting}
(*@\textcolor{greenkeyword}{[SCENE-START]}@*)(*@\textcolor{greenkeyword}{[OBJECT-START]}@*)(*@\textcolor{blueattribute}{[CATEGORY]}@*)sofa(*@\textcolor{blueattribute}{[CENTER\_CAM]}@*)-0.14 0.04 1.50(*@\textcolor{blueattribute}{[DIMENSIONS]}@*)1.70 0.80 0.90(*@\textcolor{greenkeyword}{[OBJECT-END]}@*)(*@\textcolor{greenkeyword}{[OBJECT-START]}@*)(*@\textcolor{blueattribute}{[CATEGORY]}@*)table(*@\textcolor{blueattribute}{[CENTER\_CAM]}@*)0.02 0.08 1.60(*@\textcolor{blueattribute}{[DIMENSIONS]}@*)0.50 0.40 0.50(*@\textcolor{greenkeyword}{[OBJECT-END]}@*)(*@\textcolor{greenkeyword}{[OBJECT-START]}@*)(*@\textcolor{blueattribute}{[CATEGORY]}@*)table(*@\textcolor{blueattribute}{[CENTER\_CAM]}@*)-0.30 0.22 0.00(*@\textcolor{blueattribute}{[DIMENSIONS]}@*)1.30 0.80 0.40(*@\textcolor{greenkeyword}{[OBJECT-END]}@*)(*@\textcolor{greenkeyword}{[OBJECT-START]}@*)(*@\textcolor{blueattribute}{[CATEGORY]}@*)cabinet(*@\textcolor{blueattribute}{[CENTER\_CAM]}@*)0.46 -0.02 0.00(*@\textcolor{blueattribute}{[DIMENSIONS]}@*)0.70 0.80 0.30(*@\textcolor{greenkeyword}{[OBJECT-END]}@*)(*@\textcolor{greenkeyword}{[OBJECT-START]}@*)(*@\textcolor{blueattribute}{[CATEGORY]}@*)cabinet(*@\textcolor{blueattribute}{[CENTER\_CAM]}@*)0.40 0.24 0.00(*@\textcolor{blueattribute}{[DIMENSIONS]}@*)1.90 0.90 0.70(*@\textcolor{greenkeyword}{[OBJECT-END]}@*)(*@\textcolor{greenkeyword}{[SCENE-END]}@*)
\end{lstlisting}

\subsection{Tokenization}
\label{subsec:app:tokenization}

\mypar{Text.} We employ an off-the-shelf text tokenizer from Llama-3.2~\cite{grattafiori2024llama3herdmodels} with a vocabulary size of $128,\!000$ for tokenization. 

\mypar{Images.} For image tokenization, we train a domain-specific VQGAN on the training set of each dataset. This model encodes images into discrete representations by mapping them to codebook indices, which are then used for downstream processing.

\mypar{3D Scenes.} To enable tokenization of 3D scene representations, we augment the vocabulary of the Llama-3.2 tokenizer with two additional token types: (1) special tokens that serve as markers for scene attributes and (2) numerical tokens, ensuring that each location coordinate is encoded as a distinct token.

\mypar{3D Shapes.} For encoding the explicit shape geometries for the assets, we train a 3D VQ-VAE for tokenization as discussed in Section 3.2.1 of the main paper. We provide more details on training the 3D VQ-VAE in Section~\ref{subsec:app:3d-vq-vae-training} below.

% \aanote{Add for ObjaWorld, Objectron, ARKit}

\subsection{Task sequence formation}
\label{subsec:app:task-sequence-formation}

After tokenizing all three modalities, we construct sequences tailored to the specific task, which are then used to train \llm3. In this section, we present examples of complete sequences for each of the tasks across the four datasets.

\noindent\textbf{CLEVR:}

\textit{Rendering}: $\text{3D} \rightarrow \text{Image}$

\lstset{
    basicstyle=\ttfamily\small,
    escapeinside={(*@}{@*)}, % Define escape delimiters
    breaklines=true,
    showstringspaces=false,
    breakindent=0pt,
    backgroundcolor=\color{lightgray} % Set light gray background
}
\begin{lstlisting}
(*@\textcolor{gray}{[BOS]}@*)(*@\textcolor{greenkeyword}{[SCENE-START]}@*)(*@\textcolor{greenkeyword}{[OBJECT-START]}@*)(*@\textcolor{blueattribute}{[SIZE]}@*)small(*@\textcolor{blueattribute}{[COLOR]}@*)green(*@\textcolor{blueattribute}{[MATERIAL]}@*)metal(*@\textcolor{blueattribute}{[SHAPE]}@*)sphere(*@\textcolor{blueattribute}{[LOCATION]}@*)0.85 1.85 0.35(*@\textcolor{greenkeyword}{[OBJECT-END]}@*)(*@\textcolor{greenkeyword}{[OBJECT-START]}@*)(*@\textcolor{blueattribute}{[SIZE]}@*)small(*@\textcolor{blueattribute}{[COLOR]}@*)green(*@\textcolor{blueattribute}{[MATERIAL]}@*)metal(*@\textcolor{blueattribute}{[SHAPE]}@*)sphere(*@\textcolor{blueattribute}{[LOCATION]}@*)0.80 -2.00 0.35(*@\textcolor{greenkeyword}{[OBJECT-END]}@*)(*@\textcolor{greenkeyword}{[OBJECT-START]}@*)(*@\textcolor{blueattribute}{[SIZE]}@*)large(*@\textcolor{blueattribute}{[COLOR]}@*)brown(*@\textcolor{blueattribute}{[MATERIAL]}@*)metal(*@\textcolor{blueattribute}{[SHAPE]}@*)cylinder(*@\textcolor{blueattribute}{[LOCATION]}@*)-1.35 2.65 0.70(*@\textcolor{greenkeyword}{[OBJECT-END]}@*)(*@\textcolor{greenkeyword}{[OBJECT-START]}@*)(*@\textcolor{blueattribute}{[SIZE]}@*)small(*@\textcolor{blueattribute}{[COLOR]}@*)purple(*@\textcolor{blueattribute}{[MATERIAL]}@*)rubber(*@\textcolor{blueattribute}{[SHAPE]}@*)sphere(*@\textcolor{blueattribute}{[LOCATION]}@*)-0.90 -2.20 0.35(*@\textcolor{greenkeyword}{[OBJECT-END]}@*)(*@\textcolor{greenkeyword}{[OBJECT-START]}@*)(*@\textcolor{blueattribute}{[SIZE]}@*)large(*@\textcolor{blueattribute}{[COLOR]}@*)red(*@\textcolor{blueattribute}{[MATERIAL]}@*)rubber(*@\textcolor{blueattribute}{[SHAPE]}@*)cylinder(*@\textcolor{blueattribute}{[LOCATION]}@*)-2.70 -2.90 0.70(*@\textcolor{greenkeyword}{[OBJECT-END]}@*)(*@\textcolor{greenkeyword}{[OBJECT-START]}@*)(*@\textcolor{blueattribute}{[SIZE]}@*)large(*@\textcolor{blueattribute}{[COLOR]}@*)red(*@\textcolor{blueattribute}{[MATERIAL]}@*)metal(*@\textcolor{blueattribute}{[SHAPE]}@*)cylinder(*@\textcolor{blueattribute}{[LOCATION]}@*)2.25 -1.15 0.70(*@\textcolor{greenkeyword}{[OBJECT-END]}@*)(*@\textcolor{greenkeyword}{[OBJECT-START]}@*)(*@\textcolor{blueattribute}{[SIZE]}@*)small(*@\textcolor{blueattribute}{[COLOR]}@*)red(*@\textcolor{blueattribute}{[MATERIAL]}@*)metal(*@\textcolor{blueattribute}{[SHAPE]}@*)cube(*@\textcolor{blueattribute}{[LOCATION]}@*)2.15 -2.65 0.35(*@\textcolor{greenkeyword}{[OBJECT-END]}@*)(*@\textcolor{greenkeyword}{[SCENE-END]}@*)(*@\textcolor{pink}{[OUTPUT-SEP]}@*)
(*@\textcolor{greenkeyword}{[IMAGE-START]}@*)<image-tokens>(*@\textcolor{greenkeyword}{[IMAGE-END]}@*)(*@\textcolor{gray}{[EOS]}@*)

\end{lstlisting}

\textit{Recognition}: $\text{Image} \rightarrow \text{3D}$

\lstset{
    basicstyle=\ttfamily\small,
    escapeinside={(*@}{@*)}, % Define escape delimiters
    breaklines=true,
    showstringspaces=false,
    breakindent=0pt,
    backgroundcolor=\color{lightgray} % Set light gray background
}

\begin{lstlisting}
(*@\textcolor{gray}{[BOS]}@*)(*@\textcolor{greenkeyword}{[IMAGE-START]}@*)<image-tokens>(*@\textcolor{greenkeyword}{[IMAGE-END]}@*)(*@\textcolor{pink}{[OUTPUT-SEP]}@*)(*@\textcolor{greenkeyword}{[SCENE-START]}@*)
(*@\textcolor{greenkeyword}{[OBJECT-START]}@*)(*@\textcolor{blueattribute}{[SIZE]}@*)small(*@\textcolor{blueattribute}{[COLOR]}@*)brown(*@\textcolor{blueattribute}{[MATERIAL]}@*)rubber(*@\textcolor{blueattribute}{[SHAPE]}@*)sphere(*@\textcolor{blueattribute}{[LOCATION]}@*)-2.50 0.30 0.35(*@\textcolor{greenkeyword}{[OBJECT-END]}@*)(*@\textcolor{greenkeyword}{[OBJECT-START]}@*)(*@\textcolor{blueattribute}{[SIZE]}@*)large(*@\textcolor{blueattribute}{[COLOR]}@*)blue(*@\textcolor{blueattribute}{[MATERIAL]}@*)metal(*@\textcolor{blueattribute}{[SHAPE]}@*)cylinder(*@\textcolor{blueattribute}{[LOCATION]}@*)2.95 1.60 0.70(*@\textcolor{greenkeyword}{[OBJECT-END]}@*)(*@\textcolor{greenkeyword}{[OBJECT-START]}@*)(*@\textcolor{blueattribute}{[SIZE]}@*)large(*@\textcolor{blueattribute}{[COLOR]}@*)red(*@\textcolor{blueattribute}{[MATERIAL]}@*)metal(*@\textcolor{blueattribute}{[SHAPE]}@*)cylinder(*@\textcolor{blueattribute}{[LOCATION]}@*)-0.85 0.60 0.70(*@\textcolor{greenkeyword}{[OBJECT-END]}@*)(*@\textcolor{greenkeyword}{[OBJECT-START]}@*)(*@\textcolor{blueattribute}{[SIZE]}@*)small(*@\textcolor{blueattribute}{[COLOR]}@*)green(*@\textcolor{blueattribute}{[MATERIAL]}@*)metal(*@\textcolor{blueattribute}{[SHAPE]}@*)sphere(*@\textcolor{blueattribute}{[LOCATION]}@*)0.85 1.85 0.35(*@\textcolor{greenkeyword}{[OBJECT-END]}@*)(*@\textcolor{greenkeyword}{[SCENE-END]}@*)(*@\textcolor{gray}{[EOS]}@*)

\end{lstlisting}

\textit{Instruction-Following}: $(\text{Image}, \text{3D}, \text{Text}_\text{I}) \rightarrow (\text{Image}, \text{3D})$

\lstset{
    basicstyle=\ttfamily\small,
    escapeinside={(*@}{@*)}, % Define escape delimiters
    breaklines=true,
    showstringspaces=false,
    breakindent=0pt,
    backgroundcolor=\color{lightgray} % Set light gray background
}

\begin{lstlisting}
(*@\textcolor{gray}{[BOS]}@*)(*@\textcolor{greenkeyword}{[IMAGE-START]}@*)<image-tokens>(*@\textcolor{greenkeyword}{[IMAGE-END]}@*)(*@\textcolor{greenkeyword}{[SCENE-START]}@*)(*@\textcolor{greenkeyword}{[OBJECT-START]}@*)(*@\textcolor{blueattribute}{[SIZE]}@*)
small(*@\textcolor{blueattribute}{[COLOR]}@*)brown(*@\textcolor{blueattribute}{[MATERIAL]}@*)rubber(*@\textcolor{blueattribute}{[SHAPE]}@*)sphere(*@\textcolor{blueattribute}{[LOCATION]}@*)-2.50 0.30 0.35(*@\textcolor{greenkeyword}{[OBJECT-END]}@*)(*@\textcolor{greenkeyword}{[OBJECT-START]}@*)(*@\textcolor{blueattribute}{[SIZE]}@*)large(*@\textcolor{blueattribute}{[COLOR]}@*)blue(*@\textcolor{blueattribute}{[MATERIAL]}@*)metal(*@\textcolor{blueattribute}{[SHAPE]}@*)cylinder(*@\textcolor{blueattribute}{[LOCATION]}@*)2.95 1.60 0.70(*@\textcolor{greenkeyword}{[OBJECT-END]}@*)(*@\textcolor{greenkeyword}{[OBJECT-START]}@*)(*@\textcolor{blueattribute}{[SIZE]}@*)large(*@\textcolor{blueattribute}{[COLOR]}@*)red(*@\textcolor{blueattribute}{[MATERIAL]}@*)metal(*@\textcolor{blueattribute}{[SHAPE]}@*)cylinder(*@\textcolor{blueattribute}{[LOCATION]}@*)-0.85 0.60 0.70(*@\textcolor{greenkeyword}{[OBJECT-END]}@*)(*@\textcolor{greenkeyword}{[OBJECT-START]}@*)(*@\textcolor{blueattribute}{[SIZE]}@*)small(*@\textcolor{blueattribute}{[COLOR]}@*)green(*@\textcolor{blueattribute}{[MATERIAL]}@*)metal(*@\textcolor{blueattribute}{[SHAPE]}@*)sphere(*@\textcolor{blueattribute}{[LOCATION]}@*)0.85 1.85 0.35(*@\textcolor{greenkeyword}{[OBJECT-END]}@*)(*@\textcolor{greenkeyword}{[SCENE-END]}@*)(*@\textcolor{greenkeyword}{[TEXT-START]}@*)Change the brown object to have purple color(*@\textcolor{greenkeyword}{[TEXT-END]}@*)
(*@\textcolor{pink}{[OUTPUT-SEP]}@*)(*@\textcolor{greenkeyword}{[IMAGE-START]}@*)<image-tokens>(*@\textcolor{greenkeyword}{[IMAGE-END]}@*)(*@\textcolor{greenkeyword}{[SCENE-START]}@*)(*@\textcolor{greenkeyword}{[OBJECT-START]}@*)(*@\textcolor{blueattribute}{[SIZE]}@*)
small(*@\textcolor{blueattribute}{[COLOR]}@*)purple(*@\textcolor{blueattribute}{[MATERIAL]}@*)rubber(*@\textcolor{blueattribute}{[SHAPE]}@*)sphere(*@\textcolor{blueattribute}{[LOCATION]}@*)-2.50 0.30 0.35(*@\textcolor{greenkeyword}{[OBJECT-END]}@*)(*@\textcolor{greenkeyword}{[OBJECT-START]}@*)(*@\textcolor{blueattribute}{[SIZE]}@*)large(*@\textcolor{blueattribute}{[COLOR]}@*)blue(*@\textcolor{blueattribute}{[MATERIAL]}@*)metal(*@\textcolor{blueattribute}{[SHAPE]}@*)cylinder(*@\textcolor{blueattribute}{[LOCATION]}@*)2.95 1.60 0.70(*@\textcolor{greenkeyword}{[OBJECT-END]}@*)(*@\textcolor{greenkeyword}{[OBJECT-START]}@*)(*@\textcolor{blueattribute}{[SIZE]}@*)large(*@\textcolor{blueattribute}{[COLOR]}@*)red(*@\textcolor{blueattribute}{[MATERIAL]}@*)metal(*@\textcolor{blueattribute}{[SHAPE]}@*)cylinder(*@\textcolor{blueattribute}{[LOCATION]}@*)-0.85 0.60 0.70(*@\textcolor{greenkeyword}{[OBJECT-END]}@*)(*@\textcolor{greenkeyword}{[OBJECT-START]}@*)(*@\textcolor{blueattribute}{[SIZE]}@*)small(*@\textcolor{blueattribute}{[COLOR]}@*)green(*@\textcolor{blueattribute}{[MATERIAL]}@*)metal(*@\textcolor{blueattribute}{[SHAPE]}@*)sphere(*@\textcolor{blueattribute}{[LOCATION]}@*)0.85 1.85 0.35(*@\textcolor{greenkeyword}{[OBJECT-END]}@*)(*@\textcolor{greenkeyword}{[SCENE-END]}@*)(*@\textcolor{gray}{[EOS]}@*)

\end{lstlisting}

\textit{Question-Answering}: $(\text{Image}, \text{3D}, \text{Text}_\text{Q}) \rightarrow \text{Text}_\text{A}$

\lstset{
    basicstyle=\ttfamily\small,
    escapeinside={(*@}{@*)}, % Define escape delimiters
    breaklines=true,
    showstringspaces=false,
    breakindent=0pt,
    backgroundcolor=\color{lightgray} % Set light gray background
}

\begin{lstlisting}
(*@\textcolor{gray}{[BOS]}@*)(*@\textcolor{greenkeyword}{[IMAGE-START]}@*)<image-tokens>(*@\textcolor{greenkeyword}{[IMAGE-END]}@*)(*@\textcolor{greenkeyword}{[SCENE-START]}@*)(*@\textcolor{greenkeyword}{[OBJECT-START]}@*)(*@\textcolor{blueattribute}{[SIZE]}@*)
small(*@\textcolor{blueattribute}{[COLOR]}@*)brown(*@\textcolor{blueattribute}{[MATERIAL]}@*)rubber(*@\textcolor{blueattribute}{[SHAPE]}@*)sphere(*@\textcolor{blueattribute}{[LOCATION]}@*)-2.50 0.30 0.35(*@\textcolor{greenkeyword}{[OBJECT-END]}@*)(*@\textcolor{greenkeyword}{[OBJECT-START]}@*)(*@\textcolor{blueattribute}{[SIZE]}@*)large(*@\textcolor{blueattribute}{[COLOR]}@*)blue(*@\textcolor{blueattribute}{[MATERIAL]}@*)metal(*@\textcolor{blueattribute}{[SHAPE]}@*)cylinder(*@\textcolor{blueattribute}{[LOCATION]}@*)2.95 1.60 0.70(*@\textcolor{greenkeyword}{[OBJECT-END]}@*)(*@\textcolor{greenkeyword}{[OBJECT-START]}@*)(*@\textcolor{blueattribute}{[SIZE]}@*)large(*@\textcolor{blueattribute}{[COLOR]}@*)red(*@\textcolor{blueattribute}{[MATERIAL]}@*)metal(*@\textcolor{blueattribute}{[SHAPE]}@*)cylinder(*@\textcolor{blueattribute}{[LOCATION]}@*)-0.85 0.60 0.70(*@\textcolor{greenkeyword}{[OBJECT-END]}@*)(*@\textcolor{greenkeyword}{[OBJECT-START]}@*)(*@\textcolor{blueattribute}{[SIZE]}@*)small(*@\textcolor{blueattribute}{[COLOR]}@*)green(*@\textcolor{blueattribute}{[MATERIAL]}@*)metal(*@\textcolor{blueattribute}{[SHAPE]}@*)sphere(*@\textcolor{blueattribute}{[LOCATION]}@*)0.85 1.85 0.35(*@\textcolor{greenkeyword}{[OBJECT-END]}@*)(*@\textcolor{greenkeyword}{[SCENE-END]}@*)(*@\textcolor{greenkeyword}{[TEXT-START]}@*)What size is the rubber sphere?(*@\textcolor{greenkeyword}{[TEXT-END]}@*)(*@\textcolor{pink}{[OUTPUT-SEP]}@*)(*@\textcolor{greenkeyword}{[TEXT-START]}@*)small(*@\textcolor{greenkeyword}{[TEXT-END]}@*)(*@\textcolor{gray}{[EOS]}@*)

\end{lstlisting}

Similar structure is followed for \textbf{ObjaWorld} (\textit{Rendering} and \textit{Recognition}), \textbf{Objectron} (\textit{Recognition}), and \textbf{ARKitScenes} (\textit{Recognition}). For \textbf{ObjaWorld}, when we encode the explicit shape representations, the complete sequences look like the following.

\noindent\textbf{ObjaWorld:} (with explicit shape representations) 
% \aanote{TODO}

\textit{Rendering}: $\text{3D} \rightarrow \text{Image}$

% Color definitions
\definecolor{assetcolor}{RGB}{204,0,255} % Brighter purple for [ASSET]
\definecolor{vcolor}{RGB}{0,150,255}      % Bright blue for vector elementselements
% \definecolor{greenkeyword}{RGB}{0,158,115}
% \definecolor{blueattribute}{RGB}{0,114,178}

\begin{lstlisting}
(*@\textcolor{gray}{[BOS]}@*)(*@\textcolor{greenkeyword}{[SCENE-START]}@*)(*@\textcolor{greenkeyword}{[OBJECT-START]}@*)(*@\textcolor{assetcolor}{[SHAPE]}@*)<(*@\textcolor{vcolor}{$v_{1}^{1}$}@*),(*@\textcolor{vcolor}{$v_{2}^{1}$}@*),...,(*@\textcolor{vcolor}{$v_{512}^{1}$}@*)>(*@\textcolor{blueattribute}{[LOCATION]}@*)2.10 0.15 -0.75(*@\textcolor{blueattribute}{[POSE]}@*)0.00 0.00 -2.10(*@\textcolor{greenkeyword}{[OBJECT-END]}@*)(*@\textcolor{greenkeyword}{[OBJECT-START]}@*)(*@\textcolor{assetcolor}{[SHAPE]}@*)<(*@\textcolor{vcolor}{$v_{1}^{2}$}@*),(*@\textcolor{vcolor}{$v_{2}^{2}$}@*),...,(*@\textcolor{vcolor}{$v_{512}^{2}$}@*)>(*@\textcolor{blueattribute}{[LOCATION]}@*)-1.25 2.85 0.05(*@\textcolor{blueattribute}{[POSE]}@*)0.00 0.00 1.85(*@\textcolor{greenkeyword}{[OBJECT-END]}@*)(*@\textcolor{greenkeyword}{[OBJECT-START]}@*)(*@\textcolor{assetcolor}{[SHAPE]}@*)<(*@\textcolor{vcolor}{$v_{1}^{3}$}@*),(*@\textcolor{vcolor}{$v_{2}^{3}$}@*),...,(*@\textcolor{vcolor}{$v_{512}^{3}$}@*)>(*@\textcolor{blueattribute}{[LOCATION]}@*)0.45 -2.35 -1.50(*@\textcolor{blueattribute}{[POSE]}@*)0.00 0.00 0.60(*@\textcolor{greenkeyword}{[OBJECT-END]}@*)(*@\textcolor{greenkeyword}{[SCENE-END]}@*)(*@\textcolor{pink}{[OUTPUT-SEP]}@*)
(*@\textcolor{greenkeyword}{[IMAGE-START]}@*)<image-tokens>(*@\textcolor{greenkeyword}{[IMAGE-END]}@*)(*@\textcolor{gray}{[EOS]}@*)
\end{lstlisting}

\textit{Recognition}: $\text{Image} \rightarrow \text{3D}$

\begin{lstlisting}
(*@\textcolor{gray}{[BOS]}@*)(*@\textcolor{greenkeyword}{[IMAGE-START]}@*)<image-tokens>(*@\textcolor{greenkeyword}{[IMAGE-END]}@*)(*@\textcolor{pink}{[OUTPUT-SEP]}@*)(*@\textcolor{greenkeyword}{[SCENE-START]}@*)
(*@\textcolor{greenkeyword}{[OBJECT-START]}@*)(*@\textcolor{assetcolor}{[SHAPE]}@*)<(*@\textcolor{vcolor}{$v_{1}^{1}$}@*),(*@\textcolor{vcolor}{$v_{2}^{1}$}@*),...,(*@\textcolor{vcolor}{$v_{512}^{1}$}@*)>(*@\textcolor{blueattribute}{[LOCATION]}@*)1.15 -2.85 0.40(*@\textcolor{blueattribute}{[POSE]}@*)0.00 0.00 -1.75(*@\textcolor{greenkeyword}{[OBJECT-END]}@*)(*@\textcolor{greenkeyword}{[OBJECT-START]}@*)(*@\textcolor{assetcolor}{[SHAPE]}@*)<(*@\textcolor{vcolor}{$v_{1}^{2}$}@*),(*@\textcolor{vcolor}{$v_{2}^{2}$}@*),...,(*@\textcolor{vcolor}{$v_{512}^{2}$}@*)>(*@\textcolor{blueattribute}{[LOCATION]}@*)-0.35 2.10 -0.50(*@\textcolor{blueattribute}{[POSE]}@*)0.00 0.00 2.45(*@\textcolor{greenkeyword}{[OBJECT-END]}@*)(*@\textcolor{greenkeyword}{[OBJECT-START]}@*)(*@\textcolor{assetcolor}{[SHAPE]}@*)<(*@\textcolor{vcolor}{$v_{1}^{3}$}@*),(*@\textcolor{vcolor}{$v_{2}^{3}$}@*),...,(*@\textcolor{vcolor}{$v_{512}^{3}$}@*)>(*@\textcolor{blueattribute}{[LOCATION]}@*)2.75 0.15 -1.60(*@\textcolor{blueattribute}{[POSE]}@*)0.00 0.00 -0.25(*@\textcolor{greenkeyword}{[OBJECT-END]}@*)(*@\textcolor{greenkeyword}{[SCENE-END]}@*)(*@\textcolor{gray}{[EOS]}@*)
\end{lstlisting}

\section{Additional qualitative examples}
\label{sec:app:additional-qualitative-examples}

In this section, we provide additional qualitative examples. 

\begin{figure*}[!htbp]
\begin{center}
\includegraphics[width=0.8\linewidth]{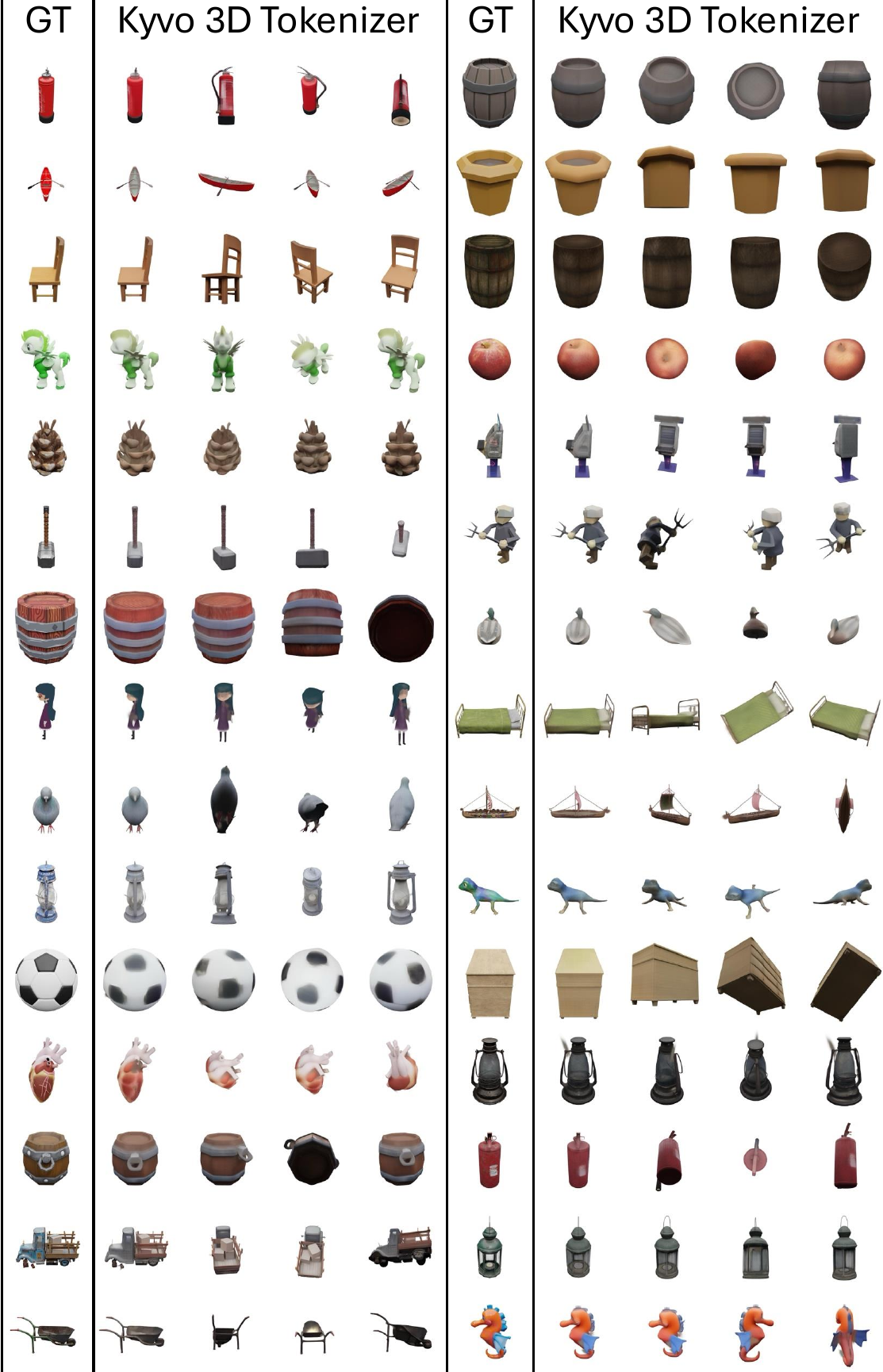} 
\end{center}
% \vspace{-3mm}
\caption{\small \textbf{Reconstruction examples for Trellis-based 3D VQ-VAE.} VQ-VAE reconstructions, shown from multiple views, compared to ground truth for unseen Objaverse assets.}
\label{fig:3d-tokenization-examples}
% \vspace{-3mm}
\end{figure*}

\begin{figure*}[!htbp]
    \centering
    \begin{minipage}[b]{0.64\textwidth}
        \centering
        \includegraphics[width=\linewidth]{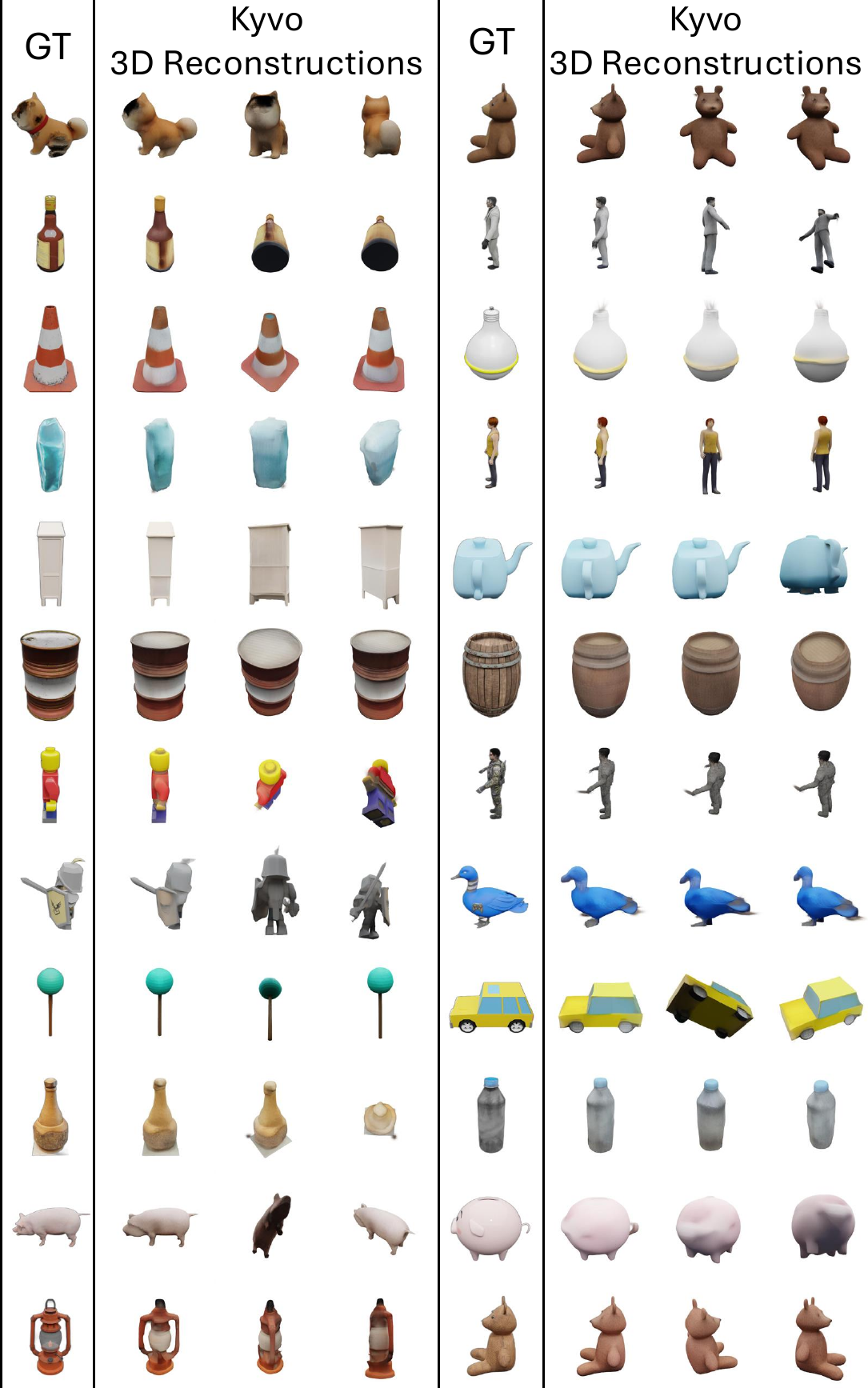}\\[0.5ex]
        \small (a)
    \end{minipage}%
    \hfill
    % vertical separator
    \begin{minipage}[b]{0.01\textwidth}
        \centering
        \rule{1.5pt}{18cm} % width, height of the vertical line
    \end{minipage}%
    \hfill
    \begin{minipage}[b]{0.33\textwidth}
        \centering
        \includegraphics[width=\linewidth]{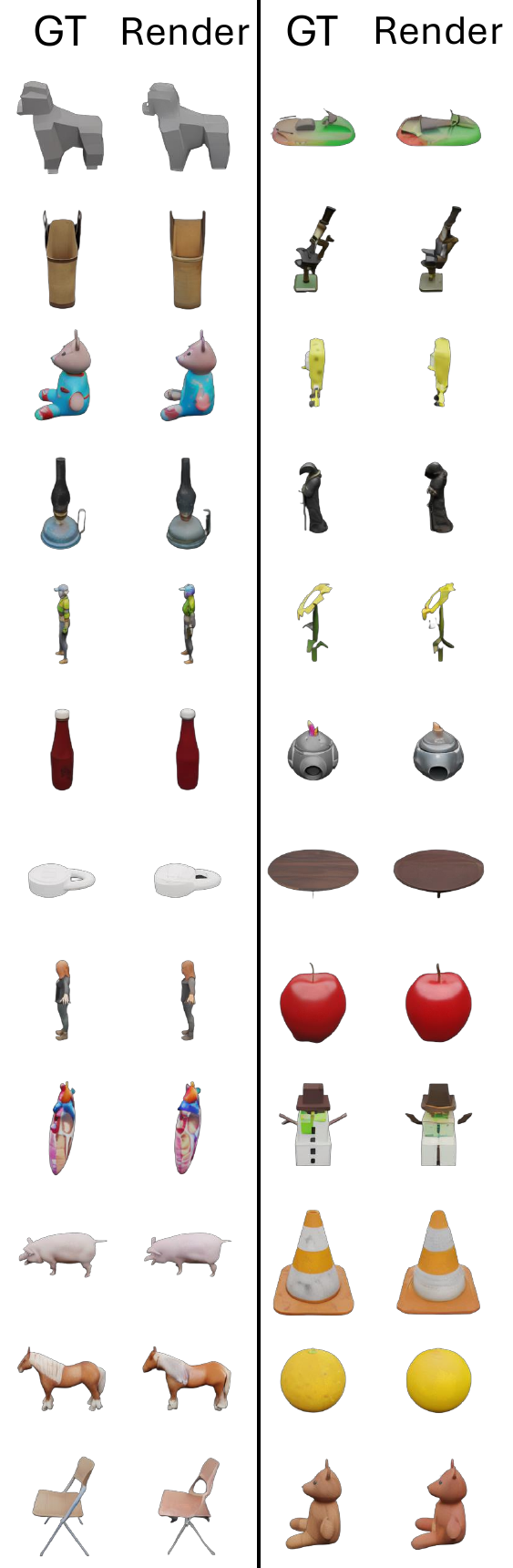}\\[0.5ex]
        \small (b)
    \end{minipage}
    \caption{\small \textbf{Decoding examples for Trellis-based 3D VQ-VAE on unseen Objaverse assets.}
    (a) Reconstruction of 3D shape encoding using Llama 3.2 as decoder. (b) Rendering from 3D shape encoding using Llama 3.2 as decoder.}
    \label{fig:3d-tokenization-decoding}
\end{figure*}

\mypar{3D Tokenization.} We present additional qualitative results for our 3D tokenization scheme. \cref{fig:3d-tokenization-examples} provides additional examples of how the 3D tokens are effective for reconstructions, as shown in Figure 4 (b) in the paper. \cref{fig:3d-tokenization-decoding} provides additional examples of how the 3D tokens are effective for decoding, as shown in Figure 4 (c) in the paper.


\begin{figure*}[!htbp]
\begin{center}
\includegraphics[width=\linewidth]{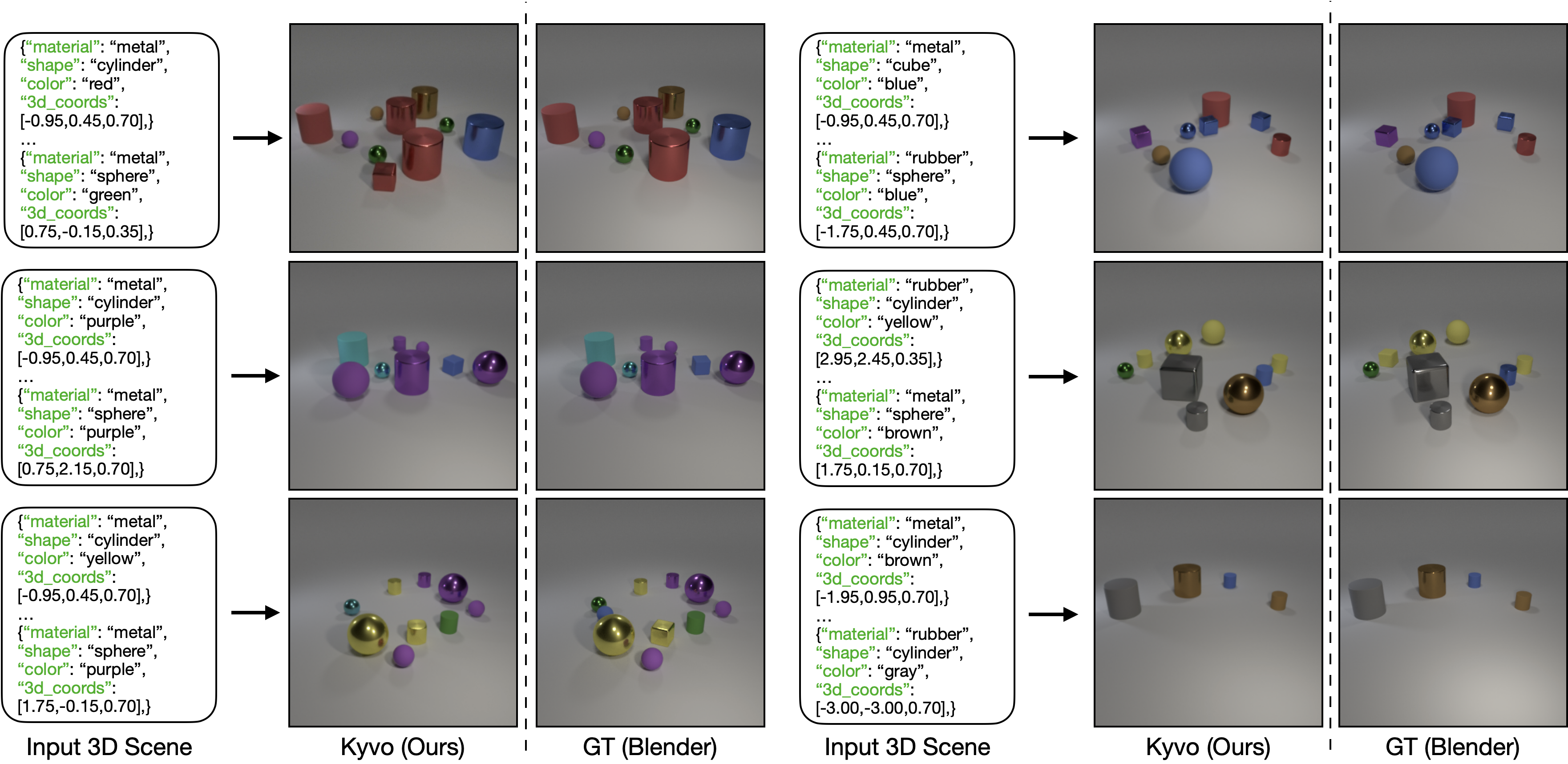} 
\end{center}
% \vspace{-3mm}
\caption{\small \textbf{Rendering examples for CLEVR.} Example image generations for the rendering task on CLEVR. The model takes a 3D scene as input and produces a corresponding image. Additionally, we show the ground-truth image rendered using Blender.}
\label{fig:app-rendering-examples-clevr}
% \vspace{-3mm}
\end{figure*}

\begin{figure*}[!htbp]
\begin{center}
\includegraphics[width=\linewidth]{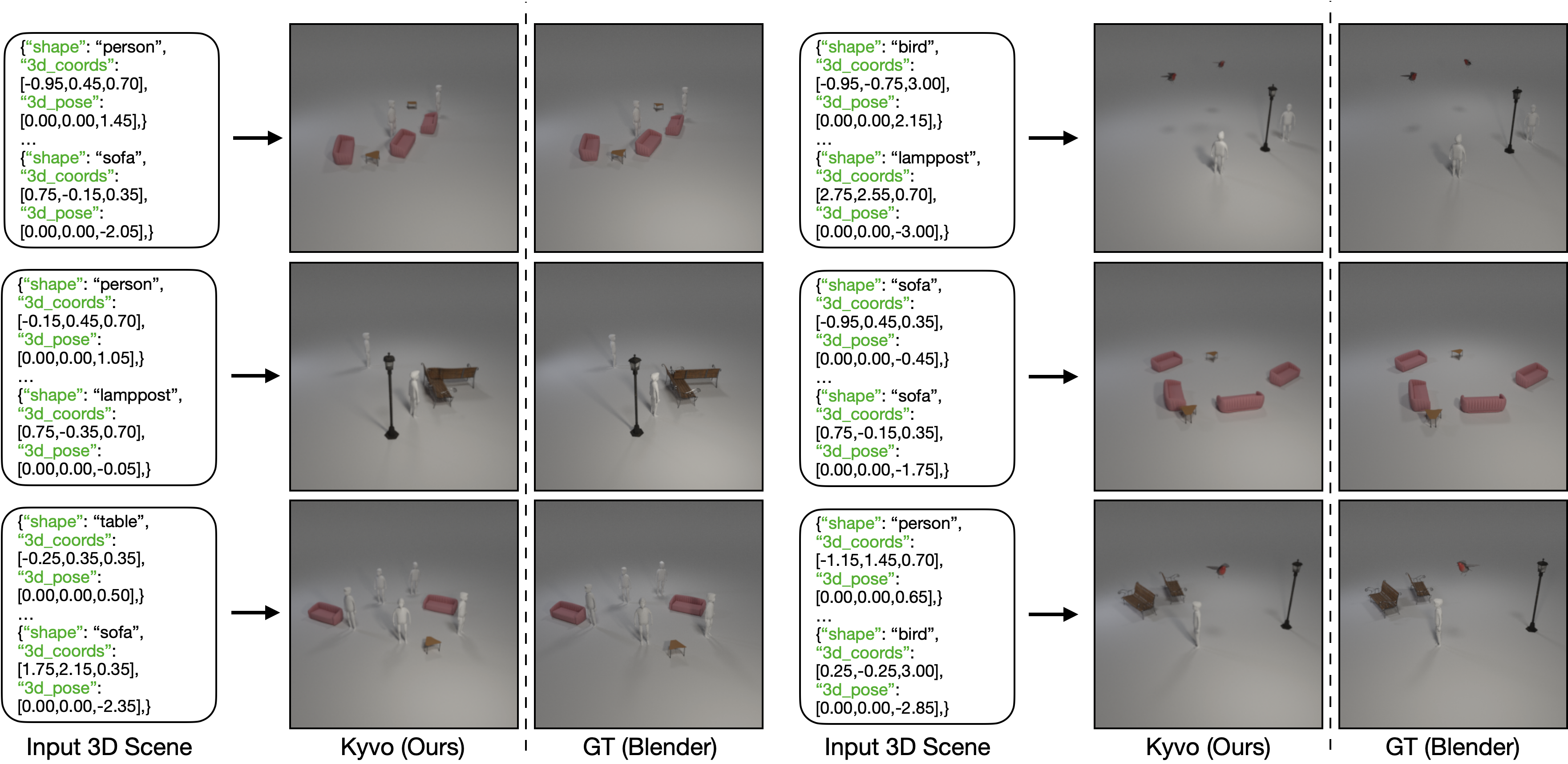} 
\end{center}
% \vspace{-3mm}
\caption{\small \textbf{Rendering examples for ObjaWorld.} Example image generations for the rendering task on ObjaWorld with complex shapes. The model takes a 3D scene as input and produces a corresponding image. Additionally, we show the ground-truth image rendered using Blender.}
\label{fig:app-rendering-examples-objaworld}
% \vspace{-3mm}
\end{figure*}

\begin{figure*}[!htbp]
\begin{center}
\includegraphics[width=\linewidth]{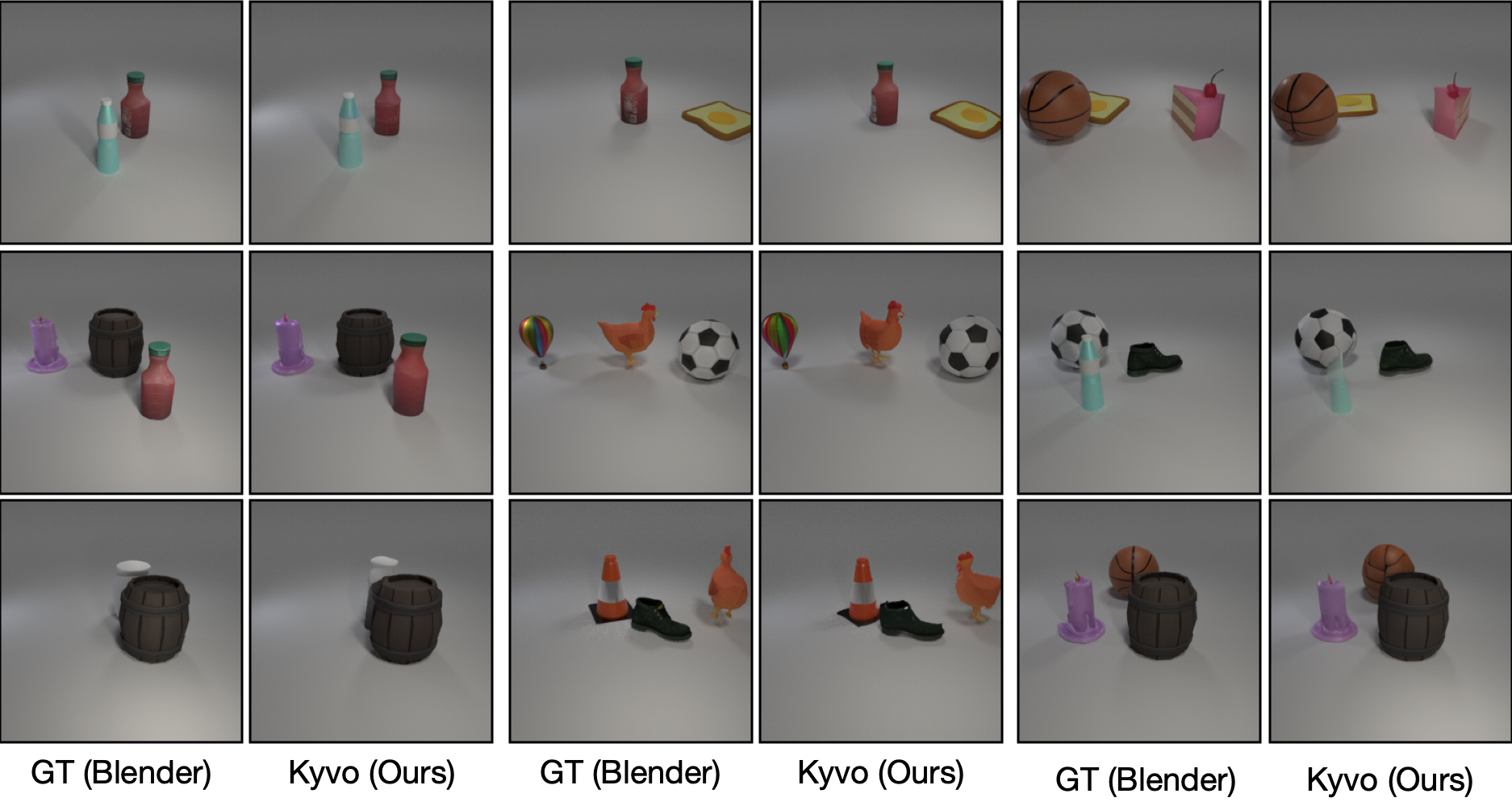} 
\end{center}
% \vspace{-3mm}
\caption{\small \textbf{Rendering examples for ObjaWorld with explicit shape encodings.} Example image generations for the rendering task on ObjaWorld. The model takes a 3D scene with embedded shape encodings as input and produces a corresponding image. Additionally, we show the ground-truth image rendered using Blender.}
\label{fig:app-rendering-examples-scenes-to-images}
% \vspace{3mm}
\end{figure*}

\mypar{Rendering.} We present qualitative results for the rendering task on CLEVR and ObjaWorld with complex shapes in Figures~\ref{fig:app-rendering-examples-clevr} and ~\ref{fig:app-rendering-examples-objaworld}, respectively. Results on ObjaWorld with explicit shape representations are shown in Figure~\ref{fig:app-rendering-examples-scenes-to-images}. The model takes only the structured 3D scene representation as input and predicts the corresponding image tokens. These tokens are then decoded using the VQGAN decoder, which reconstructs the image by mapping them from token-space to pixel-space.
For each example, we also provide the ground truth image rendered using Blender, allowing direct comparison with the model-generated output. As observed, the model effectively captures the 3D scene structure based solely on the JSON input and accurately synthesizes the corresponding image. Notably, Figure~\ref{fig:app-rendering-examples-scenes-to-images} demonstrates the model's ability to integrate multiple information sources: it successfully maps spatial arrangements from JSON scene descriptions and visual properties from asset sequences to generate coherent, realistic pixel-space representations.
% \aanote{Add ObjaWorld rendering examples}

However, certain failure cases highlight the model’s limitations. For instance, in the first example of Figure~\ref{fig:app-rendering-examples-clevr}, the model fails to predict a small red cube positioned at the front. Similarly, in the last example of Figure~\ref{fig:app-rendering-examples-objaworld}, the model mispredicts the bird's pose, causing it to face the wrong direction. In the third column of comparisons in Figure~\ref{fig:app-rendering-examples-scenes-to-images}, some distortions in the shapes and poses can be seen as well. Despite these occasional errors, the overall rendering quality demonstrates strong spatial understanding and scene reconstruction capabilities.


\begin{figure*}[!htbp]
\begin{center}
\includegraphics[width=\linewidth]{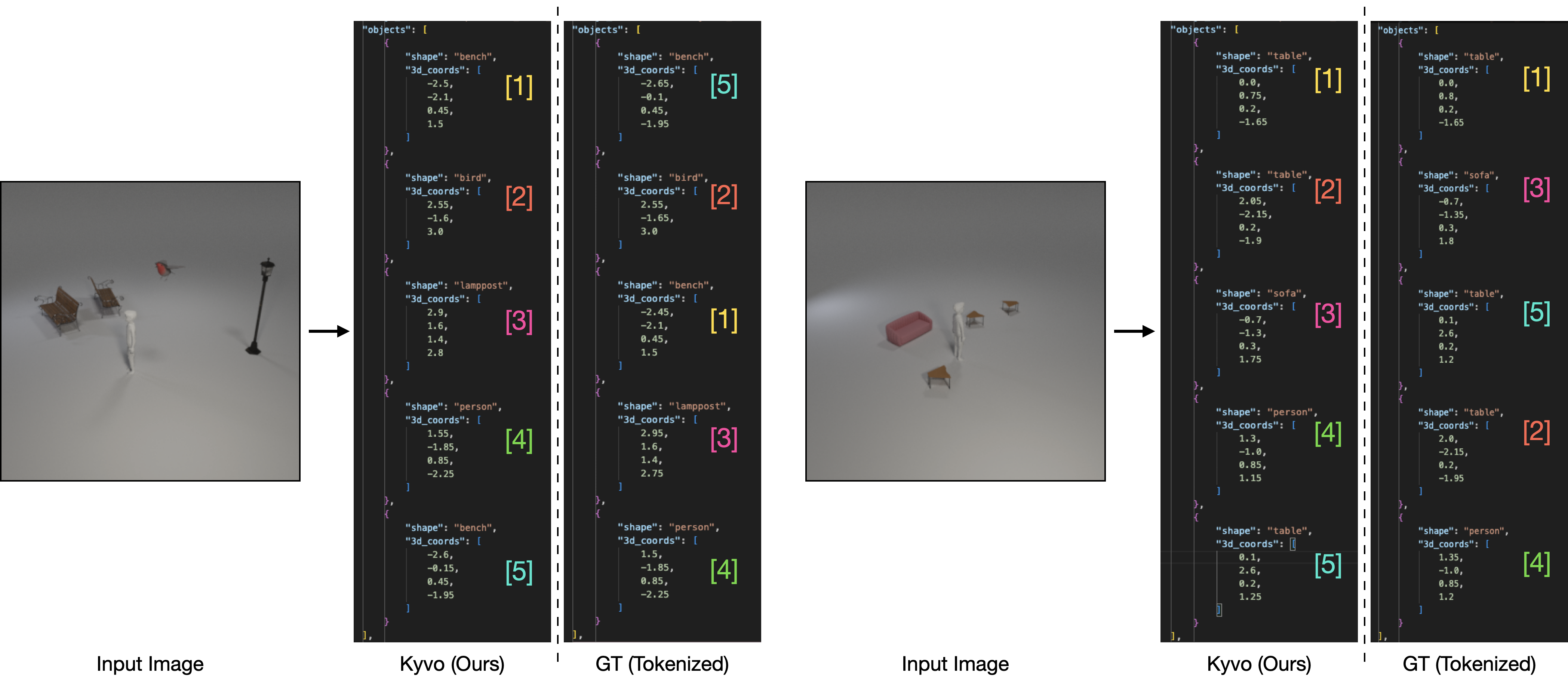} 
\end{center}
% \vspace{-3mm}
\caption{\small \textbf{Recognition examples for ObjaWorld.} Two example predictions from the recognition task on ObjaWorld. The colored numbers indicate object matching between the predicted and ground-truth scenes, based on the criteria for Jaccard Index as defined in Algorithm~\ref{alg:jaccard}. Note that the fourth number in the list is the azimuth pose value, this format of prediction saves sequence length.}
\label{fig:app-recognition-examples-clevr}
% \vspace{-3mm}
\end{figure*}

\mypar{Recognition.} In Figure~\ref{fig:app-recognition-examples-clevr} we show example recognition results on the ObjaWorld dataset with complex shapes. The model takes the image (tokenized) as input and outputs the sequential 3D representation. We then parse the predicted 3D scene into JSON format and display it alongside the ground truth JSON in the figure.
To evaluate recognition performance, we match the predicted objects with ground truth objects to compute the Jaccard Index (see Algorithm~\ref{alg:jaccard}). This matching is based on attribute similarity and spatial location criteria. Additional details on the Jaccard Index calculation are provided in Section~\ref{sec:app:evaluation}.
The matching results, visualized using colored numbers in Figure~\ref{fig:app-recognition-examples-clevr}, illustrate the model's strong ability to accurately predict object attributes and their 3D spatial locations almost accurately. While minor spatial deviations may occur, the model effectively reconstructs the structured 3D scene from image input.

\begin{figure*}[!htbp]
\begin{center}
\includegraphics[width=\linewidth]{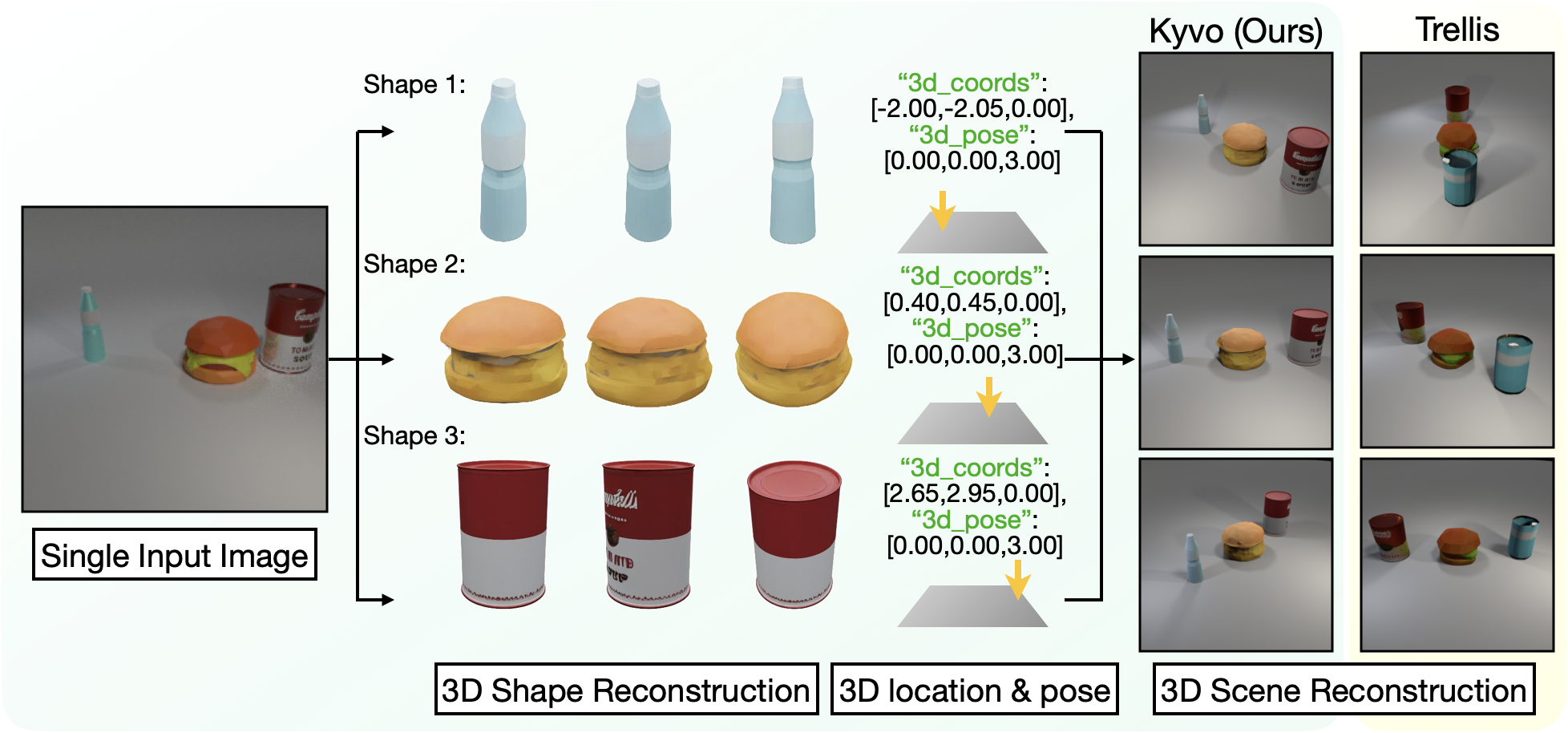}
\includegraphics[width=\linewidth]{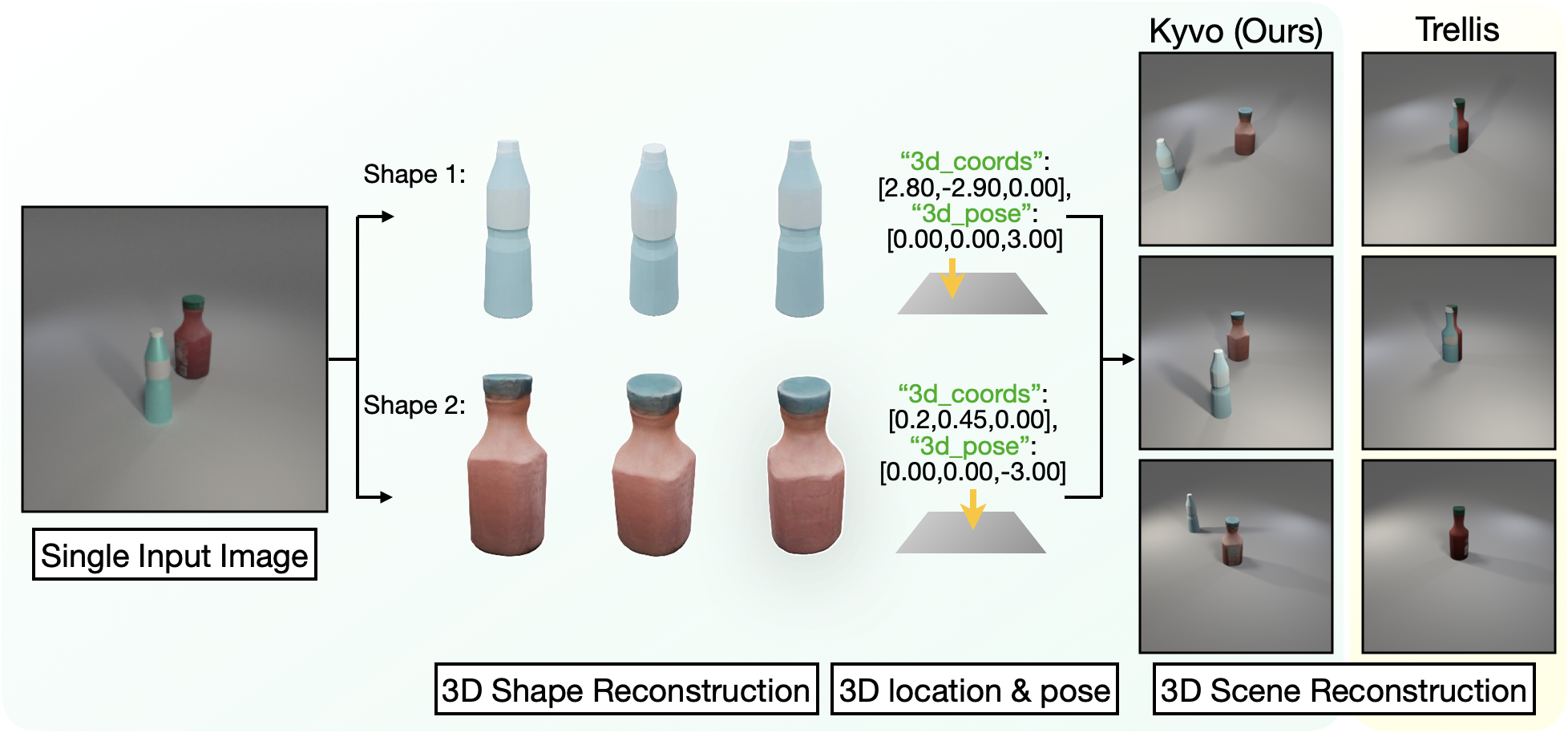}
\end{center}
% \vspace{-3mm}
\caption{\small \textbf{Unified shape and scene reconstruction examples.} Given a single input image, Kyvo predicts shape sequences and reconstructs individual objects (bottle, cheeseburger, \etc) along with their 3D locations and poses via our structured 3D modality, effectively reconstructing the 3D scene with consistent spatial relations between the objects, visualized using Blender.}
\label{fig:app-rendering-examples-images-to-scenes}
% \vspace{-3mm}
\end{figure*}

In Figure~\ref{fig:app-rendering-examples-images-to-scenes} we show examples of recognition task on ObjaWorld with explicit shape representations on unseen scenes. Specifically, the model takes a single images as input and reconstructs the full 3D geometry per object and infers each object's 3D position and pose to accurately reconstruct the 3D scene. As can be seen from Figure~\ref{fig:app-rendering-examples-images-to-scenes}, \llm3 accurately recovers object geometries and spatial layouts via our structured, object-centric 3D modality. In contrast, Trellis~\cite{xiang2024structured} often merges two objects into one (e.g. the second example) or hallucinates shapes (e.g.,
blue can in the first example) and misaligned layouts.


\begin{figure*}[!htbp]
\begin{center}
\includegraphics[width=\linewidth]{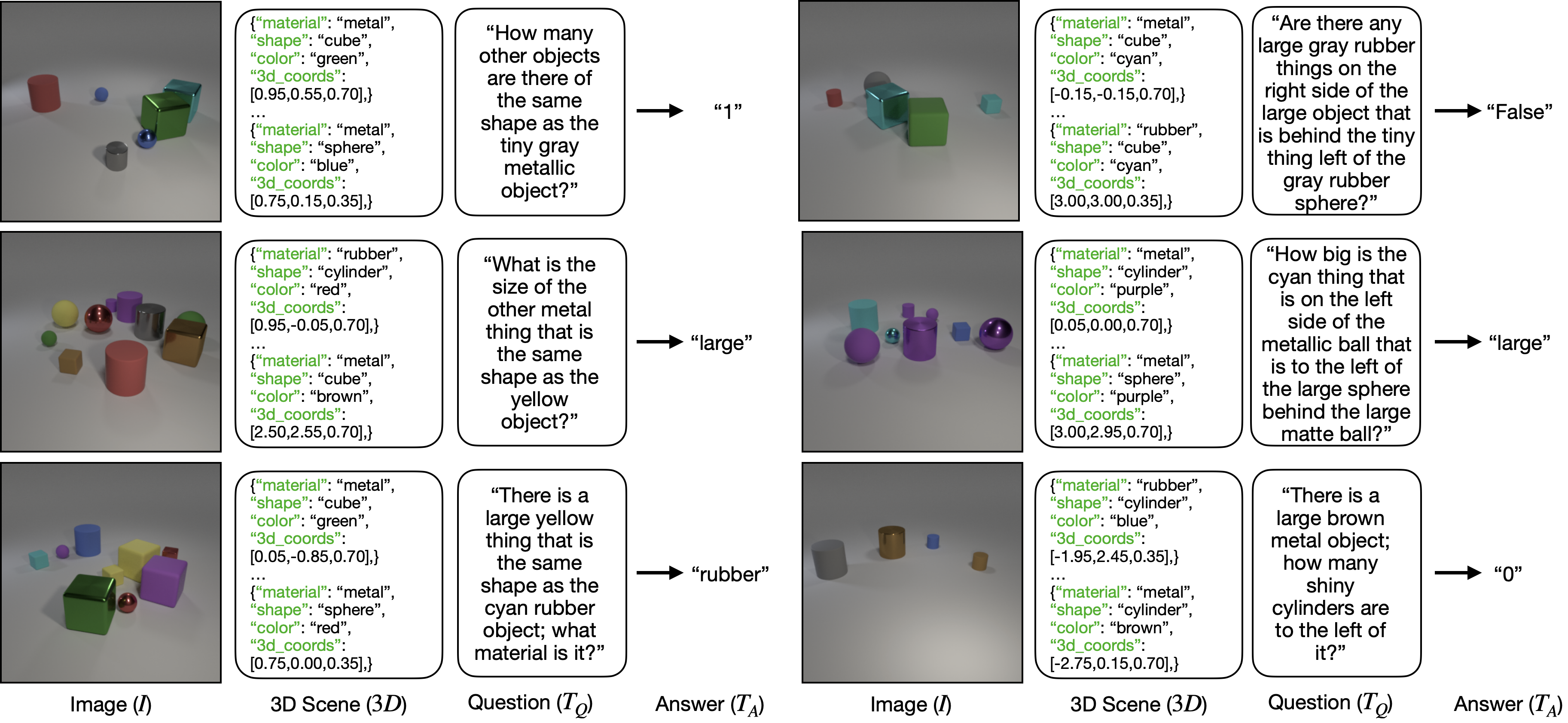} 
\end{center}
% \vspace{-3mm}
\caption{\small \textbf{Question-answering examples for CLEVR.} Example cases from the question-answering task on CLEVR. The model takes an image, a 3D scene, and a question as input to generate the corresponding answer.}
\label{fig:app-qa-examples-clevr}
% \vspace{-3mm}
\end{figure*}

\mypar{Question-answering.} In Figure~\ref{fig:app-qa-examples-clevr}, we present qualitative examples from the question-answering task on the CLEVR dataset. The figure showcases model predictions across a diverse set of question types, including binary (True/False) responses, categorical answers such as object sizes (``small'' or ``large''), and numerical values.
These examples highlight the model's ability to understand and reason about structured 3D scenes, demonstrating accurate comprehension of spatial relationships, object attributes, and numerical reasoning.

\section{Evaluation}
\label{sec:app:evaluation}

In this section we provide more details on the evaluation strategies that we adopt. We discuss the evaluation method for the three primary modalities as well as for the 3D shape encodings.

\subsection{3D scenes}
\label{subsec:app:3d-scenes}

We evaluate predicted 3D scenes using the Jaccard Index, computed via Algorithm~\ref{alg:jaccard}. Objects are matched between predicted and ground-truth scenes based on attribute similarity and spatial proximity.

\textit{Matching criteria by dataset:}
\begin{itemize}
\item CLEVR: Match shape, size, color, and material
\item ObjaWorld: Match shape with pose constraint (predicted pose within $\pm 0.15$ radians)
\item Objectron: Match category with dimension constraint (mean absolute error $\leq 0.05$)
\item ARKitScenes: Match category with dimension constraint (mean absolute error $\leq 1.00$)
\end{itemize}

\textit{Spatial proximity thresholds ($\tau$):}
We average Jaccard Index across multiple threshold values:
\begin{itemize}
\item CLEVR, ObjaWorld, Objectron: $\tau \in \{0.05, 0.10, 0.15, 0.20, 0.25\}$
\item ARKitScenes: $\tau \in \{1.25, 1.50, 1.75, 2.00, 2.25\}$
\end{itemize}

\begin{figure*}[!htbp]
\begin{center}
\includegraphics[width=0.6\linewidth]{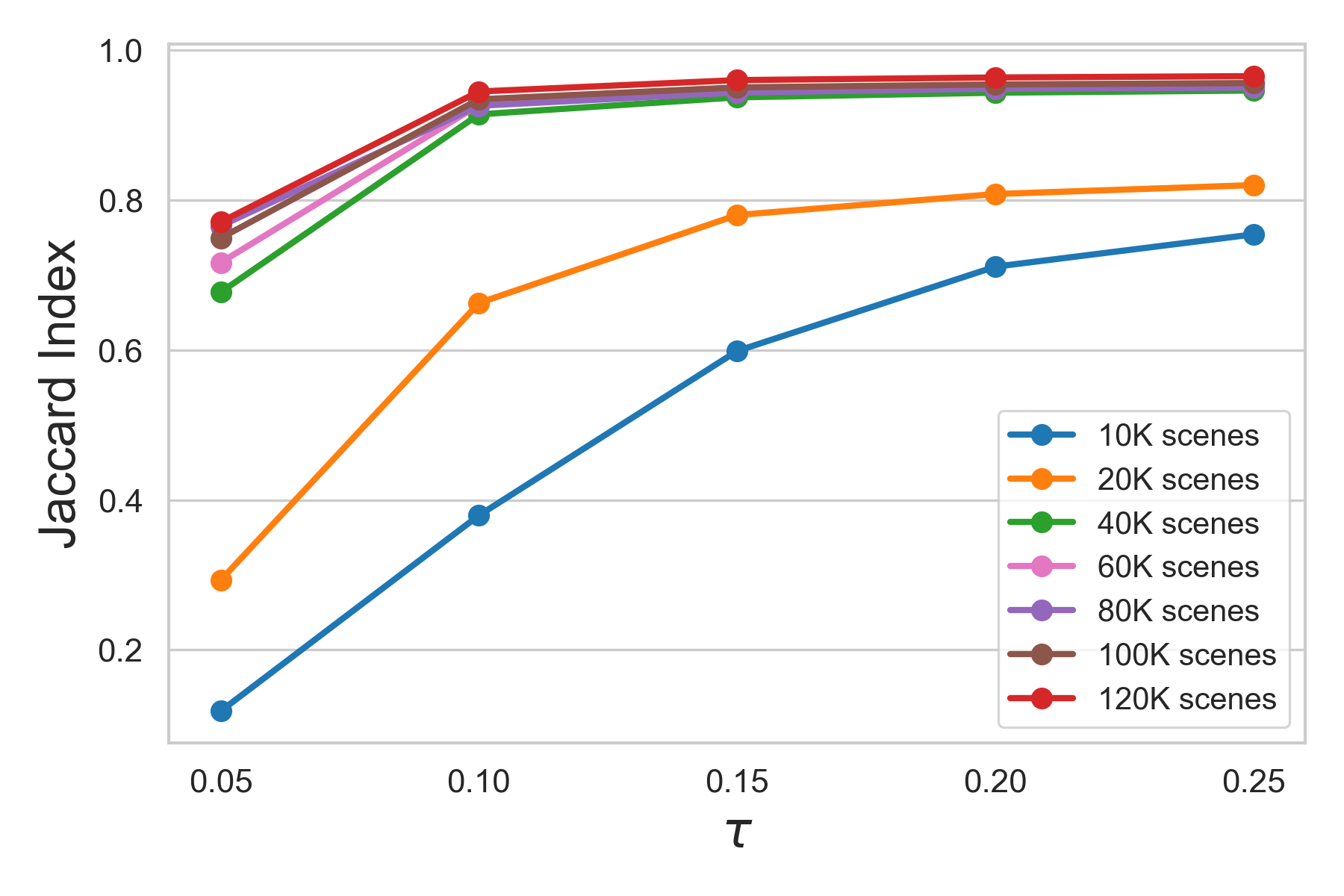} 
\end{center}
% \vspace{-3mm}
\caption{\small \textbf{Effect of $\tau$.} The plot shows the impact of $\tau$ on the Jaccard Index for models trained with increasing amounts of training data on CLEVR for the recognition task. The drop in Jaccard Index with decreasing $\tau$ is more pronounced for models trained on smaller datasets. Higher-performing models demonstrate greater robustness to changes in $\tau$.}
\label{fig:app-effect-of-tau}

\end{figure*}

Lower $\tau$ values impose stricter spatial constraints, requiring predicted objects to be closer to ground-truth positions. Figure~\ref{fig:app-effect-of-tau} illustrates how $\tau$ affects Jaccard Index on CLEVR across different training data sizes.

\mypar{Comparison with Trellis.} We compare our unified shape and scene reconstruction against Trellis (Figure 9, main paper; Figure~\ref{fig:app-rendering-examples-images-to-scenes}). Trellis reconstructs scenes by inputting an image to a rectified flow transformer (DiT) that generates a scene-level SLAT representation, which is subsequently decoded into a single holistic 3DGS. In contrast, Kyvo employs a different approach through two key distinctions: (1) \textit{Scene decomposition}: Kyvo decomposes scenes into constituent objects, each parameterized by shape, 3D location, and orientation within our structured 3D modality. (2) \textit{Quantized representation}: While both methods utilize SLAT representations for object shapes, Kyvo vector-quantizes these representations via our 3D VQ-VAE, to slot naturally into Kyvo's autoregressive generation framework. This decomposition enables Kyvo to achieve precise reconstruction of individual objects while simultaneously inferring their 3D spatial locations and relationships. Moreover, we obtained an average Jaccard Index of $0.666$ averaged over $\tau \in \{0.50, 0.75, 1.00, 1.25, 1.50\}$ for this recognition model. We observed that the model seems to have a relatively harder time predicting the location coordinates when explicit shape sequences are involved as compared to when the shapes are identified using a word token like in CLEVR.
% \aanote{add JI value}
% \aanote{Elaborate more}

\begin{algorithm*}[htb]
   \caption{Compute Jaccard Index}
   \label{alg:jaccard}
\begin{algorithmic}[1]
   \STATE \textbf{Input:} GTS (list of ground-truth scenes), PS (list of predicted scenes), distance threshold \(\tau\)
   \STATE \textbf{Output:} Average JaccardIndex

   \STATE Initialize \texttt{JaccardIndex} = 0

   \FOR{(G, P) in zip(GTS, PS)}
       \STATE \texttt{GObjs} \(\leftarrow\) G.objects
       \STATE \texttt{PObjs} \(\leftarrow\) P.objects
       \STATE \texttt{matchedFlags} \(\leftarrow\) Boolean array of length \(|\texttt{GObjs}|\), initialized to \texttt{False}
       \STATE Initialize \texttt{TP} = 0, \texttt{FP} = 0, \texttt{FN} = 0

       \FOR{\texttt{p} in \texttt{PObjs}}
           \STATE \texttt{foundMatch} \(\leftarrow\) \texttt{False}
           \FOR{\(j = 1\) to \(|\texttt{GObjs}|\)}
               \IF{\(\neg \texttt{matchedFlags}[j]\) 
                   and \(\texttt{attributesMatch}(\texttt{p}, \texttt{GObjs}[j])\) 
                   and \(\texttt{dist}(\texttt{p.coords}, \texttt{GObjs}[j].\texttt{coords}) < \tau\)}
                   \STATE \(\texttt{matchedFlags}[j] \leftarrow \texttt{True}\)
                   \STATE \(\texttt{foundMatch} \leftarrow \texttt{True}\)
                   \STATE \(\texttt{TP} \leftarrow \texttt{TP} + 1\)
                   \STATE \textbf{break}
               \ENDIF
           \ENDFOR
           \IF{\texttt{foundMatch} = \texttt{False}}
               \STATE \(\texttt{FP} \leftarrow \texttt{FP} + 1\)
           \ENDIF
       \ENDFOR

       \STATE \(\texttt{FN} \leftarrow \texttt{FN} + \text{count}(\texttt{matchedFlags} = \texttt{False})\)

       \STATE \(\texttt{JaccardIndex} += \frac{\texttt{TP}}{\texttt{TP} + \texttt{FP} + \texttt{FN}}\)
   \ENDFOR

    \STATE \(\texttt{JaccardIndexAvg} \leftarrow \texttt{JaccardIndex} / |\text{PS}|\)

   \STATE \textbf{return} \(\texttt{JaccardIndexAvg}\)

\end{algorithmic}
\end{algorithm*}

\subsection{Images}
\label{subsec:app:images}

\begin{figure}[!htbp]
\begin{center}
    \includegraphics[width=\linewidth]{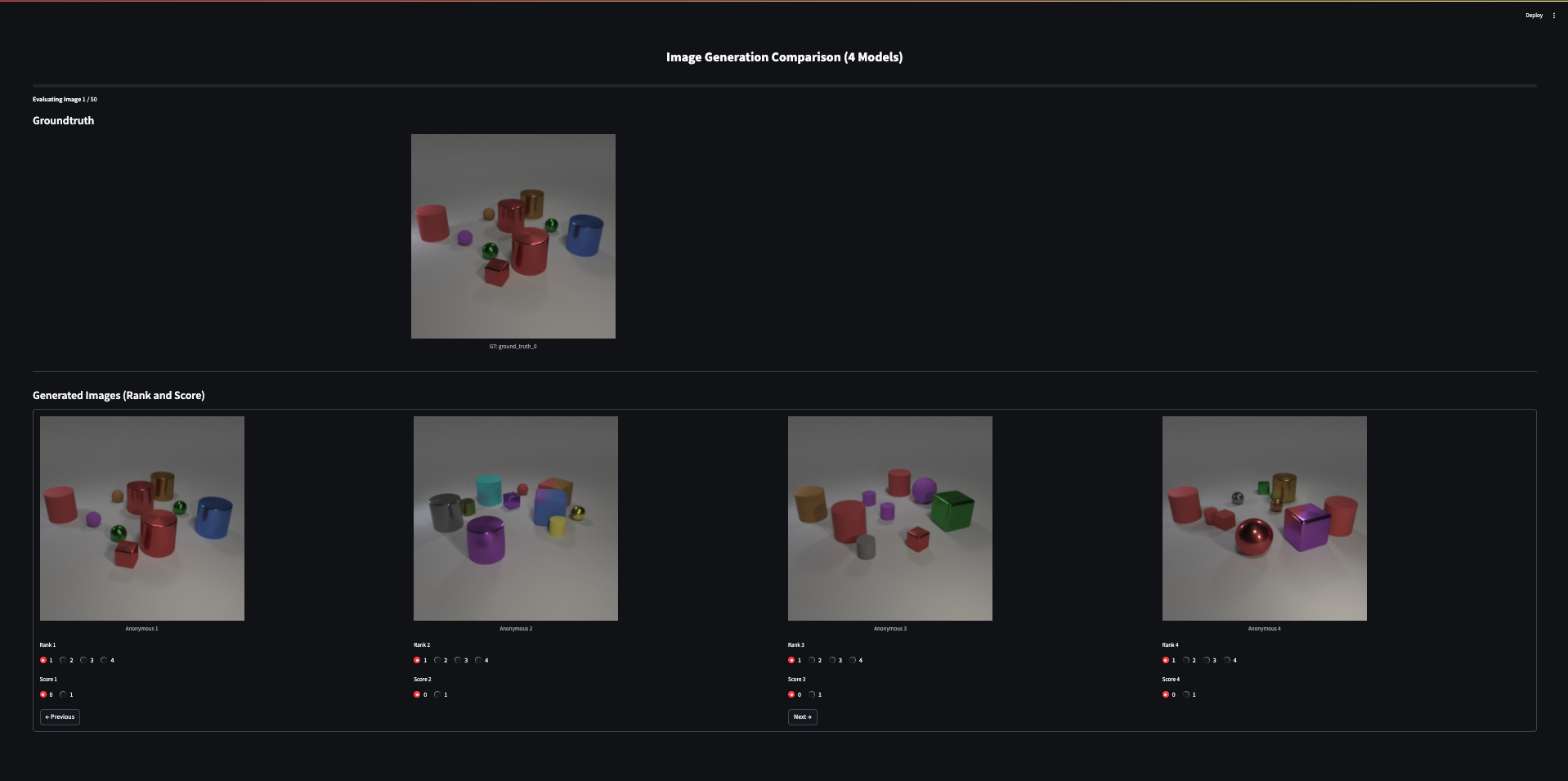}
\end{center}
% \vspace{-3mm}
\caption{\textbf{Human Evaluation User Interface.} A snapshot of the user interface used for human evaluation of generated images. This example is taken from an experiment analyzing the effect of center-token reordering and weighted loss, comparing four models. The results of this experiment are presented in Table 2 of the main paper.}
\label{fig:human-eval-ui}
\end{figure}

\mypar{Human Evaluation.} As discussed in the main paper, we rely on human evaluations to assess image generations by the model, as object-level nuances often go beyond the scope of quantitative metrics. To facilitate this evaluation, we designed a user interface, a snapshot of which is shown in Figure~\ref{fig:human-eval-ui}. For each comparison involving $\text{N}$ models, the interface presents users with $\text{N}$ generated images alongside the ground truth image, all displayed in a shuffled and anonymized order. This ensures that users remain unaware of which model generated each image, mitigating potential biases in evaluation. In each comparison, users are asked to assign both a \textit{score} and a \textit{rank} to every generated image based on its visual fidelity and alignment with the ground truth. Figure~\ref{fig:human-eval-ui} shows a snapshot of the evaluation of images from the experiment where we studied the effect of center-token reordering and weighted loss on model generation involving four models (results reported in Table 4 of the main paper).

\myparit{Score:} The score takes a binary value of ``$0$'' or ``$1$'' and signifies the complete correctness of an image generation. The user is asked to provide a score of ``$1$'' only if the user believes that all the objects in the scene were accurately generated and accurately placed spatially. If the generated image has any differences with the groundtruth, \eg a cube was not generated correctly, the user provides a score of ``$0$". The score value is independent of any other model involved in the comparison and solely depends on the model under consideration and the groundtruth.

\myparit{Rank:} The rank takes a value from $\{$``$1$'', ``$2$'', ..., ``$\text{N}$''$\}$. The user is expected to rank the $\text{N}$ images from ``$1$" to ``$\text{N}$" by comparing the generation quality among them. If the user is unable to decide between two images, then we allow equal rank assignment to more than one image, \eg if two models equally perform for a given scene, they get the same rank value.

We consider a test set of $50$ scenes and use the user interface for scoring and ranking the generations. Specifically, let's say we want to have an evaluation on the effect of granularity, then we consider the image generated by the models for the $50$ scenes and score and rank them as discussed above. In the main paper, we report the \textit{mean rank} (the lower, the better) over the $50$ images in the tables. In addition to the \textit{mean rank}, we obtain the \textit{mean score} (the higher, the better) for every model. We also compute the \textit{winning rate} (the higher, the better) that is defined as the fraction of times a given model was assigned a rank of $1$. We report the \textit{mean score} and \textit{winning rate} for all the models in Table~\ref{table:app-image-metrics}.

\begin{figure*}[!htbp]
\begin{center}
\includegraphics[width=0.75\linewidth]{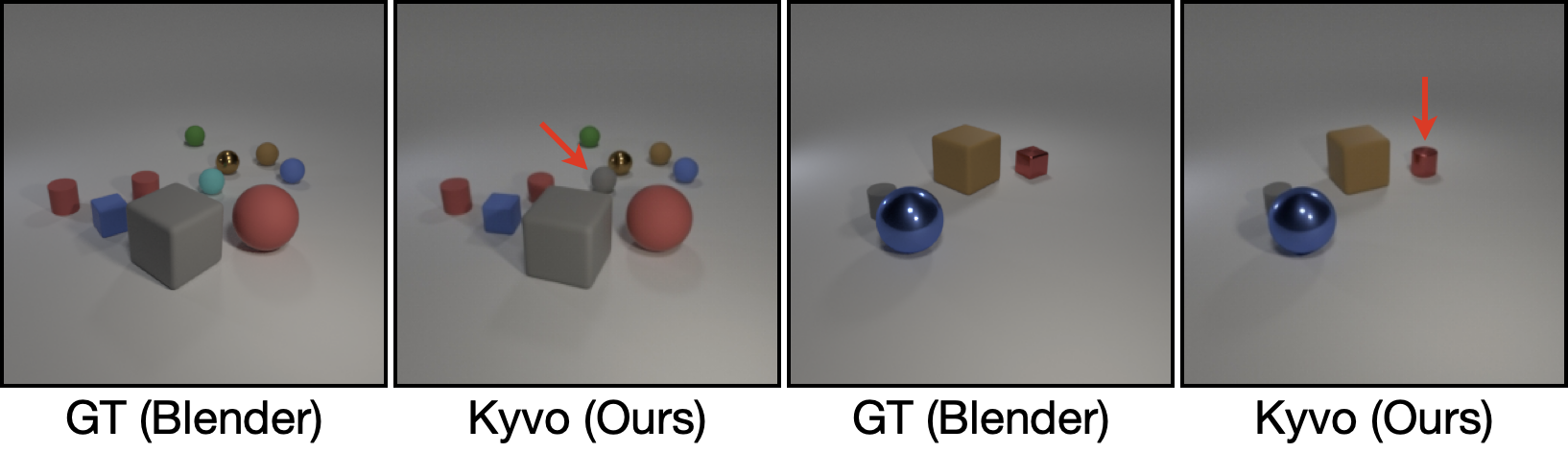} 
\end{center}
% \vspace{-3mm}
\caption{\small Object-level nuances are challenging for image metrics, like SSIM and L2-loss, to capture, prompting the need for human evaluation. The predicted image is incorrect in both cases but differs only subtly from the groundtruth.}
\label{fig:app-ssim-visual}

\end{figure*}

\mypar{SSIM and L2-loss.} Evaluation of the generated images requires careful assessment of the objects and their attributes in the scene. For example, consider the example predictions in Figure~\ref{fig:app-ssim-visual}. For both cases, the predicted image is incorrect and minutely differs from the groundtruth. For the first case, the small cyan sphere gets an incorrect gray color, while in the second case, the small cube gets mispredicted as a cylinder. Quantitative metrics like SSIM and L2-loss fail to capture these subtle differences between the predicted and the groundtruth images that occupy a very small pixel region, leading us to qualitative human evaluations. However, for experimental completeness, we computed the SSIM and pixel-wise L2-loss for all the models and reported them in Table~\ref{table:app-image-metrics}. While the values show similar trends to human evals, we report human evals in the main paper as they directly assess image correctness.

\begin{table*}[htbp]

\definecolor{Gray}{gray}{0.90}
\definecolor{LightCyan}{rgb}{0.88,1,1}

\newcolumntype{a}{>{\columncolor{Gray}}c}
\renewcommand{\arraystretch}{1.2}

\scriptsize
\begin{center}
    
\resizebox{0.8\textwidth}{!}{
\begin{tabular}{l|ccc|cc}
\toprule
\textbf{Comparison} & \multicolumn{5}{c}{\textbf{Metrics}} \\
\hline
\textbf{Granularity} & Mean Rank ($\downarrow$) & Winning Rate ($\uparrow$) & Mean Score ($\uparrow$) & SSIM ($\uparrow$) & L2-loss ($\downarrow$) \\
\hline
$0.005$ & $1.38$ & $0.70$ & $0.74$ & $0.9527$ & $0.0021$ \\
$0.05$ & $1.20$ & $0.82$ & $0.80$ & $0.9505$ & $0.0010$ \\
$0.5$ & $2.02$ & $0.28$ & $0.52$ & $0.9030$ & $0.0010$ \\
\hline
\textbf{Number Encoding} & Mean Rank ($\downarrow$) & Winning Rate ($\uparrow$) & Mean Score ($\uparrow$) & SSIM ($\uparrow$) & L2-loss ($\downarrow$) \\
\hline
Fixed Sine-Cosine & $1.44$ & $0.64$ & $0.64$ & $0.9525$ & $0.0010$ \\
Learned & $1.58$ & $0.60$ & $0.60$ & $0.9487$ & $0.0011$ \\
Fixed Sine-Cosine + Learned & $1.28$ & $0.78$ & $0.78$ & $0.9505$ & $0.0010$ \\
\hline
\textbf{CT Reordering, Weighted Loss} & Mean Rank ($\downarrow$) & Winning Rate ($\uparrow$) & Mean Score ($\uparrow$) & SSIM ($\uparrow$) & L2-loss ($\downarrow$) \\
\hline
\ding{55}, \ding{55} & $2.66$ & $0.00$ & $0.00$ & $0.8575$ & $0.0049$ \\
\ding{51}, \ding{55} & $3.56$ & $0.00$ & $0.00$ & $0.8410$ & $0.0061$ \\
\ding{55}, \ding{51} & $2.78$ & $0.00$ & $0.00$ & $0.8668$ & $0.0044$ \\
\ding{51}, \ding{51} & $1.00$ & $1.00$ & $0.84$ & $0.9505$ & $0.0010$ \\
\hline
\textbf{Recipe} & Mean Rank ($\downarrow$) & Winning Rate ($\uparrow$) & Mean Score ($\uparrow$) & SSIM ($\uparrow$) & L2-loss ($\downarrow$) \\
\hline
Scratch & $1.36$ & $0.68$ & $0.72$ & $0.9515$ & $0.0010$ \\
LoRA & $1.82$ & $0.48$ & $0.56$ & $0.9507$ & $0.0010$ \\
FFT & $1.26$ & $0.8$ & $0.82$ & $0.9505$ & $0.0010$ \\
\hline
\textbf{Backbone} & Mean Rank ($\downarrow$) & Winning Rate ($\uparrow$) & Mean Score ($\uparrow$) & SSIM ($\uparrow$) & L2-loss ($\downarrow$) \\
\hline
Llama-3.2-1B & $1.38$ & $0.76$ & $0.78$ & $0.9521$ & $0.0010$ \\
Llama-3.2-1B-Instruct & $1.28$ & $0.72$ & $0.80$ & $0.9505$ & $0.0010$ \\
Llama-3.2-3B-Instruct & $1.18$ & $0.84$ & $0.86$ & $0.9539$ & $0.0010$ \\
\bottomrule
\end{tabular}}
\end{center}

\caption{\small \textbf{Image Metrics.} Comparison of mean rank, winning rate, and mean score from human evaluation across all models for the CLEVR rendering task. Additionally, we provide SSIM and pixel-wise L2 loss values for each model.}
\label{table:app-image-metrics} 

\end{table*}

\subsection{3D assets}
\label{subsec:app:assets}
For the same reasons as above, we use human evaluations to assess the quality of 3D assets when it comes to the 3D tokenizer. In Table 1 in the main paper, we use human evaluations to quantify the effect of auxiliary reconstruction losses on the reconstruction quality of the 3D VQ-VAE, and report mean rank. In Table 2 in the main paper, we use human evaluations to compare the reconstruction quality of our Trellis-based 3D VQ-VAE with that of SAR3D, and report mean rank. In both cases, the human evaluation is facilitated using an anonymized, shuffled interface as described above, and only relative rank is assessed and reported.

\subsection{Text} 
\label{subsec:app:text}
Among the four tasks we consider, only question-answering produces text output. The question templates used in CLEVR cover a diverse range of answer types. Some questions require binary (True/False) responses, others expect numerical values ranging from 0 to 10, while some answers involve text words describing attributes like ``small'',``green'', ``metal'', \etc. Table~\ref{table:app-qa-data-summary} provides a breakdown of the different question types in the training set.

Using these distributions, we establish two baseline accuracies on the test set: random and frequency. For the random baseline, we predict answers uniformly at random for each question type, yielding a mean accuracy of $0.359$ with a standard deviation of $0.009$ over $100$ runs. For the frequency baseline, we predict the most common answer for each question type based on its distribution in the training data (as shown in Table~\ref{table:app-qa-data-summary}), achieving a test accuracy of $0.430$.

\begin{table}[htbp]
  \centering
  \scriptsize
  {%
    % local number formatting for this table only
    \sisetup{group-separator={,}, group-minimum-digits=4}

    \resizebox{0.5\linewidth}{!}{%
    \begin{tabular}{@{}l
      S[table-format=4.0]   % “# questions”
      l@{}}
      \toprule
      \textbf{Question type} & \textbf{\# Questions} & \textbf{Majority answer} \\
      \midrule
      True/False        & 8080 & False     \\
      Number (0–10)     & 4401 & 0         \\
      Shape             & 1840 & cylinder  \\
      Color             & 1919 & cyan      \\
      Material          & 1840 & metal     \\
      Size              & 1920 & small     \\
      \midrule
      \multicolumn{3}{@{}c@{}}{\textbf{Baseline accuracy (test set)}} \\ \midrule
      Random (100 runs) & \multicolumn{2}{c}{0.359 ± 0.009} \\
      Frequency         & \multicolumn{2}{c}{0.430}         \\
      \bottomrule
    \end{tabular}}%
  }
  \vspace{2mm}
  \caption{\small\textbf{Question-Answering Data.} Statistics of various question types in the training dataset and baseline accuracies on the test set.}
  \vspace{-6mm}
  \label{table:app-qa-data-summary}
\end{table}

\section{Additional implementation details}
\label{sec:app:additional-implementation-details}

\subsection{Image VQGAN architecture}
\label{subsec:app:vqgan-architecture}

\begin{figure*}[!htbp]
\begin{center}
% \textbf{Pretrained VQGAN Architecture:}
% \includegraphics[width=0.8\linewidth]{figures-appendix/PretrainedVQGANArchitecture.png}
% \\
% \textbf{Domain-Specific VQGAN Architecture:}
\includegraphics[width=0.8\linewidth]{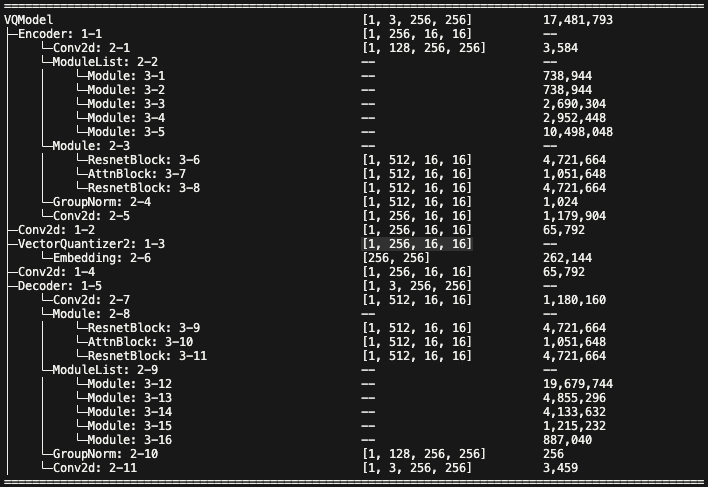} 
 
\end{center}
% \vspace{-3mm}
\caption{\small \textbf{VQGAN Architecture.} We show the detailed architecture of the VQGAN model that we use to train a domain-specific codebook. We show the output shapes for an input shape ($1$, $3$, $256$, $256$).}
\label{fig:app-vqgan-architecture-comparison}

\end{figure*}

In this section, we provide additional implementation details for the VQGAN architecture employed in our experiments to represent images. Figure~\ref{fig:app-vqgan-architecture-comparison} illustrates the complete network architecture, with output shapes shown for each layer given an input of shape ($1$, $3$, $256$, $256$). Our VQGAN configuration uses a $1024$-entry codebook with $256$-dimensional embeddings, trained on $256\times256$ resolution images. The encoder produces quantized embeddings of shape ($1$, $256$, $16$, $16$), yielding $16\times16=256$ discrete tokens per image that correspond to learned codebook entries.
For each dataset, we initialize the VQGAN model with ImageNet pretrained weights from~\cite{esser2021taming}, then fine-tune on the respective training set for 100 epochs. This fine-tuned model serves as both the encoder for converting images to token sequences and the decoder for reconstructing images from predicted tokens during evaluation.

\subsection{3D VQ-VAE training}
\label{subsec:app:3d-vq-vae-training}

In Section 3.2.1 of the main paper we describe how we train a 3D VQ-VAE to compress the $64^3\times8$ slats to a dense $8^3\times128$ latent. Here, we provide some additional details on the training. A slat contains $\sim\!20$k sparse voxels (in a $64^3$ grid, with vectors of dimension $8$ at active positions). We train the 3D VQ-VAE with these slats in 3 steps: (i) it \emph{densifies} the sparse tensor onto a $64^{3}\!\times8$ grid, (ii) encodes that grid with a 3-D convolutional U-Net down to an $8^{3}\!\times128$ latent volume, and (iii) vector-quantizes every latent with an $8192$-entry codebook. 

\paragraph{Architecture.}
The original Trellis SLAT encoder and decoder remain frozen and provide supervision.  
Sparse latents $(z_i,p_i)$ are first rasterised into a zero-initialised $64^{3}\!\times8$ grid; using a learnable padding vector instead of zeros showed no measurable gain and is therefore omitted.  
The dense grid is processed by a 3D U-Net that downsamples as $64^{3}\!\to32^{3}\!\to16^{3}\!\to8^{3}$ with channel widths $(32,128,512,1024)$.  
Each $8^{3}$ cell outputs a $128$-dim vector, quantized to the nearest of $8192$ code vectors updated by exponential moving average ($\tau{=}0.99$).  Gradients flow to the encoder through the straight-through estimator.

\paragraph{Optimization.}
We train for $200$k steps on $\sim168$k Sketchfab assets from Objaverse (further details provided in \cref{subsec:app:data-generation}) with a batch size of $8$. We use an AdamW optimizer ($\beta_1{=}0.9,\beta_2{=}0.999$) with a constant learning-rate of $3{\times}10^{-4}$, no weight decay, mixed precision, and adaptive gradient clipping.  

\smallskip
The training objective combines four terms:
\begin{equation}
\begin{aligned}
\mathcal{L}
= {} &
\underbrace{\|x-\hat{x}\|_2^{2}}_{\text{dense-SLAT recon}}
+\beta\,\underbrace{\|\,z-\mathrm{sg}[e]\|_2^{2}}_{\text{commit}}
\\
&\quad
+\lambda_{\text{KL}}\,D_{\text{KL}}
+\gamma\,\mathcal{L}_{\text{render}}
\end{aligned}
\tag{1}
\end{equation}
where  
\(x\in\mathbb{R}^{64^3\times8}\) is the rasterised SLAT voxel,  
\(\hat{x}\) its VQ–VAE reconstruction,  
\(\mathrm{sg}[\cdot]\) denotes the stop-gradient operator. In the commitment term, \(z\in\mathbb{R}^{8^{3}\times128}\) denotes the pre-quantization output of the encoder, while \(e\) is the corresponding vector-quantized output selected from the codebook.  The \(\mathcal{L}_{\text{render}}\) follows TRELLIS:
\[
\mathcal{L}_{\text{render}}
  = \mathcal{L}_{1}(I,\hat{I})
  + 0.2\!\left(1-\text{SSIM}(I,\hat{I})\right)
  + 0.2\,\text{LPIPS}(I,\hat{I}),
\]
computed between images rendered from ground-truth images \(I\) and rendering from Gaussian reconstructions \(\hat{I}\). This is the auxiliary pixel-space reconstruction loss discussed in the main paper.
We set \(\beta=0.25,\;\lambda_{\text{KL}}=10^{-6},\;\gamma=0.1\).  

The resulting codebook is used to represent objects using $8^3 = 512$ discrete tokens, enabling unified processing with structured 3D representations for autoregressive modeling. We employ bi-directional attention within shape token sequences for full intra-sequence connectivity when training the LLM.

\begin{figure*}[!htbp]
\begin{center}

\includegraphics[width=0.8\linewidth]{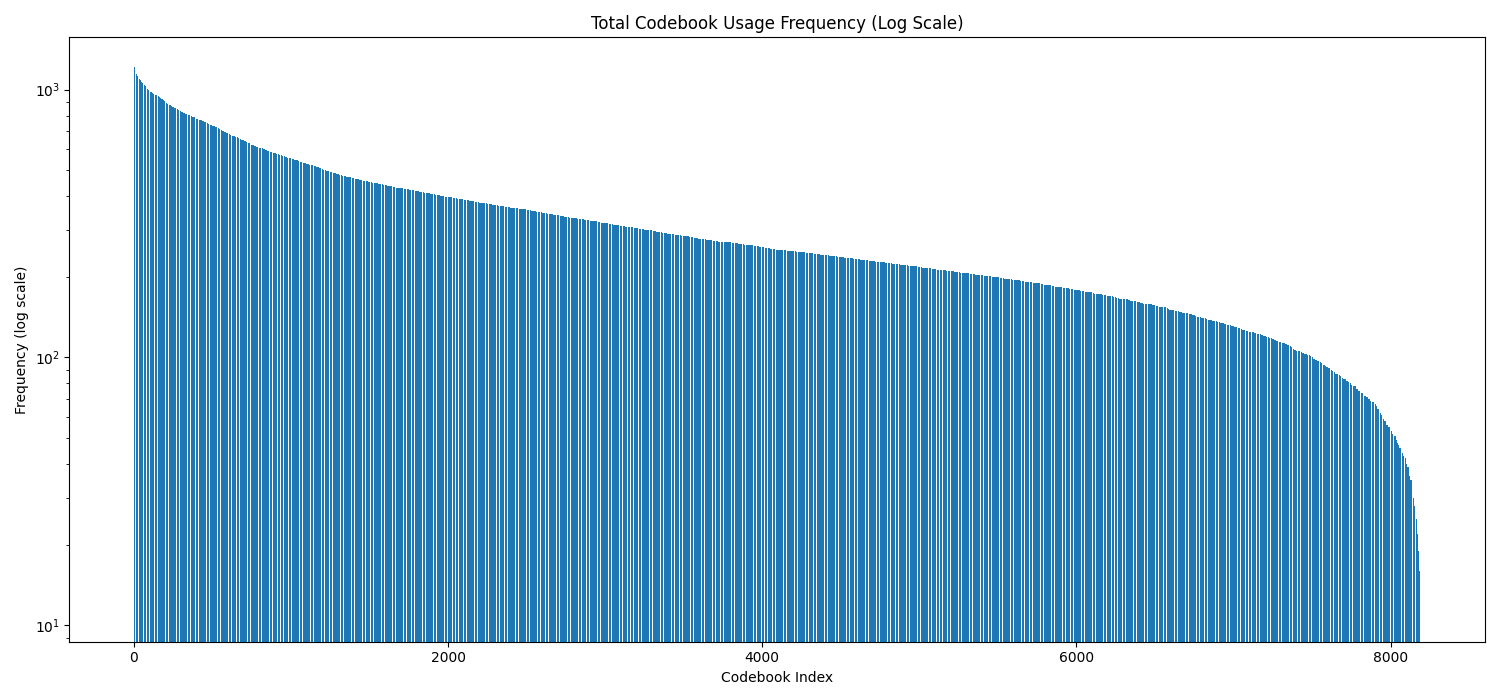} 
 
\end{center}
% \vspace{-3mm}
\caption{\small \textbf{Codebook usage.} Histogram of usage counts (log scale) for the $8192$ latent codes on $5000$ assets used for training.  The heavy-tailed distribution reveals a compact core of frequently reused codes that capture ubiquitous 3D primitives, while the long tail represents rarer geometric variations. In total, all of the vocabulary is exercised at least once, indicating that the codebook is not over-parameterised. Additionally, increasing codebook size did not help empirically indicating it is not under-parameterised.}
\label{fig:codebook-usage}

\end{figure*}

\mypar{Codebook usage.} Figure \ref{fig:codebook-usage} shows codebook utilization across 5000 training assets (log-scale, decreasing order). The heavy-tailed distribution indicates effective learning: frequent codes capture common 3D primitives while rare codes encode geometric variations. Active utilization of the whole codebook indicates it is not over-parameterised. Additionally, increasing codebook size did not help empirically indicating it is not under-parameterised.

The full training script and code for 3D VQ-VAE are included in the \texttt{code.zip} file. The code was built on top of Trellis~\cite{xiang2024structured} code.

\subsection{Encoding of numbers}
\label{subsec:app:encoding-of-numbers}

In Section 3.2.2 (Figure 6), we showed the robustness of a hybrid approach that we used for encoding numbers, where the embeddings are learned but are augmented with sine-cosine encodings. Here, we provide more details on how we implement the encodings. Specifically, if we have $N$ numbers to encode, then the $n^{th}$ number is encoded using an embedding of dimension $d$. First, we obtain the sine-cosine encoding as follows:

\begin{align}
NE_{(pos, 2i)}   &= \sin \left( \frac{pos}{10000^{\frac{2i}{d}}} \right), \\
NE_{(pos, 2i+1)} &= \cos \left( \frac{pos}{10000^{\frac{2i}{d}}} \right)
\end{align}

where $i$ indexes the embedding dimensions. On top of this, we incorporate a learned embedding layer, implemented as $\texttt{nn.Embedding}$ in PyTorch, of the same dimensionality $d$. The final representation for each number is obtained by summing its learned embedding with the corresponding static sine-cosine encoding. This hybrid approach allows the model to leverage both the flexibility of learned embeddings and the structured inductive bias introduced by the sine-cosine encoding for numbers.

\subsection{Compute resources and time}
\label{subsec:app:compute-resources-and-time}

Experiments were executed on a single Ubuntu 22.04.5 server equipped with 8$\times$NVIDIA A100–SXM4 GPUs (80 GB each, driver 570.124, CUDA 12.8), dual-socket AMD EPYC 7763 CPUs (256 threads) and 2 TB RAM. The software stack comprised Python 3.11 and PyTorch 2.4.1 built against CUDA 12.1 and cuDNN 9.1. Each model was trained on a single GPU, with an average training throughput of $\sim 8,\!800$ tokens sec$^{-1}$ per GPU for all language model training. The LLM experiments were implemented using the torchtune PyTorch library, providing a unified framework for fine-tuning language models. We employed PagedAdamW8bit optimizer with learning rate $1 \times 10^{-4}$, batch size 32, and trained for 10 epochs using bfloat16 precision and used the cross-entropy loss function.

The code and configs are included in the \texttt{code.zip} file. The code was built on top of the torchtune code.

\section{Additional experiments and observations}
\label{sec:app:additional-experiments-and-observations}

\begin{figure*}[!htbp]
\begin{center}
\includegraphics[width=0.6\linewidth]{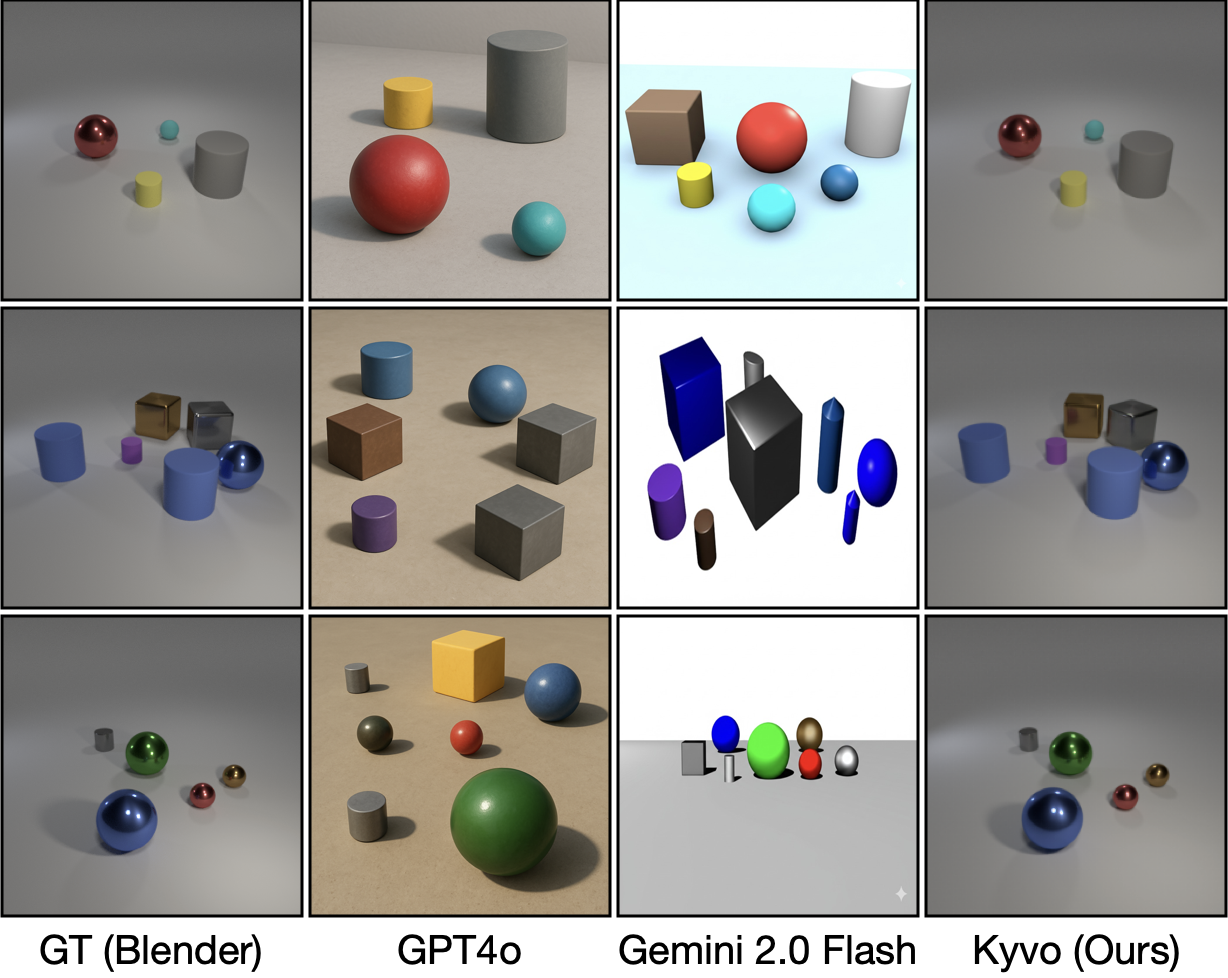} 
\end{center}
% \vspace{-3mm}
\caption{\small \textbf{Failure of modern-day LLMs on rendering task.} Each row depicts one test scene described by our 3D structured modality; columns show the ground-truth Blender render (GT) and images produced by GPT4o, Gemini 2.0 Flash, and \llm3 (ours). GPT4o and Gemini frequently hallucinate extra objects, omit specified ones, or displace shapes, violating fundamental xyz and relational constraints.}
\label{fig:llm-failure}

\end{figure*}

\begin{figure*}[!htbp]
\begin{center}
\includegraphics[width=0.9\linewidth]{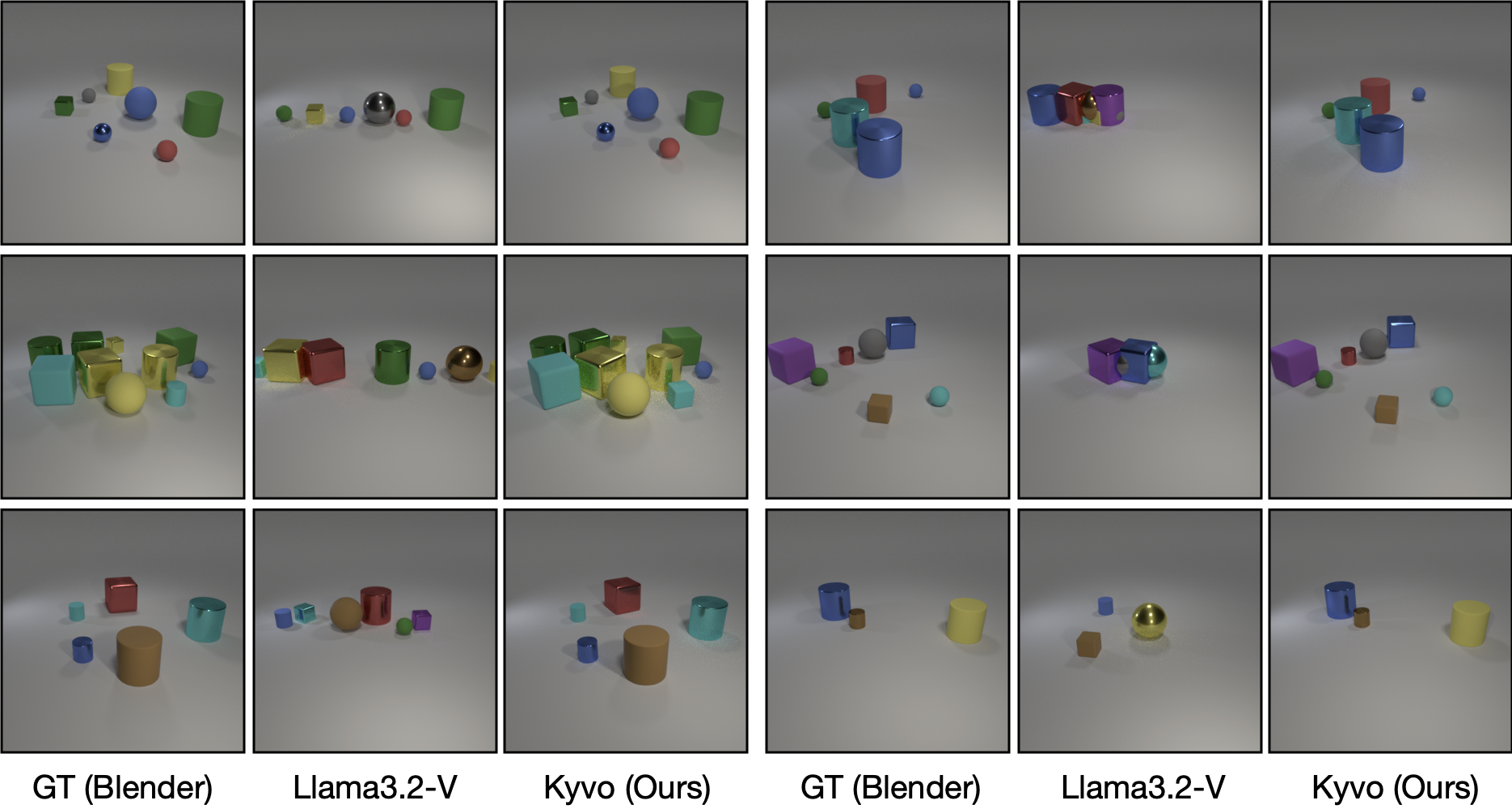} 
\end{center}
% \vspace{-3mm}
\caption{\small \textbf{Failure of Llama3.2-V on recognition task.} For six randomly selected scenes, we render (from left to right) the ground-truth scene, the recognition prediction scene by Llama 3.2-Vision, and the predicted scene by Kyvo (ours). Llama 3.2-Vision frequently collapses objects toward the centre or misplaces them entirely, failing to recover the true spatial layout even in this simple synthetic setting. Moreover, it also misidentifies some objects in the scene.}
\label{fig:llm-failure-llamav}

\end{figure*}

\mypar{Failure of modern-day LLMs.} To demonstrate the complexity of our 3D tasks, we evaluated several state-of-the-art large language models on simple CLEVR scenes. While CLEVR scenes appear visually simple, the tasks they address are complex. For instance, Google’s Gemini-Pro and OpenAI’s latest GPT4o, when prompted to generate an image with CLEVR’s 3D scene specifications, produce wrong images -- failing to adhere to relative object positions and often hallucinating new objects (see Figure~\ref{fig:llm-failure}). For the recognition task, Meta’s state-of-the-art VLM, Llama3.2-Vision (from the same family of backbones as ours), achieves near-zero performance. Figure~\ref{fig:llm-failure-llamav} visualizes Llama3.2-Vision's predictions rendered in Blender alongside ground truth scenes and the predictions from \llm3 recognition model. As can be observed, it fails to accurately predict xyz coordinates, despite the simplicity of CLEVR scenes and its objects. These failures demonstrate that 3D spatial reasoning remains a significant challenge for internet-scale trained models, even within controlled synthetic environments, underscoring the difficulty of the tasks we address.


% ---------------- wrapped, compact table ----------------
\setlength{\tabcolsep}{3pt}
% \begin{wraptable}{r}{0.55\linewidth}   % right, 42 % of text width
\begin{table}[htbp]
  % \vspace{-2mm}
  \centering
  \scriptsize
  \renewcommand{\arraystretch}{0.95}    % slightly tighter rows

  \resizebox{0.9\linewidth}{!}{%
    \begin{tabular}{@{}l
      S[table-format=1.2]   % Rendering
      S[table-format=1.4]   % Recognition
      S[table-format=1.4]   % Instruction
      S[table-format=1.4]@{}}% QA
      \toprule
      \textbf{Number Encoding} &
        {Rendering$({\downarrow})$} &
        {Recognition$({\uparrow})$} &
        {Instruction$({\uparrow})$} &
        {QA$({\uparrow})$} \\ \midrule
      Fixed Sine–Cosine              & 1.44 & $\mathbf{0.9229}$ & $\mathbf{0.8678}$ & 0.4845 \\
      Learned                        & 1.58 & 0.9185          & 0.8572          & 0.4680 \\
      \makecell[l]{Fixed Sine–Cosine + Learned} &
                                      $\mathbf{1.28}$ & 0.9212 & 0.8666 & $\mathbf{0.4980}$ \\
      \bottomrule
    \end{tabular}}%
  \vspace{2mm}
  \caption{\small\textbf{Encoding strategies for numbers.}  
           Performance of each strategy across the four tasks.}
  \label{table:static-learned-encoding}
  % \vspace{-3mm}
\end{table}
% \end{wraptable}
% --------------------------------------------------------

\mypar{Number encodings.} As detailed in Section 3.2.2 of the main paper, we employ a hybrid approach for number encoding that combines learned representations with sine-cosine positional encodings. We demonstrated the robustness of this hybrid encoding strategy across varying data regimes in Figure 6 of the main paper, with implementation details provided in Section~\ref{subsec:app:encoding-of-numbers}. While Figure 6 assessed the performance of these strategies across data regimes, Table~\ref{table:static-learned-encoding} summarizes the performance of these encoding strategies across all four
tasks for the largest training set size.

\mypar{Embedding dimension projector.} 
We use a projector to embed the VQGAN codes and the 3D VQ-VAE codes into the same dimensional space as the Llama embeddings, creating a unified sequence of dimensions. A simple linear layer without biases proves sufficient for this task. We tried more complex alternatives, such as a two-layer MLP with ReLU activation, but did not show any notable performance improvements.

\begin{figure}[htbp]
  \centering
  % first subfigure
  \begin{subfigure}[b]{0.48\textwidth}
    \centering
    \includegraphics[width=\linewidth]{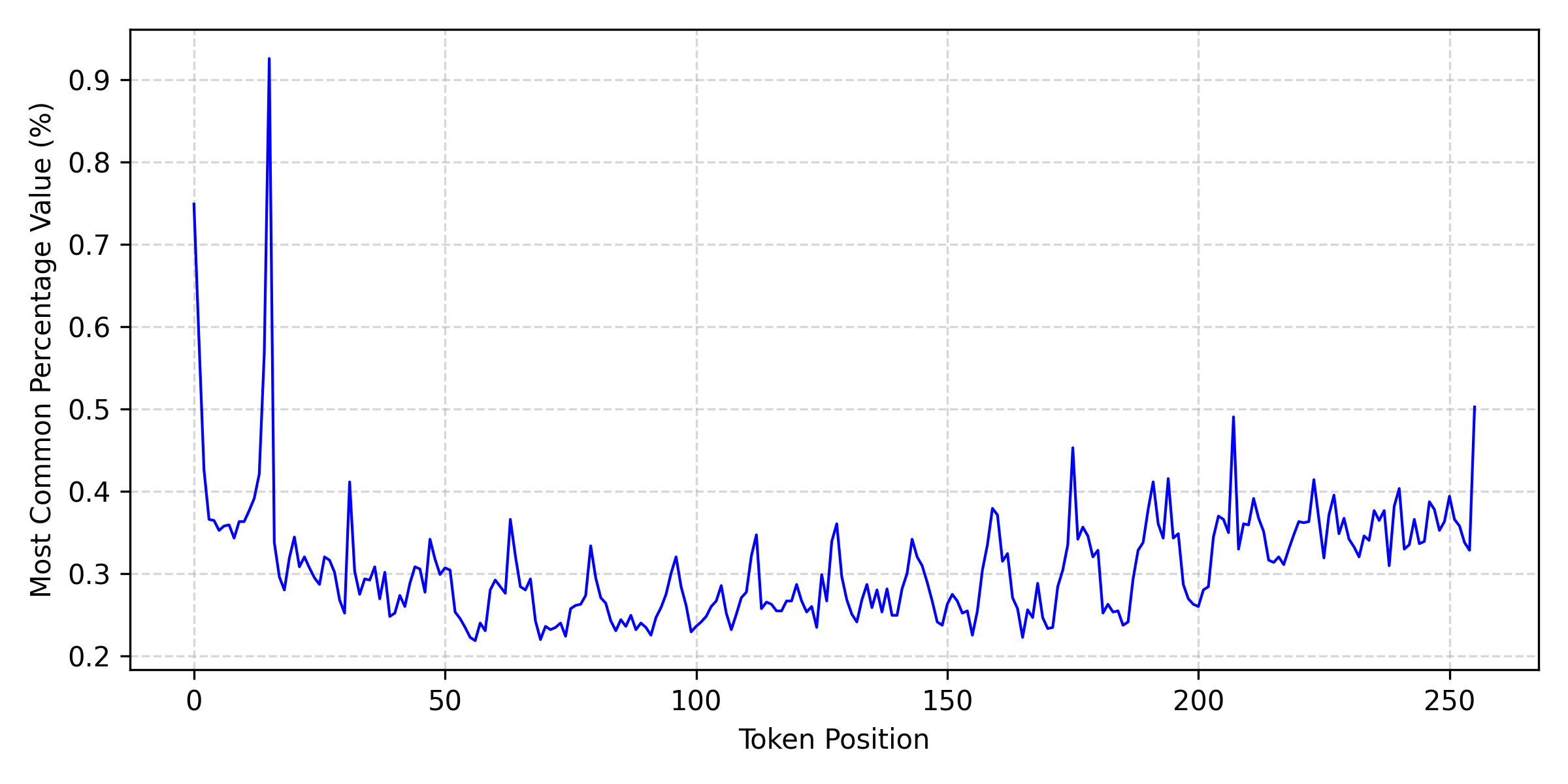}
    \caption{\small\textbf{Objectron}}
    \label{fig:img1}
  \end{subfigure}
  \hfill
  % second subfigure
  \begin{subfigure}[b]{0.48\textwidth}
    \centering
    \includegraphics[width=\linewidth]{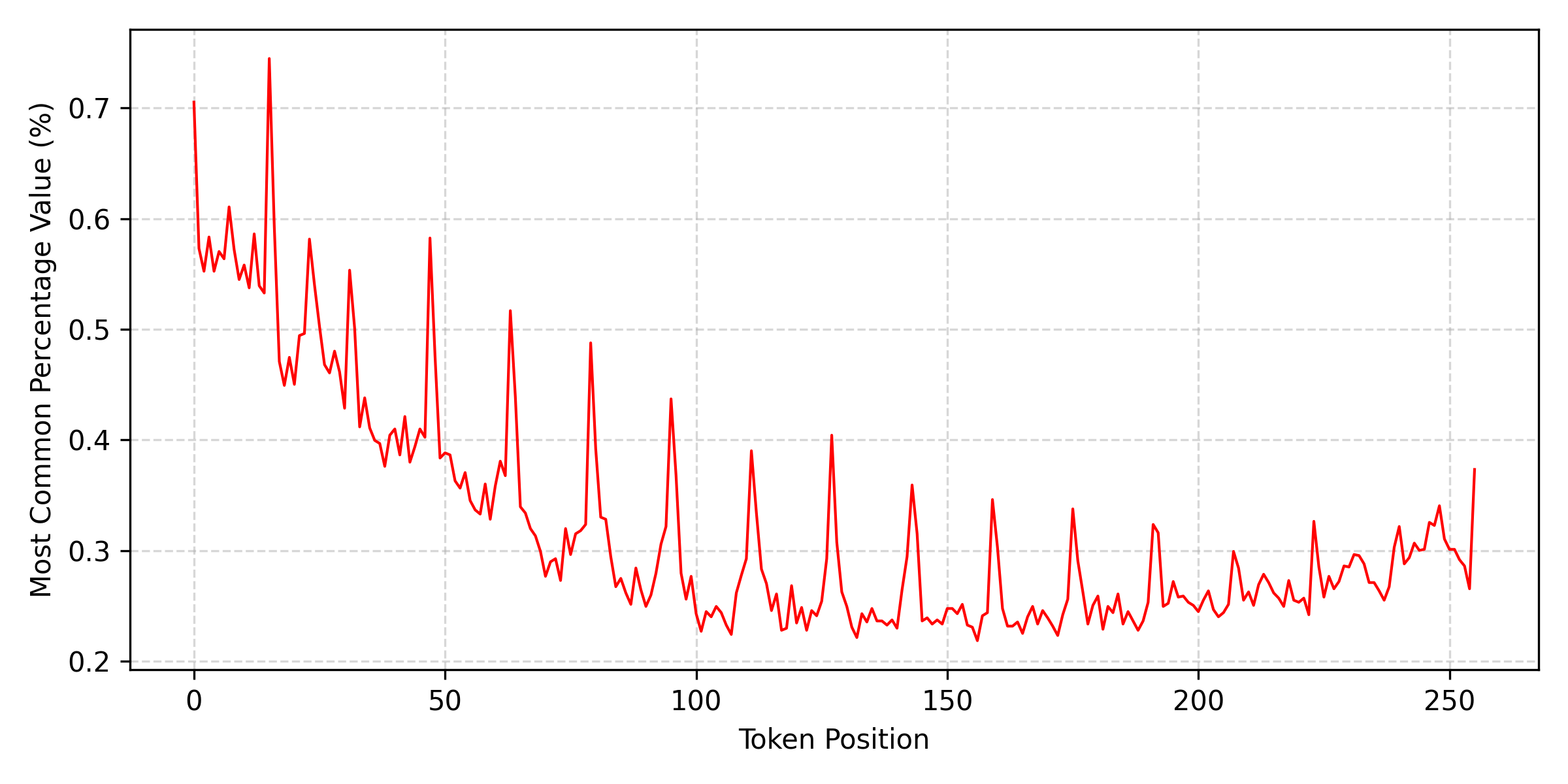}
    \caption{\small\textbf{ARKitScenes}}
    \label{fig:img2}
  \end{subfigure}
  \hfill
  \caption{\small For 256-token image sequences, the plots show the percentage of images that have the most common token at each position in Objectron (left, blue) and ARKitScenes (right, red). Early positions repeat more often, so there is still some bias, but it is weaker than in synthetic CLEVR scenes.}
  \label{fig:token-visual-real-world}
\end{figure}

\mypar{Token visualization plot.} As discussed in Section 3.2.3 of the main paper, our analysis of 256-token image sequences revealed that over $25\%$ of CLEVR images shared identical first tokens, creating a substantial bias attributed to the uniform gray backgrounds prevalent in synthetic scenes. Interestingly, in Figure~\ref{fig:token-visual-real-world} we demonstrate that similar positional biases emerge in the Objectron and ARKitScenes datasets, which contain images of real-world 3D scenes, though with notably reduced magnitude. These plots illustrate the percentage of images sharing the most frequent token value at each sequence position.

\mypar{3D scene conditioning in QA.} To assess the impact of 3D scene data on question-answering, we train a model excluding 3D scene inputs, i.e., $(\text{I}, \text{T}_\text{Q}) \rightarrow \text{T}_\text{A}$. This model achieves a Jaccard accuracy of $0.4465$, compared to $0.4980$ for the $(\text{I}, \text{3D}, \text{T}_\text{Q}) \rightarrow \text{T}_\text{A}$ model, demonstrating the critical role of 3D semantic information for accurate answers.

\section{Failure cases}
\label{sec:app:failure-cases}
In this section, we discuss a failure case of \llm3 and potential areas for improvement. Among the four tasks used in our experiments for CLEVR, \textit{Instruction-Following} presents the highest complexity. This task requires the model to process three modalities -- 3D scenes, images, and text instructions -- as input and generate both a modified image and a modified 3D scene. This requires precise comprehension of the text instruction and accurate application of the specified modifications across both the image and 3D scene sequences.

\begin{figure*}[!htbp]
\begin{center}
\includegraphics[width=0.85\linewidth]{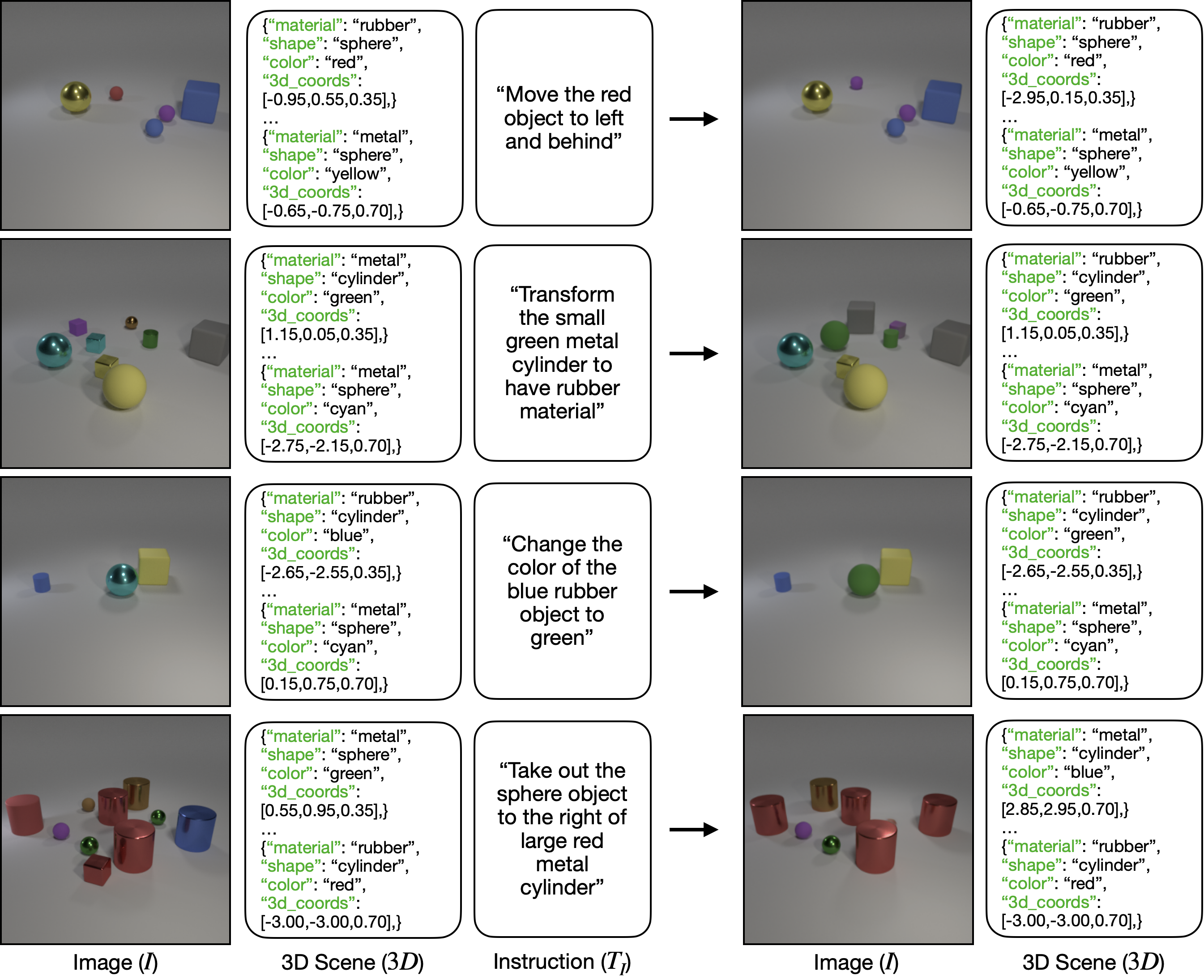} 
\end{center}
% \vspace{-3mm}
\caption{\small \textbf{Instruction-following failure cases for the image modality on CLEVR.}. As observed, the images generated by the model do not accurately reflect the intended modifications based on the input image and text instruction. On the other hand, the output 3D scenes are correct, meaning that our \llm3 accurately modified them based on the instructions. This suggests that a better avenue for predicting instruction-modified images is by task decomposition: first predict the modified 3D scene and then render the final image.}
\label{fig:instruction-following-failure}
\end{figure*}

Our experiments indicate that while \llm3 effectively predicts the modified 3D scene sequence, it struggles with image sequence modifications. In the main paper, we report Jaccard Index values for the instruction-following task, demonstrating the model's effectiveness in handling 3D scenes. Additionally, Figure~\ref{fig:instruction-following-failure} presents qualitative examples of the model's image outputs. Although the predicted images are somewhat close to the groundtruth, the model often fails to accurately modify the scene within the image sequence. For instance, in the first example, the red sphere was incorrectly assigned a purple color and was moved behind but not to the left as instructed.

While this highlights areas for improvement, an interesting direction for future work is exploring whether decomposing complex tasks like instruction-following into sequences of simpler tasks can enhance performance. For example, instead of predicting both images and 3D scenes simultaneously, the task could be divided into two stages: first, predicting the modified 3D scene, and then using a rendering model to generate the corresponding image.

\clearpage

\end{document}